\documentclass[manuscript, screen]{acmart}
\usepackage{adjustbox}
\usepackage{array}
\usepackage{amssymb}
\usepackage{amsmath, amsthm}
\usepackage{boldline, makecell}
\usepackage{booktabs} 
\usepackage{bm} 
\usepackage{csquotes}
\usepackage{colortbl}
\usepackage{diagbox}
\usepackage{dsfont}
\usepackage{epsfig}
\usepackage[mathscr]{euscript}
\usepackage{float}
\usepackage{geometry}
\usepackage{longtable}
\usepackage{lscape}
\usepackage{multirow}
\usepackage{marvosym}
\usepackage{paralist}
\usepackage{enumitem}
\usepackage[figuresright]{rotating}
\usepackage[super]{nth}
\usepackage{subfigure}
\usepackage{slashbox}
\usepackage[para,online,flushleft]{threeparttable}
\usepackage{tabularx}
\usepackage{wrapfig}
\usepackage{xcolor}
\usepackage{setspace}

\definecolor{myGray}{gray}{.9}
\newcolumntype{g}{>{\columncolor{myGray}}c}

\newcolumntype{R}[2]{%
    >{\adjustbox{angle=#1,lap=\width-(#2)}\bgroup}%
    l%
    <{\egroup}%
}
\newcommand*\rot{\multicolumn{1}{R{90}{1em}}}
\newtheorem{problemDef}{Problem}
\newtheorem{myDef}{Definition}

\usepackage[ruled]{algorithm2e} 

\acmJournal{CSUR}
\acmYear{2020} 
\acmVolume{1} 
\acmNumber{1} 
\acmArticle{1} 
\acmMonth{1} 

\acmPrice{15.00}
\acmDOI{10.1145/3395046}

\setcopyright{acmcopyright}

\acmDOI{}

\DeclareMathAlphabet\mathbfcal{OMS}{cmsy}{b}{n}
\begin{document} 
\title{{A Survey of Fake News:\\ Fundamental Theories, Detection Methods, and Opportunities}}



\author{Xinyi Zhou}
\orcid{0000-0002-2388-254X}
\affiliation{%
  \department{Data Lab, EECS Department}
  \institution{Syracuse University}
  \city{Syracuse}
  \state{NY}
  \postcode{13244}
  \country{USA}}
\email{zhouxinyi@data.syr.edu}
\author{Reza Zafarani}
\affiliation{%
  \department{Data Lab, EECS Department}
  \institution{Syracuse University}
  \city{Syracuse}
  \state{NY}
  \postcode{13244}
  \country{USA}}
\email{reza@data.syr.edu}

\begin{abstract}
The explosive growth in fake news and its erosion to democracy, justice, and public trust has increased the demand for fake news detection and intervention. This survey reviews and evaluates methods that can detect fake news from four perspectives: (1) the false \textit{knowledge} it carries, (2) its writing \textit{style}, (3) its \textit{propagation} patterns, and (4) the credibility of its \textit{source}. The survey also highlights some potential research tasks based on the review. In particular, we identify and detail related fundamental theories across various disciplines to encourage interdisciplinary research on fake news.
We hope this survey can facilitate collaborative efforts among experts in computer and information sciences, social sciences, political science, and journalism to research fake news, where such efforts can lead to fake news detection that is not only efficient but more importantly, explainable.
\end{abstract}


%
%


\maketitle
\section{Introduction} 
\label{sec::intro}

Fake news is now viewed as one of the greatest threats to democracy, journalism, and freedom of expression. It has weakened public trust in governments and its potential impact on the contentious ``Brexit'' referendum and the equally divisive 2016 U.S. presidential election -- which it might have affected~\cite{pogue2017stamp} -- is yet to be realized~\cite{zhou2019tutorial,reza2019tutorial,allcott2017social}. The reach of fake news was best highlighted during the critical months of the 2016 U.S. presidential election campaign.  During that period, the top twenty frequently-discussed fake election stories generated 8,711,000 shares, reactions, and comments on Facebook, ironically, more than the 7,367,000 for the top twenty most-discussed election stories posted by 19 major news websites~\cite{silverman2016analysis}. {Research has shown that compared to the truth, fake news on Twitter is typically retweeted by many more users and spreads far more rapidly, especially for political news~\cite{vosoughi2018spread}}. Our economies are not immune to the spread of fake news either, with fake news being connected to stock market fluctuations and large trades. For example, fake news claiming that Barack Obama, the 44th President of the United States, was injured in an explosion wiped out \$130 billion in stock value~\cite{rapoza2017can}. These events and losses have motivated fake news research and sparked the discussion around fake news, as observed by skyrocketing usage of terms such as ``post-truth'' -- selected as the international word of the year by Oxford Dictionaries in 2016~\cite{wang2016post}. 

While fake news is not a new phenomenon~\cite{tandoc2018defining}, questions such as why  it has emerged as a global topic of interest and why it is attracting increasingly more public attention are particularly relevant at this time.
The leading cause is that fake news can be created and published online faster and cheaper when compared to traditional news media such as newspapers and television~\cite{shu2017fake}. The rise of social media and its popularity also plays an essential role in this surge of interest~\cite{zafarani2014social,olteanu2019social}. As of August 2018, around two thirds (68\%) of Americans get their news from social media.\footnote{\url{https://www.journalism.org/2018/09/10/news-use-across-social-media-platforms-2018/}} With the existence of an \textit{echo chamber effect} on social media, biased information is often amplified and reinforced~\cite{jamieson2008echo}. As an ideal platform to accelerate fake news dissemination, social media breaks the physical distance barrier among individuals, provides rich platforms to share, forward, vote, and review, and encourages users to participate and discuss online news. This surge of activity around online news can lead to grave repercussions and substantial potential political and economic benefits. Such generous benefits encourage malicious entities to create, publish, and spread fake news.

Take the dozens of ``well-known'' teenagers in the Macedonian town of Veles as an example of users who created fake news for millions on social media and became wealthy by penny-per-click advertising during the U.S. presidential election. As reported by the NBC, each individual ``has earned at least \$60,000 in the past six months -- far outstripping their parents' income and transforming his prospects in a town where the average annual wage is \$4,800.''~\cite{smith2016fake}. The tendency of individuals to overestimate the benefits associated with disseminating fake news rather than its costs, as the \textit{valence effect} indicates~\cite{jones1976correspondent}, further attracts individuals to engage in fake news activities. Clearly, when governments, parties, and business tycoons are standing behind fake news generation, seeking its tempting power and profits, there is a greater motivation and capability to make fake news more persuasive and indistinguishable from truth to the public. But, how can fake news gain public trust?

Social and psychological factors play an important role in fake news gaining public trust and further facilitate the spread of fake news. For instance, humans have been proven to be irrational and vulnerable when differentiating between truth and falsehood while overloaded with deceptive information. Studies in social psychology and communications have demonstrated that human ability to detect deception is only slightly better than chance: typical accuracy rates are in the 55\%-58\% range, with a mean accuracy of 54\% over 1,000 participants in over 100 experiments~\cite{rubin2010deception}. The situation is more critical for fake news than other types of information. For news, where one expects authenticity and objectivity, it is relatively easier to gain public trust. In addition, individuals tend to trust fake news after repeated exposures (\textit{validity effect}~\cite{boehm1994validity}), or if it confirms their preexisting beliefs (\textit{confirmation bias}~\cite{nickerson1998confirmation}) { or attitudes (\textit{selective exposure}~\cite{freedman1965selective,metzger2015cognitive}), or if it pleases them (\textit{desirability bias}~\cite{fisher1993social})}. \textit{Peer pressure} can also at times ``control'' our perception and behavior (e.g., \textit{bandwagon effect}~\cite{leibenstein1950bandwagon}). 

The many perspectives on what fake news is, what characteristics and nature fake news or those who disseminate it share, and how fake news can be detected motivate the need for a comprehensive introduction and in-depth analysis, which this survey aims to develop. {In addition, this survey aims to attract researchers within general areas of data mining, machine learning, graph mining, Natural Language Processing (NLP), and Information Retrieval (IR). More importantly, we hope to boost collaborative efforts among experts in computer and information sciences, political science, journalism, social sciences, psychology, and economics to study fake news, where such efforts can lead to fake news detection that is not only efficient but more importantly, explainable~\cite{reza2019tutorial,zhou2019tutorial}. 

To achieve these goals, we first discuss the ways to define fake news (see Section \ref{sec::def}) and summarize related fundamental theories across disciplines (e.g., in social sciences and economics) that can help study fake news (see Section \ref{sec::theories}). Before further specification, we present an overview of this survey in Section \ref{sec:overview}.

\subsection{What is Fake News?}
\label{sec::def}

There has been no universal definition for fake news, even in journalism. A clear and accurate definition helps lay a solid foundation for fake news analysis and evaluating related studies. Here we (I) distinguish between several concepts that frequently co-occur or have overlaps with fake news, (II) present a broad and a narrow definition for the term \textit{fake news}, justifying each definition, and (III) further discuss the potential research problems raised by such definitions.

\paragraph{I. Related Concepts}
Existing studies often connect fake news to terms and concepts such as \textit{deceptive news}~\cite{allcott2017social,shu2017fake,lazer2018science}, \textit{false news}~\cite{vosoughi2018spread}, \textit{satire news}~\cite{rubin2015deception,tandoc2018defining,wardle2017fake}, \textit{disinformation}~\cite{kshetri2017economics,wardle2017fake}, \textit{misinformation}~\cite{kucharski2016post,wardle2017fake}, {\textit{cherry-picking}~\cite{asudeh13detecting}, \textit{clickbait}~\cite{chen2015misleading}} and \textit{rumor}~\cite{zubiaga2018detection}. Based on how these terms and concepts are defined, we can distinguish one from the others based on three characteristics: 
(i)~\textbf{authenticity} ({containing \underline{any} non-factual statement or not}), 
(ii)~\textbf{intention} (aiming to mislead or entertain the public), and 
(iii)~\textbf{whether the information is news}. Table \ref{tab::def} summarizes these related concepts based on these characteristics. For example, disinformation is false information [news or not-news] with a malicious intention to mislead the public.

\begin{table}[t]
\small
\caption{A Comparison between Concepts related to Fake News}
\label{tab::def}
\begin{threeparttable}
\begin{tabular}{|l|c|c|c|} \hline
\rowcolor{gray!20}
\hline
\textbf{Concept} & \textbf{Authenticity} & \textbf{Intention} & \textbf{News?} \\ \hline \hline
\textbf{Deceptive news} & Non-factual & Mislead & Yes \\ \hline
\textbf{False news} & Non-factual & Undefined & Yes \\ \hline
\textbf{Satire news} & Non-unified\footnotemark & Entertain & Yes \\ \hline 
\textbf{Disinformation} & Non-factual & Mislead & Undefined \\ \hline 
\textbf{Misinformation} & Non-factual & Undefined & Undefined \\ \hline 
{\textbf{Cherry-picking}} & Commonly factual & Mislead & Undefined \\ \hline 
{\textbf{Clickbait}} & Undefined & Mislead & Undefined \\ \hline
\textbf{Rumor} & Undefined & Undefined & Undefined \\ \hline 
\end{tabular}
\end{threeparttable}
\end{table}

\footnotetext{ For example, Golbeck et al. regard satire news as ``factually incorrect''~\cite{golbeck2018fake} while Tandoc Jr et al. state that ``where parodies differ from satires is their use of non-factual information to inject humor.''~\cite{tandoc2018defining}}

\paragraph{II. Defining Fake News}
{Challenges of fake news research start from defining fake news. To date no universal definition is provided for fake news, where it has been looked upon as ``a news article that is intentionally and verifiably false''~\cite{allcott2017social,shu2017fake} (deceptive news), ``a news article or message published and propagated through media, carrying false information regardless of the means and motives behind it'' which overlaps with false news, disinformation~\cite{kshetri2017economics}, misinformation~\cite{kucharski2016post}, satire news~\cite{rubin2015deception}, or even the stories that a person does not like (considered improper)~\cite{golbeck2018fake}. 
Furthermore, what news is has become harder to define as it can range from an account of a recent, interesting, and significant event, to a dramatic account of something novel or deviant; in particular, ``the digitization of news has challenged traditional definitions of news. Online platforms provide space for non-journalists to reach a mass audience.''~\cite{tandoc2018defining}. Under these circumstances, we first broadly define fake news as:}

\begin{myDef}[Broad definition of fake news]
\label{def::generalized}
Fake news is false news,
\end{myDef}

\noindent where news broadly includes articles, claims, statements, speeches, posts, among other types of information related to public figures and organizations. It can be created by journalists and non-journalists. {Such definition of news raises some social concerns, e.g., the term ``fake news'' should be ``about more than news'' and ``about the entire information ecosystem.''~\cite{wardle2017fake}}
The broad definition aims to impose minimum constraints in accord with the current resources: it emphasizes information authenticity, purposefully adopts a broad definition for the term news~\cite{vosoughi2018spread}, and weakens the requirement for information intention {due to the difficulty in obtaining the ground truth (true intention)}. This definition supports most existing fake-news-related studies and datasets, as provided by the existing fact-checking websites (Section \ref{subsec::manual} provides a detailed introduction). Current fake news datasets often provide ground truth for the authenticity of claims, statements, speeches, or posts related to public figures and organizations, while limited information is provided on intentions. 

We provide a more narrow definition of fake news, which satisfies the overall requirements for fake news as follows.

\begin{myDef}[Narrow definition of fake news]
 \label{def::narrow}
Fake news is intentionally false news published by a news outlet.
\end{myDef}

{This narrow definition supports recent advancements in fake news studies~\cite{allcott2017social,shu2017fake}.} It addresses the public's perception of fake news, especially following the 2016 U.S. presidential election. Note that deceptive news is more harmful and less distinguishable than incautiously false news, as the former pretends to be truth to mislead the public better. The narrow definition emphasizes both news authenticity and intentions; it also ensures the posted information is news by investigating if its publisher is a news outlet { (e.g., CNN and New York Times). Often news outlets publish news in the form of articles with fixed components: a title, author(s), a body text, image(s) and/or video(s) that include the claims made by, or about, public figures and organizations.} 

{Both definitions require the authenticity of fake news to be false (i.e., being non-factual). As the goal is to provide a scientific definition for fake news; hence, news falsity should be derived by comparing with \underline{objective} facts and not with individual viewpoints (preferences). Hence, it is improper to consider fake news as articles that do not agree with individuals' or groups' interests or viewpoints, which is sometimes how the term fake news is used by the general public or in politics~\cite{golbeck2018fake}.} Such falsity can be assigned to the whole or part of the news content, or even to true news when subsequent events have rendered the original truth outdated (e.g., ``Britain has control over fifty-six colonial countries''). In this general framework, a more comprehensive strategy for automatic fake news detection is needed, as the aforementioned fake news types emphasize various aspects of detection (see Section \ref{sec::discussion} for a discussion).

\paragraph{III. Discussion}
We have differentiated between fake and fake news-related terms based on three properties (authenticity, intention, and if it is news). We have also defined fake news as (1) deceptive news, narrowly and as (2) false news, broadly. Questions are thus left, such as how to [manually or automatically] identify the authenticity and intention of the given information (news)? Many domain-experts and platforms have investigated ways to analyze news authenticity manually; however, how one can automatically assess news authenticity in an effective and explainable manner is still an open issue. We will detail both manual and automatic assessment of news authenticity (also known as \textit{fact-checking}) in Section \ref{sec::knowledge}.
To assess information (news) intention, analyzing the news (i) writing style and its (ii) propagation patterns can be useful. First, the information (news) created to mislead or deceive the public intentionally (e.g., deceptive news) should look or sound ``more persuasive'' compared to the news without such intentions (e.g., satire news). Secondly, malicious users should play a part in the propagation of deceptive information (news) to enhance its social influence. Both news writing styles and propagation characteristics will be discussed with current methods in Sections \ref{sec::style}-\ref{sec::credibility}. For intention analysis, often some level of manual news annotation is necessary. The accuracy of such annotations dramatically impacts automatic intention analysis within a machine learning framework. When the intentions behind non-factual information are determined, intervention strategies can be more appropriate and effective. For example, punitive measures should be taken against non-factual information and those who intentionally create it. 

{
\subsection{Fundamental Theories} 
\label{sec::theories}

Fundamental human cognition and behavior theories developed across various disciplines, such as social sciences and economics, provide invaluable insights for fake news analysis. These theories can introduce new opportunities for qualitative and quantitative studies of \underline{big} fake news data~\cite{zhou2019content}. These theories can also facilitate building well-justified and explainable models for fake news detection and intervention, which, to date, have been rarely available~\cite{miller2017explainable}.
We have conducted a comprehensive literature survey across various disciplines and have identified well-known theories that can be potentially used to study fake news. These theories are provided in Table \ref{tab::theories} along with short descriptions, which are related to either (I) the news itself or (II) its spreaders. 

\paragraph{I. News-related theories}
News-related theories reveal the possible characteristics of fake news content compared to true news content. For instance, theories have implied that fake news potentially differs from the truth in terms of, e.g., writing style and quality (by \textit{Undeutsch hypothesis})~\cite{undeutsch1967beurteilung},  quantity such as word counts (by \textit{information manipulation theory})~\cite{mccornack2014information}, and sentiments expressed (by \textit{four-factor theory})~\cite{zuckerman1981verbal}.
It should be noted that these theories, developed by forensic psychology, target deceptive statements or testimonies (i.e., disinformation) but not fake news, though these are similar concepts (see Section \ref{sec::def} for details). Thus, one research opportunity is to verify whether these attributes (e.g., information sentiment polarity) are statistically distinguishable among disinformation, fake news, and the truth, in particular, using big fake news data. On the other hand, these (discriminative) attributes identified can be used to automatically detect fake news using its writing style, where a typical study using supervised learning can be seen in \cite{zhou2019content}; we will provide further details in Section \ref{sec::style}.

\begin{table}[t]
\centering
\caption{Fundamental Theories in Social Sciences (Including Psychology and Philosophy) and Economics}
\label{tab::theories}
\begin{adjustbox}{max width=\textwidth}
\begin{tabular}{|c|c|l|l|}
\hline
\rowcolor{gray!20}
\multicolumn{2}{|l|}{} & \textbf{Theory} & \textbf{Phenomenon}  \\ \hline \hline                                                  
\multicolumn{2}{|c|}{\multirow{8}{*}{\rotatebox{90}{\begin{tabular}[c]{@{}c@{}} \textbf{News-related} \\   \textbf{Theories}\end{tabular}}}}

& \textit{Undeutsch hypothesis}  & \multirow{2}{13cm}{A statement based on a factual experience differs in content style and quality from that of fantasy.}\\ 
\multicolumn{2}{|c|}{} & \cite{undeutsch1967beurteilung} & \\ \cline{3-4}

\multicolumn{2}{|c|}{} & \textit{Reality monitoring}  & \multirow{2}{13cm}{Actual events are characterized by higher levels of sensory- perceptual information.}  \\ 
\multicolumn{2}{|c|}{} & \cite{johnson1981reality} & \\ \cline{3-4}      
\multicolumn{2}{|c|}{} & \textit{Four-factor theory}  & \multirow{2}{13cm}{Lies are expressed differently in terms of arousal, behavior control, emotion, and thinking from truth.}  \\ 
\multicolumn{2}{|c|}{} & \cite{zuckerman1981verbal} & \\ \cline{3-4}

\multicolumn{2}{|c|}{} & { \textit{Information manipulation theory}}  & \multirow{2}{13cm}{Extreme information quantity often exists in deception.}  \\ 
\multicolumn{2}{|c|}{} & \cite{mccornack2014information} & \\ 
\hline

\multirow{36}{*}{\rotatebox{90}{ \textbf{User-related Theories} (User's Engagements and Roles in Fake News Activities)}}
& \multirow{18}{*}{\rotatebox{90}{\textbf{Social Impacts}}} &
\textit{Conservatism bias}  & \multirow{2}{13cm}{The tendency to revise one's belief insufficiently when presented with new evidence.}  \\ 
& & \cite{basu1997conservatism} & \\ \cline{3-4}

& & \textit{Semmelweis reflex} & \multirow{2}{13cm}{Individuals tend to reject new evidence because it contradicts with established norms and beliefs.}  \\ 
& & \cite{balint2009semmelweis} & \\ \cline{3-4}     

& & \textit{Echo chamber effect} & \multirow{2}{13cm}{Beliefs are amplified or reinforced by communication and repetition within a closed system.}  \\
& &
\cite{jamieson2008echo} & \\ \cline{3-4}

& &\textit{Attentional bias} & \multirow{2}{13cm}{An individual's perception is affected by his or her recurring thoughts at the time.} \\ 
& & \cite{macleod1986attentional} &  \\\cline{3-4}

& & \textit{Validity effect}  & \multirow{2}{13cm}{Individuals tend to believe information is correct after repeated exposures.}  \\  
& &
\cite{boehm1994validity} & \\ \cline{3-4}

& & \textit{Bandwagon effect} & \multirow{2}{13cm}{Individuals do something primarily because others are doing it.}   \\
& &
\cite{leibenstein1950bandwagon} & \\ \cline{3-4}

& & \textit{Normative influence theory} & \multirow{2}{13cm}{The influence of others leading us to conform to be liked and accepted by them.} \\ 
& &
\cite{deutsch1955study} & \\ \cline{3-4}

& & \textit{Social identity theory}  & \multirow{2}{13cm}{An individual's self-concept derives from perceived membership in a relevant social group.}  \\
& &
\cite{ashforth1989social} & \\ \cline{3-4}
 
& & \textit{Availability cascade}  & \multirow{2}{13cm}{Individuals tend to adopt insights expressed by others when such insights are gaining more popularity within their social circles}  \\ 
& &
\cite{kuran1999availability} & \\ \cline{2-4}

 &  \multirow{12}{*}{\rotatebox{90}{\textbf{Self-impact}}} &\textit{Confirmation bias} &  \multirow{2}{13cm}{Individuals tend to trust information that confirms their preexisting beliefs or hypotheses.}  \\
& & \cite{nickerson1998confirmation} & \\ \cline{3-4}

& & { \textit{Selective exposure}} &\multirow{2}{13cm}{Individuals prefer information that confirms their preexisting attitudes.}   \\ 
& &
\cite{freedman1965selective} & \\ \cline{3-4} 

& & { \textit{Desirability bias}} &\multirow{2}{13cm}{Individuals are inclined to accept information that pleases them.}   \\
& & \cite{fisher1993social} & \\ \cline{3-4}

& &
\textit{Illusion of asymmetric insight} &\multirow{2}{13cm}{Individuals perceive their knowledge to surpass that of others.}   \\ 
& &
\cite{pronin2001you} & \\ \cline{3-4}

& &
\textit{Na\"ive realism}
&\multirow{2}{13cm}{The senses provide us with direct awareness of objects as they really are.}
\\ 
& & 
\cite{ward1997naive} & \\ \cline{3-4}

& & 
\textit{Overconfidence effect} & \multirow{2}{13cm}{A person's subjective confidence in his judgments is reliably greater than the objective ones.} \\ 
& & \cite{dunning1990overconfidence} & \\ \cline{2-4}

 &  \multirow{6}{*}{\rotatebox{90}{\textbf{Benefits}}} &
\textit{Prospect theory} & \multirow{2}{13cm}{People make decisions based on the value of losses and gains rather than the outcome.} \\
& &
\cite{kahneman2013prospect} & \\ \cline{3-4}

& &
\textit{Contrast effect} & \multirow{2}{13cm}{The enhancement or diminishment of cognition due to successive or simultaneous exposure to a stimulus of lesser or greater value in the same dimension.}  \\ 
& &
\cite{hovland1957assimilation} & \\ \cline{3-4}

& &
\textit{Valence effect} & \multirow{2}{13cm}{People tend to overestimate the likelihood of good things happening rather than bad things.}  \\ 
& & \cite{frijda1986emotions} & \\ \hline 
\end{tabular}
\end{adjustbox}
\end{table}

\paragraph{II. User-related theories}
User-related theories investigate the characteristics of users involved in fake news activities, e.g., posting, forwarding, liking, and commenting. Fake news, unlike information such as fake reviews~\cite{jindal2008opinion}, can ``attract'' both malicious and normal users~\cite{shao2018spread}.
Malicious users (e.g., some social bots~\cite{ferrara2016rise}) spread fake news often intentionally and are driven by \textit{benefits}~\cite{kahneman2013prospect,hovland1957assimilation}. Some normal users (which we denote as \textit{vulnerable} normal users) can frequently and unintentionally spread fake news without recognizing the falsehood. Such vulnerability psychologically stems from 
(i)~\textit{social impacts} and 
(ii)~\textit{self-impact}, where theories have been accordingly categorized and detailed in Table \ref{tab::theories}. Specifically, as indicated by the \textit{bandwagon effect}~\cite{leibenstein1950bandwagon}, \textit{normative influence theory}~\cite{deutsch1955study}, \textit{social identity theory}~\cite{ashforth1989social}, and \textit{availability cascade}~\cite{kuran1999availability}, to be liked and/or accepted by the community, normal users are encouraged to engage in fake news activities when many users have done so (i.e., \textit{peer pressure}). One's trust to fake news and his or her unintentional spreading can be promoted as well when being \textit{exposed} more to fake news (i.e., \textit{validity effect})~\cite{boehm1994validity}, which often takes place due to the \textit{echo chamber} effect on social media~\cite{jamieson2008echo}. Such trust to fake news can be built when the fake news confirms one's \textit{preexisting attitudes, beliefs or hypotheses} (i.e., \textit{confirmation bias}~\cite{nickerson1998confirmation}, \textit{selective exposure}~\cite{freedman1965selective}, and \textit{desirability bias}~\cite{fisher1993social}), which are often perceived to surpass that of others~\cite{pronin2001you,ward1997naive,dunning1990overconfidence} and tend to be insufficiently revised when new refuting evidence is presented~\cite{basu1997conservatism,balint2009semmelweis}. In such settings, strategies for intervening fake news from a user perspective (more discussions on fake news intervention are in Section \ref{sec::discussion}) should be cautiously designed for users with different levels of credibility or intentions, even though they might all engage in the same fake news activity. For instance, it is reasonable to intervene with the spread of fake news by penalizing (e.g., removing) malicious users, but not for normal accounts. Instead, education and personal recommendations of true news articles and refuted fake ones can be helpful for normal users~\cite{vo2018rise}. Such recommendations should not only cater to the topics that the users want to read but should also capture topics that users are most gullible to. In Section \ref{sec::credibility}, we will provide the path for utilizing these theories, i.e.,  quantifying social and self-impact, to enhance fake news research by identifying user intent and evaluating user credibility. 

\vspace{0.5em}
Meanwhile, we should point out that clearly understanding the potential roles that the fundamental theories listed in Table \ref{tab::theories} can play in fake news research requires further in-depth investigations of interdisciplinary nature.}
}

\begin{figure}[t]
  \includegraphics[width=1\textwidth]{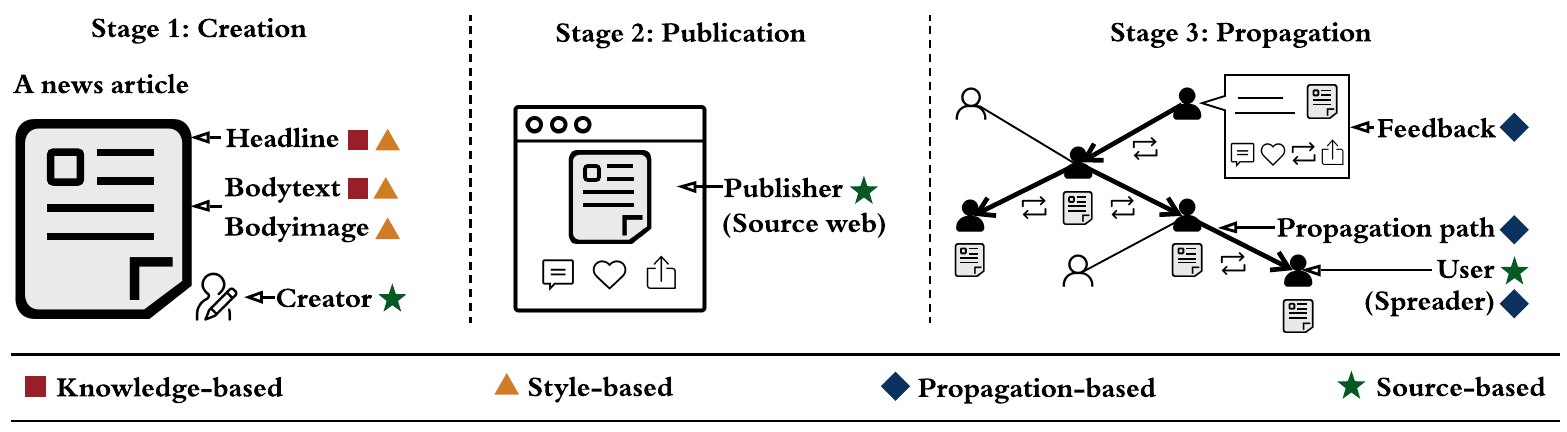}
  \caption{Fake News Life Cycle and Connections to the Four Fake News Detection Perspectives Presented in this Survey}
\label{fig::framework}
\end{figure}

\subsection{An Overview of this Survey}
\label{sec:overview}

We have defined fake news (Section \ref{sec::def}) and presented relevant fundamental theories in various disciplines (Section \ref{sec::theories}). The rest of this survey is organized as follows. We detail the detection of fake news from four perspectives (see Fig. \ref{fig::framework} for an overview): (I) \textit{Knowledge-based} methods (Section \ref{sec::knowledge}), which detect fake news by verifying if the knowledge within the news content (text) is consistent with facts (true knowledge); 
(II) \textit{Style-based} methods (Section \ref{sec::style}) are concerned with how fake news is written (e.g., if it is written with extreme emotions); 
(III) \textit{Propagation-based} methods (Section \ref{sec::propagation}), where they detect fake news based on how it spreads online; and 
(IV) \textit{Source-based} methods (Section \ref{sec::credibility}) detect fake news by investigating the credibility of news sources at various stages (being created, published online, and spread on social media).
In Section \ref{sec::discussion}, we discuss open issues in current fake news studies and in fake news detection. We highlight six potential research tasks, hoping to facilitate the development of fake news research. We conclude in Section \ref{sec::conclusion}.

\paragraph{Comparison to Related Surveys} Our survey varies from related surveys from three perspectives. First, we discuss the ways fake news is defined in the current fake news research and its harmfulness to the public. We detail how fake news is related to terms such as    deceptive news, false news, satire news, disinformation, misinformation, cherry-picking, clickbait, and rumor. Compared to related surveys and forums that often provide a specific definition for fake news, this survey highlights the challenges of defining fake news and introduces both a narrow and a broad definition for it.

Second, though recent studies have highlighted the importance of multidisciplinary fake news research~\cite{lazer2018science}, we provide a path towards it by conducting an extensive literature survey across various disciplines, identifying a comprehensive list of well-known theories. We demonstrate how these theories relate to fake news and its spreaders and illustrate technical methods utilizing these theories both in fake news detection and intervention.

Third, for fake news detection, current surveys have mostly limited their scope to reviewing research from a certain perspective (or within a certain research area, e.g., NLP~\cite{oshikawa2018survey} and data mining~\cite{shu2017fake}). These surveys generally classify fake news detection models by the types of [deep] machine learning methods used~\cite{oshikawa2018survey} or by whether they utilize social context information~\cite{shu2017fake}. In our survey, we categorize automatic fake news detection methods from four perspectives: \textit{knowledge} (Section \ref{sec::knowledge}), \textit{style} (Section \ref{sec::style}), \textit{propagation} (Section \ref{sec::propagation}), and \textit{source} (Section \ref{sec::credibility}). Reviewing and organizing fake news detection studies in such a way allows analyzing both \textit{news content} (mainly Sections \ref{sec::knowledge}-\ref{sec::style}) and the \textit{medium} (often, social media) on which the news spreads (Sections \ref{sec::propagation}-\ref{sec::credibility}), where fake news detection can be defined as a \textit{probabilistic/regression problem} linked to, e.g., \textit{entity resolution} and \textit{link prediction} tasks (Section \ref{sec::knowledge}), or a \textit{classification problem} that relies on, e.g., \textit{feature engineering} and \textit{[text and graph] embedding} techniques (Sections \ref{sec::style}-\ref{sec::credibility}). In our survey of fake news detection, \textit{patterns} of fake news in terms of its content (text and images, see Figs. \ref{fig:pattern_text}-\ref{fig:pattern_image}) or how it propagates (see Fig. \ref{fig::pattern_propagation} and Fig. \ref{fig:pattern_network}) are revealed, \textit{algorithms} and \textit{model architectures} are presented (e.g., Figs. \ref{fig:multimodal}, \ref{fig::cascadeDetection}, \ref{fig:homoNet}-\ref{fig:hieraNet}), and \textit{performance} of various fake news detection methods are compared (e.g., Table~\ref{tab:mlModel}). 
We point out that our survey focuses more on how to construct a fake news dataset, i.e., ground truth data, and the possible sources to obtain such ground truth (e.g., Section~\ref{subsec::manual}), rather than detailing existing datasets, which have been provided in past surveys~\cite{shu2017fake,oshikawa2018survey}. Nevertheless, we acknowledge the contributions to automatic fake news detection by these existing datasets (e.g., CREDBANK~\cite{mitra2015credbank}, 
LIAR~\cite{wang2017liar},
FakeNewsNet~\cite{shu2018fakenewsnet},
FakevsSatire~\cite{golbeck2018fake},
NELA-GT-2018~\cite{norregaard2019nela},
FEVER~\cite{thorne2018fever},
PHEME~\cite{kochkina2018all}, and
Emergent~\cite{ferreira2016emergent}) and systems that can be used for building datasets (e.g., 
ClaimBuster~\cite{hassan2017claimbuster},
XFake~\cite{yang2019xfake},
Hoaxy~\cite{shao2016hoaxy},
MediaRank~\cite{ye2019mediarank},
Botometer~\cite{davis2016botornot}, and
RumorLens~\cite{resnick2014rumorlens}.

\section{Knowledge-based Fake News Detection} 
\label{sec::knowledge}

When detecting fake news from a knowledge-based perspective, one often uses a process known as \textit{fact-checking}. Fact-checking, initially developed in journalism, aims to \underline{assess news authenticity} by comparing the knowledge extracted from to-be-verified news content (e.g., its claims or statements) with known facts. In this section, we will discuss the traditional fact-checking (also known as \textit{manual fact-checking}) and how it can be incorporated into automatic means to detect fake news (i.e., \textit{automatic fact-checking}).

\subsection{Manual Fact-checking}
\label{subsec::manual}

Broadly speaking, manual fact-checking can be divided into (I) expert-based and (II) crowd-sourced fact-checking.

\paragraph{I. Expert-based Manual Fact-checking} Expert-based fact-checking relies on domain-experts as \textit{fact-checkers} to verify the given news contents. Expert-based fact-checking is often conducted by a small group of highly credible fact-checkers, is easy to manage, and leads to highly accurate results, but is costly and poorly scales with the increase in the volume of the to-be-checked news contents.

\begin{table}[t]
\small
\caption{A Comparison among Expert-based Fact-checking Websites}
\label{tab::factCheckWeb}
\begin{minipage}{\columnwidth}
\begin{adjustbox}{max width=\textwidth}
\begin{tabular}{|p{2.4cm}|p{3.1cm}|p{2.9cm}|p{5.3cm}|} \hline
\rowcolor{gray!20} \textbf{Website} & {\textbf{Topics Covered}} & {\textbf{Content Analyzed}} & {\textbf{Assessment Labels}}  \\ \hline \hline
\textbf{PolitiFact}\footnotemark & American politics & Statements & True; Mostly true; Half true; Mostly false; False; Pants on fire \\ \hline
\textbf{The Washington Post Fact Checker}\footnotemark & American politics & Statements and claims & One pinocchio; Two pinocchio; Three pinocchio; Four pinocchio; The Geppetto checkmark; An upside-down Pinocchio; Verdict pending \\ \hline
\textbf{FactCheck}\footnotemark & American politics & TV ads, debates, speeches, interviews, and news & True; No evidence; False   \\ \hline
\textbf{Snopes}\footnotemark & Politics and other social and topical issues & News articles and videos & True; Mostly true; Mixture; Mostly false; False; Unproven; Outdated; Miscaptioned; Correct attribution; Misattributed; Scam; Legend   \\ \hline
\textbf{TruthOrFiction}\footnotemark & Politics, religion, nature, aviation, food,  medical, etc. & Email rumors & Truth; Fiction; etc.  \\ \hline
\textbf{FullFact}\footnotemark & Economy, health, education, crime, immigration, law & Articles & Ambiguity (no clear labels)  \\ \hline
\textbf{HoaxSlayer}\footnotemark & Ambiguity & Articles and messages & Hoaxes, scams, malware, bogus warning, fake news, misleading, true, humour, spams, etc. \\ \hline
\textbf{ GossipCop}\footnotemark & Hollywood and celebrities & Articles & 0-10 scale, where 0 indicates completely fake news and 10 indicates completely true news \\ \hline 
\end{tabular}
\end{adjustbox}
\end{minipage}
\end{table}
\footnotetext[5]{\url{http://www.politifact.com/} \label{url:politifact}}
\footnotetext[6]{\url{https://www.factcheck.org/}}
\footnotetext[7]{\url{https://www.washingtonpost.com/news/fact-checker}}
\footnotetext[8]{\url{https://www.snopes.com/}}
\footnotetext[9]{\url{https://www.truthorfiction.com/}}
\footnotetext[10]{\url{https://fullfact.org/}}
\footnotetext[11]{\url{http://hoax-slayer.com/}}
\footnotetext[12]{\url{https://www.gossipcop.com}}

\begin{figure}[t]
  \begin{center}
  \subfigure[(Expert-based) PolitiFact: the PolitiFact scorecard]{\label{fig::politiFact}
   \begin{minipage}[b]{0.47\textwidth}
     \includegraphics[width=0.85\textwidth]{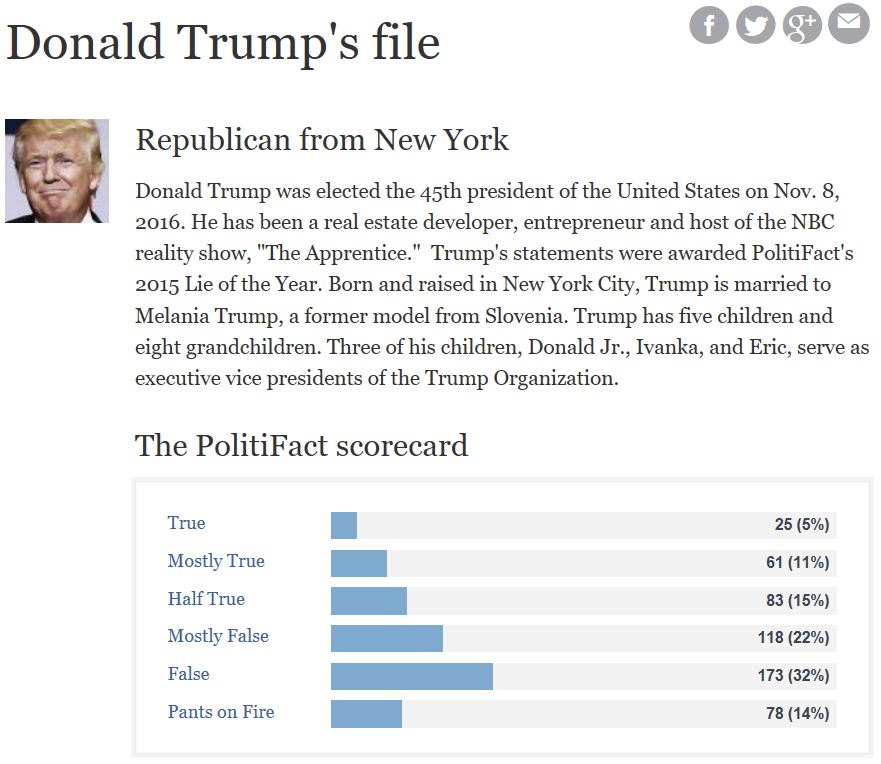}
     \end{minipage}}
  \subfigure[(Crowd-sourced) Fiskkit: the tag distribution]{ \label{fig::fiskkit}
   \begin{minipage}[b]{0.51\textwidth}
     \includegraphics[width=\textwidth]{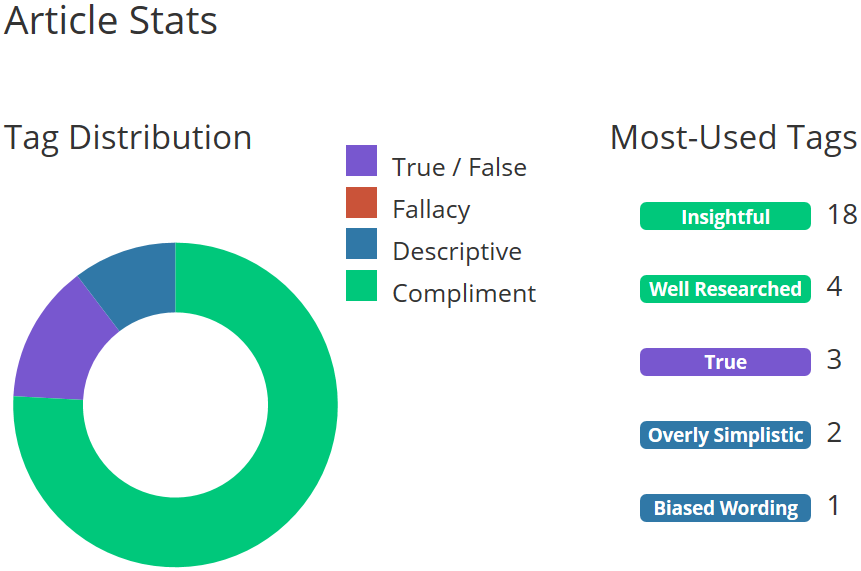}
     \end{minipage}}\vspace{-5mm}
  \caption{Illustrations of Manual Fact-checking Websites}
\label{fig::factCheckWeb}
\end{center}
\end{figure}

\paragraph{$\blacktriangleright$ Expert-based Fact-checking Websites} Recently, many websites have emerged to allow expert-based fact-checking better serve the public. We list and provide details on the well-known websites in Table \ref{tab::factCheckWeb}. Some websites provide further information, for instance, \textit{PolitiFact} provides ``the PolitiFact scorecard'', which presents statistics on the authenticity distribution of all the statements related to a specific topic (see an example on Donald Trump, the 45th President of the United States, in Fig. \ref{fig::politiFact}). { This information can provide the ground truth on the credibility of a topic~\cite{zhang2018fake}}, and help identify check-worthy topics (see Section \ref{sec::discussion} for details) that require further scrutiny for verification. Another example is \textit{HoaxSlayer}, which is different from most fact-checking websites that focus on information authenticity because it further classifies the articles and messages into, e.g., hoaxes, spams, and fake news. Though the website does not provide clear definitions for these categories, its information can be potentially exploited as ground-truth for comparative studies of fake news. In addition to the list provided here, a comprehensive list of fact-checking websites is provided by Reporters Lab at Duke University,\footnote{\url{https://reporterslab.org/fact-checking/}} where over two hundred fact-checking websites across countries and languages are listed. Generally, these expert-based fact-checking websites can provide ground-truth for the detection of fake news, in particular, under the broad definition (Definition \ref{def::generalized}). {Among these websites \textit{PolitiFact} and \textit{GossipCop} have supported the development of fake news datasets that are publicly available (e.g., LIAR~\cite{wang2017liar} and FakeNewsNet~\cite{shu2018fakenewsnet})}. The detailed expert-based analysis that these websites provide for checked contents (e.g., what is false and why is it false) carries invaluable insights for various aspects of fake news analysis, e.g., for \textit{identifying check-worthy content}~\cite{hassan2017toward} and \textit{explainable fake news detection}~\cite{shu2019defend} (see Section \ref{sec::discussion} for more discussions); however, to date, such insights have not been well utilized.

\paragraph{II. Crowd-sourced Manual Fact-checking} Crowd-sourced fact-checking relies on a large population of regular individuals acting as fact-checkers (i.e., the collective intelligence). {Such large population of fact-checkers can be gathered within some common crowd-sourcing marketplaces such as Amazon Mechanical Turk, based on which CREDBANK~\cite{mitra2015credbank}, a publicly available large-scale fake news dataset, has been constructed.}
Compared to expert-based fact-checking, crowd-sourced fact-checking is relatively difficult to manage, less credible and accurate due to the political bias of fact-checkers and their conflicting annotations, and has better (though insufficient) scalability. Hence, in crowd-sourced fact-checking, one often needs to (1) filter non-credible users and (2) resolve conflicting fact-checking results; both requirements become more critical as the number of fact-checkers grows. Nevertheless, crowd-sourcing platforms often allow fact-checkers to provide more detailed feedback (e.g., their sentiments or stances), which can be further explored in fake news studies.

\paragraph{$\blacktriangleright$ Crowd-sourced Fact-checking Websites} Unlike expert-based fact-checking, crowd-sourced fact-checking websites are still in early development. An example is \textit{Fiskkit},\footnote{\url{http://fiskkit.com/}} where users can upload articles, provide ratings for sentences within articles, and choose tags that best describe the articles. The given sources of articles help (i)~distinguish the types of content (e.g., news vs. non-news) and (ii) determine its credibility (Section \ref{sec::credibility} provides the details). The tags categorized into multiple dimensions allow one to study the patterns across fake and non-fake news articles (see Fig. \ref{fig::fiskkit} for an example).
While crowd-sourced fact-checking websites are not many, we believe more crowd-sourced platforms or tools will arise as major Web and social media websites start to realize their importance in identifying fake news (e.g., Google,\footnote{\url{https://blog.google/topics/journalism-news/labeling-fact-check-articles-google-news/}} Facebook,\footnote{\url{https://newsroom.fb.com/news/2016/12/news-feed-fyi-addressing-hoaxes-and-fake-news/}} Twitter,\footnote{\url{https://blog.twitter.com/2010/trust-and-safety}} and Sina Weibo\footnote{\url{http://service.account.weibo.com/} (sign in required)}).

\subsection{Automatic Fact-checking} 
\label{subsec::automatic}

Manual fact-checking does not scale with the volume of newly created information, especially on social media. To address scalability, automatic fact-checking techniques have been developed, heavily relying on Information Retrieval (IR), Natural Language Processing (NLP), and Machine Learning (ML) techniques, as well as on network/graph theory~\cite{cohen2011computational}. {To review these techniques, a unified standard representation of knowledge is first presented that can be automatically processed by machines and has been widely adopted in related studies~\cite{nickel2016review}:}

\begin{myDef}[Knowledge]
\label{def::knowledge}
A set of \texttt{(\textbf{S}ubject, \textbf{P}redicate, \textbf{O}bject)} (\textbf{SPO}) triples extracted from the given information that well-represent the given information.
\end{myDef}

\noindent For instance, the knowledge within sentence ``Donald Trump is the president of the U.S.'' can be (\texttt{DonaldTrump}, \texttt{Profession}, \texttt{President}). 
{ Based on the above representation of knowledge, we present the following widely-accepted definitions for the key terms that often appear in automatic fact-checking  literature for a better understanding~\cite{dong2014knowledge,ciampaglia2015computational,shi2016discriminative}:}\vspace{-2mm}

\begin{myDef}[Fact]
\label{def::fact}
A fact is a knowledge (SPO triple) verified as truth.
\end{myDef}

\begin{myDef}[Knowledge Base]
\label{def::kb}
Knowledge Base (KB) is a set of facts.
\end{myDef}

\begin{myDef}[Knowledge Graph]
\label{def::kg}
Knowledge Graph (KG) is a graph structure representing the SPO triples in a knowledge base, where the entities (i.e., subjects or objects in SPO triples) are represented as nodes and relationships (i.e., predicates in SPO triples) are represented as edges.
\end{myDef}

\begin{figure}[t]
\centering
\includegraphics[width=\textwidth]{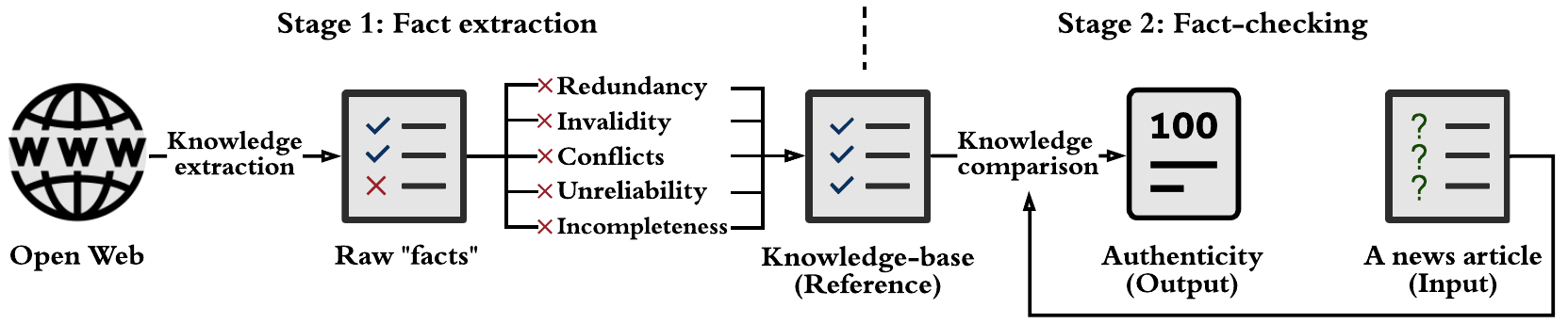}
\vspace{-5mm}
\caption{Automatic News Fact-checking Process}
\label{fig::factChecking}
\vspace{-2mm}
\end{figure}

{As a systematic approach for automatic fact-checking of news has never been presented before, here we prioritize organizing the related research to clearly present the automatic news fact-checking process over presenting each related study in detail.}
The automatic fact-checking process is shown in Fig. \ref{fig::factChecking}. It can be divided into two stages: fact extraction (a.k.a. \textit{knowledge-base construction}, see Section \ref{subsubsec:fact_extraction}) and fact-checking (a.k.a. \textit{knowledge comparison}, see Section \ref{subsubsec:knowledge_comparison}). 

\subsubsection{Fact Extraction}
\label{subsubsec:fact_extraction}
To collect facts and construct a KB (KG), knowledge is first extracted from the open Web as raw ``facts'' that need further processing. Such process often refers to knowledge extraction or \textit{relation extraction}~\cite{nickel2016review}.
Knowledge extraction can be classified into \textit{single-source} or \textit{open-source} knowledge extraction. Single-source knowledge extraction, which relies on one comparatively reliable source (e.g., Wikipedia) to extract knowledge, is relatively efficient but often leads to incomplete knowledge (see related studies in \cite{suchanek2007yago,auer2007dbpedia,bollacker2008freebase}). Open-source knowledge extraction aims to fuse knowledge from distinct sources; hence, it is less efficient that single-source knowledge extraction, but leads to more complete knowledge (see related studies in \cite{dong2014knowledge,nakashole2012patty,carlson2010toward,niu2012elementary}). More recent studies in relation extraction can be seen in \cite{di2019relation,lin2019learning,ijcai2019-750}. Finally, to form a KB (KG) from these extracted raw ``facts'', they  need to be further cleaned-up and completed by addressing the following issues:
\begin{itemize}
    \item \textit{Redundancy}. For example, \texttt{(DonaldJohnTrump, profession, President)} is redundant when having \texttt{(Donald- Trump, profession, President)} as \texttt{DonaldTrump} and \texttt{DonaldJohnTrump} match the same entity. The task to reduce redundancy is often referred to as \textit{entity resolution}~\cite{getoor2012entity}, a.k.a. \textit{deduplication} or \textit{record linkage}~\cite{nickel2016review} (see related studies in, e.g., \cite{altowim2014progressive,christen2008automatic,steorts2016bayesian,bhattacharya2007collective,whang2012joint});
    \item \textit{Invalidity}. The correctness of some facts depends on a specific time interval, for example, \texttt{(Britain, joinIn, EuropeanUnion)} has been outdated and should be updated. One way to address this issue is to allow facts to have beginning and ending dates~\cite{bollacker2008freebase}; or one can \textit{reify} current facts by adding extra assertions to them~\cite{hoffart2013yago2};
    \item \textit{Conflicts}. For example,  \texttt{(DonaldTrump, bornIn, NewYorkCity)} and \texttt{(DonaldTrump, bornIn, LosAngeles)} are a pair with conflicting knowledge. {Conflicts can be resolved by Multi-Criteria Decision-Making (MCDM) methods~\cite{deng2015generalized,MaximumDengetnropy,viviani2016multi,pasi2019multi};
    \item \textit{Unreliability (Low-credibility)}. For example, the knowledge extracted from The Onion,\footnote{\url{https://www.theonion.com/}} a satire news organization, is unreliable knowledge. Unreliable knowledge can be reduced by filtering low-credibility website(s) from which the knowledge is extracted. The credibility of website(s) can be obtained from resources such as NewsGuard\footnote{\url{http://www.newsguardtech.com/} \label{note:newsguard}} (expert-based), or systems such as MediaRank~\cite{ye2019mediarank}.} Section \ref{sec::credibility} provides more details; and
    \item \textit{Incompleteness}. Raw facts extracted from online resources, particularly, using a single source, are far from complete. Hence, reliably inferring new facts based on existing facts, a.k.a. \textit{KG completion}, is necessary to improve the KG being built. KG completion performs link prediction between entities, where the methods can be classified into three groups based on their assumptions: 
(1)~\textit{latent feature models}, that assume the existence of KB triples is conditionally independent given latent features and parameters (e.g., 
    RESCAL~\cite{nickel2012factorizing},
    NTN~\cite{socher2013reasoning},
    DistMult~\cite{yang2014embedding},
    TransE~\cite{bordes2013translating}, TransH~\cite{wang2014knowledge}, TransR~\cite{lin2015learning},
    ComplEx~\cite{trouillon2016complex},
    {SimplE~\cite{kazemi2018simple},
    and ConMask~\cite{shi2018open}});
(2)~\textit{graph feature models} that assume the existence of triples is conditionally independent given observed graph features and parameters (e.g., Path Ranking Algorithm (PRA)~\cite{lao2010relational}); and 
(3)~\textit{Markov Random Field (MRF) models}, that assume existing triples have local interactions~\cite{nickel2016review}.
\end{itemize}

Note that instead of building a KB (KG) from scratch, one can rely on existing large-scale ones, e.g., YAGO~\cite{suchanek2007yago,hoffart2013yago2}, Freebase~\cite{bollacker2008freebase}, NELL~\cite{carlson2010toward}, PATTY~\cite{nakashole2012patty}, DBpedia~\cite{auer2007dbpedia}, Elementary/DeepDive~\cite{niu2012elementary}, and Knowledge Vault \cite{dong2014knowledge}.

\subsubsection{Fact-checking}
\label{subsubsec:knowledge_comparison}

To assess the authenticity of news articles, we need to compare the knowledge extracted from to-be-verified news content (i.e., SPO triples) with the facts (i.e., true knowledge). KBs (KGs) are suitable resources for providing ground truth for news fact-checking, i.e., we can reasonably assume that the existing triples in a KB (KG) represent facts. However, for non-existing triples, their authenticity relies on assumptions made -- we list three common assumptions below -- and we may need further inference: 
\begin{itemize}
\item \emph{Closed-world assumption}: non-existing triples indicate false knowledge;
\item \emph{Open-world assumption}: non-existing triples indicate unknown knowledge that can be either true or false; and
\item \emph{Local closed-world assumption~\cite{dong2014knowledge}}: the authenticity of non-existing triples can be determined by the following rule: let $T(s,p)$ denote the set of existing triples in the KB (KG) for subject $s$ and predicate $p$. For any $(s,p,o) \notin T(s,p)$, if $|T(s,p)|>0$, we say the triple is false; if $|T(s,p)|=0$, its authenticity is unknown. 
\end{itemize}

Generally, the fact-checking strategy for a (\texttt{Subject}, \texttt{Predicate}, \texttt{Object}) triple is to
evaluate the \textit{possibility} that the edge labeled \texttt{Predicate} exists from the node labeled \texttt{Subject} to the node representing \texttt{Object} in a KG. Specifically, 
\begin{enumerate}
\item[Step 1:] \textit{Entity locating}.  \texttt{Subject} (Similarly, \texttt{Object}) is first matched with a node in the KG that represents the same entity as the \texttt{Subject} (similarly, \texttt{Object}), where \textit{entity resolution} techniques (e.g., \cite{altowim2014progressive,bhattacharya2007collective,trivedi2018linknbed}) can be used to identify proper matchings.
\item[Step 2:] \textit{Relation verification}. Triple (\texttt{Subject}, \texttt{Predicate}, \texttt{Object}) is considered as truth if an edge labeled \texttt{Predicate} from the node representing \texttt{Subject} to the one representing \texttt{Object} exists in the KG. Otherwise, its authenticity is (1) false based on the aforementioned closed-world assumption, or (2) determined after \textit{knowledge inference}.
\item[Step 3:] \textit{Knowledge inference}. When the triple (\texttt{Subject}, \texttt{Predicate}, \texttt{Object}) does not exist in the KG, the probability for the edge labeled \texttt{Predicate} to exist from the node representing \texttt{Subject} to the one representing \texttt{Object} in the KG can be computed, e.g., using \textit{link prediction} methods such as semantic proximity~\cite{ciampaglia2015computational}, discriminative predicate path~\cite{shi2016discriminative}, or LinkNBed~\cite{trivedi2018linknbed}.
\end{enumerate}

Finally, we conclude this section by formally defining the problem of automatic news fact-checking in Problem \ref{pro::factChecking}, and discussing the potential research avenues in automatic news fact-checking in Section \ref{subsubsec:knowledge_discussion}.

\begin{problemDef}[Fact-checking]
\label{pro::factChecking}
Assume a to-be-verified news article is represented as a set of knowledge statements, i.e., SPO triples $(s_i,p_i,o_i), i=1,2,\dots,n$. Let $G_\mathit{KB}$ refer to a knowledge graph containing a set of facts (i.e., true knowledge) denoted by $(s_{t_j},p_{t_j},o_{t_j}), j=1,2,\dots,m$. News fact-checking is to identify a function $\mathcal{F}$ that assigns an authenticity value $A_i \in [0,1]$ to each corresponding $(s_i, p_i, o_i)$ by comparing it with every $(s_{t_j},p_{t_j},o_{t_j})$ in the knowledge graph, where $A_i = 1$ ($A_i = 0$) indicates the triple is true (false). The final authenticity index $A \in [0, 1]$ of the to-be-verified news article is obtained by aggregating all $A_i$'s. To summarize,
\begin{equation}
\begin{matrix}
\mathcal{F}: (s_i,p_i,o_i)  \xrightarrow{G_{\mathit{KB}}} A_i,\\ 
A = \mathcal{I}(A_1, A_2, \cdots, A_n),
\end{matrix}
\end{equation}
where $\mathcal{I}$ is an aggregation function of choice ({ e.g., weighted or arithmetic average}). The to-be-verified news article is true if $A=1$, and is [completely] false if $A=0$.
Specifically, function $\mathcal{F}$ can be formulated as
\begin{equation}
\mathcal{F}((s_i,p_i,o_i),G_{\mathit{KB}}) = P(\mbox {edge labeled } p_i \mbox{ linking } s'_i \mbox{ to } o'_i \mbox{ in }G_{\mathit{KB}}),
\end{equation}
where $P(\cdot)$ denotes the probability, and $s'_i$ and $o'_i$ are the matched entities to $s_i$ and $o_i$ in $G_{\mathit{KB}}$, respectively:
{ 
\begin{eqnarray} \label{eq:distance}
s'_i = \underset{s_{t_j}}{\arg\min}~ |\mathcal{D}(s_i,s_{t_j})|<\theta,  &
o'_i = \underset{o_{t_j}}{\arg\min}~|\mathcal{D}(o_i,o_{t_j})|<\theta.
\end{eqnarray}

In Eq. (\ref{eq:distance}), $\mathcal{D}(a,b)$ measures the distance  between entity $a$ and $b$. Such distance can be computed by, e.g., Jaccard distance directly, or by cosine distance after entity embedding. When the distance between $a$ and $b$ is zero (i.e., $\mathcal{D}(a,b)=0$), or the distance is less than a certain threshold $\theta$ (i.e., $|\mathcal{D}(a,b)|<\theta$), one can regard $a$ as the same entity as $b$.}
\end{problemDef}
 
\subsubsection{Discussion}
\label{subsubsec:knowledge_discussion}

We have detailed fact extraction (i.e., KB/KG construction) and fact-checking (i.e., knowledge comparison), the two main components of automatic news fact-checking. There are some open issues and several potential research tasks. First, when collecting facts to construct KB (KG), one concern is the source(s) from which facts are extracted. In addition to the traditional sources such as Wikipedia, some other sources, e.g., fact-checking websites that contain \textit{expert analysis and justifications} for checked news content, might help provide high-quality domain knowledge. However, such sources have rarely been considered in current research. Second, we highlight the value of research on \textit{dynamic KBs (KGs)} for news fact-checking that can automatically remove invalid knowledge and introduce new facts. Such properties are especially important due to news timeliness -- news articles are often not about ``common knowledge'', but around recent events. Third, it has been verified that fake news spreads faster than true news~\cite{vosoughi2018spread}, which attaches great importance to fast news fact-checking to achieve \textit{fake news early detection} (see Section \ref{sec::discussion} for a summary and discussion). Current research in building KBs (KGs) has focused on constructing KBs (KGs) with as many facts as possible. However, fast news fact-checking requires not only identifying parts of the to-be-verified news that is check-worthy (see Section \ref{sec::discussion} for a discussion on identifying check-worthy content), but also a KB (KG) that only stores as many ``valuable'' facts as possible (i.e., a \textit{KB (KG) simplification process}). 

\section{Style-based Fake News Detection}
\label{sec::style} 

Similar to knowledge-based fake news detection (Section \ref{sec::knowledge}), style-based fake news detection also focuses on analyzing the news content. However, knowledge-based methods mainly evaluate the authenticity of the given news, while style-based methods can \underline{assess news intention}, i.e., is there an intention to mislead the public or not? The intuition and assumption behind style-based methods is that malicious entities prefer to write fake news in a ``special'' style to encourage others to read and convince them to trust. Before discussing how such ``special'' content styles can be automatically identified, we first define fake news style in a way that facilitates use of machine learning:

\begin{myDef}[Fake News Style]
\label{def::style}
A set of quantifiable characteristics (e.g., machine learning features) that can well represent fake news content and differentiate it from true news content.
\end{myDef}

Based on this definition for fake news style, style-based fake news detection is often formulated as a binary (or at times, a multi-label) classification problem:

\begin{problemDef}[Style-based Fake News Detection] \label{pro::style}
Assume a to-be-verified news article $\mathcal{N}$ can be represented as a set of $k$ content features denoted by feature vector $\mathbf{f} \in \mathbb{R}^k$. The task to verify the news article based on its content style is to identify a function $\mathcal{S}$, such that 
\begin{equation}
\begin{matrix}
\mathcal{S}: \mathbf{f} \xrightarrow{TD} \hat{y}
\end{matrix}
\end{equation}
where $\hat{y} \in \{0~\text{(\texttt{true})},1~\text{(\texttt{fake})}\}$ is the predicted news label and $TD= \{ (\mathbf{f}_{l},y_{l}) \mid \mathbf{f}_{l} \in \mathbb{R}^k, y_{l} \in \{0,1\}, l=1\dots n \}$ is the training dataset. The training dataset helps estimate the parameters within $\mathcal{S}$ and consists of a set of $n$ news articles represented by the same set of features ($\mathbf{f}_{l}$) with known news labels ($y_{l}$).
\end{problemDef}

Hence, the performance of style-based fake news detection methods rely on 
(I)~how well the style of news content (text and images) can be captured and represented (see Section \ref{subsec:style_representation}); and
(II)~how the classifier (model) is performing based on different news content representations (see Section \ref{subsec:style_classification}). In addition, we summarize
(III)~some verified patterns of fake news content style in Section \ref{subsec:style_pattern}, and provide (IV) our discussion on style-based methods in Section \ref{subsec:style_discussion}.

\subsection{Style Representation}
\label{subsec:style_representation}

As provided in Definition \ref{def::style}, the content style is commonly represented by a set of quantifiable characteristics, often machine learning features. Generally, these features can be grouped into \textit{textual features} (see Section \ref{subsubsec:text_representation}) and \textit{visual features} (see Section \ref{subsubsec:image_representation}), representing news text and images, respectively.

\subsubsection{News Text}
\label{subsubsec:text_representation}

Broadly speaking, textual features can be grouped into (I) \textit{general features} and (II) \textit{latent features}.

\begin{figure}[t]
    \begin{minipage}{0.475\textwidth}
    \centering
    \includegraphics[width=\textwidth]{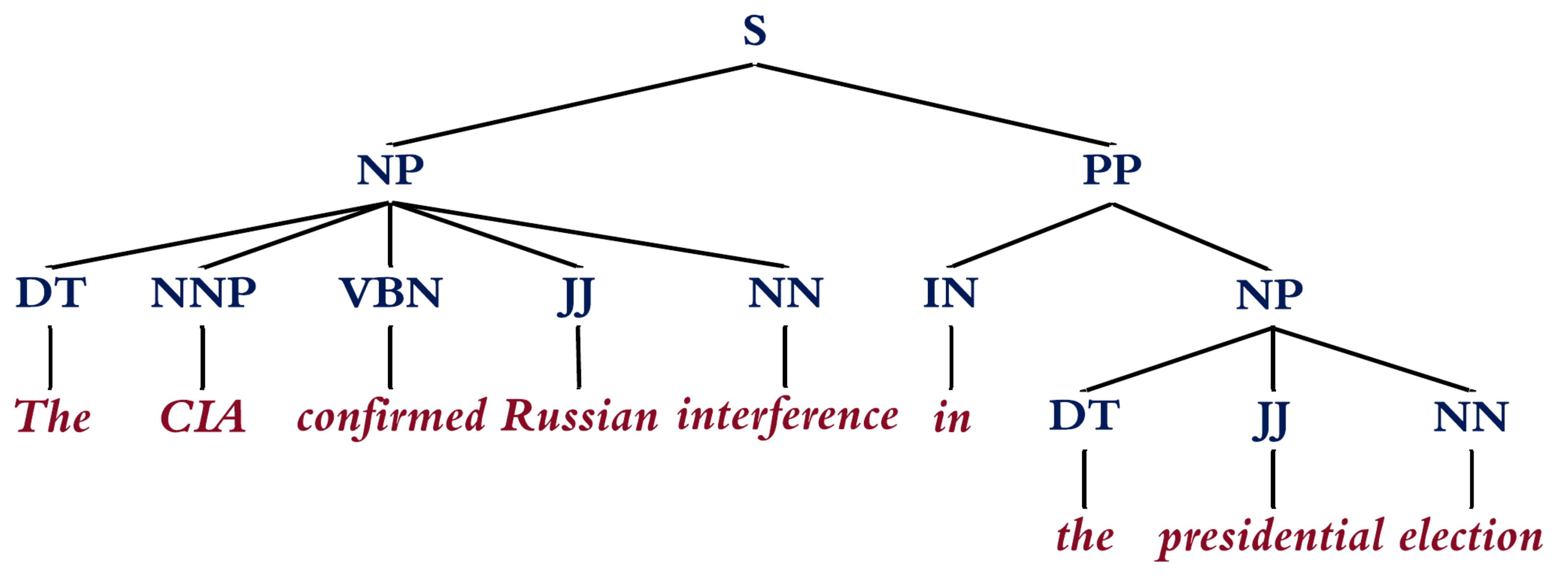}
    \caption{ \footnotesize{PCFG Parse Tree
    for the sentence ``The CIA confirmed Russian interference in the presidential election'' in a fake news article (directly from \cite{zhou2019content}). 
    The lexicalized rewrite rules of this sentence are: 
    S$\rightarrow$NP PP, 
    NP$\rightarrow$DT NNP VBN JJ NN, 
    PP$\rightarrow$IN NP, 
    NP$\rightarrow$DT JJ NN, 
    DT$\rightarrow$``the'', 
    NNP$\rightarrow$``CIA'', 
    VBN$\rightarrow$``confirmed'', 
    JJ$\rightarrow$``Russian'', 
    NN$\rightarrow$``interference'', 
    IN$\rightarrow$``in'', 
    JJ$\rightarrow$``presidential'', and NN$\rightarrow$``election''.}
    }
    \label{fig:cfg}
    \end{minipage} \quad
    \begin{minipage}{0.485\textwidth}
    \centering
    \includegraphics[width=\textwidth]{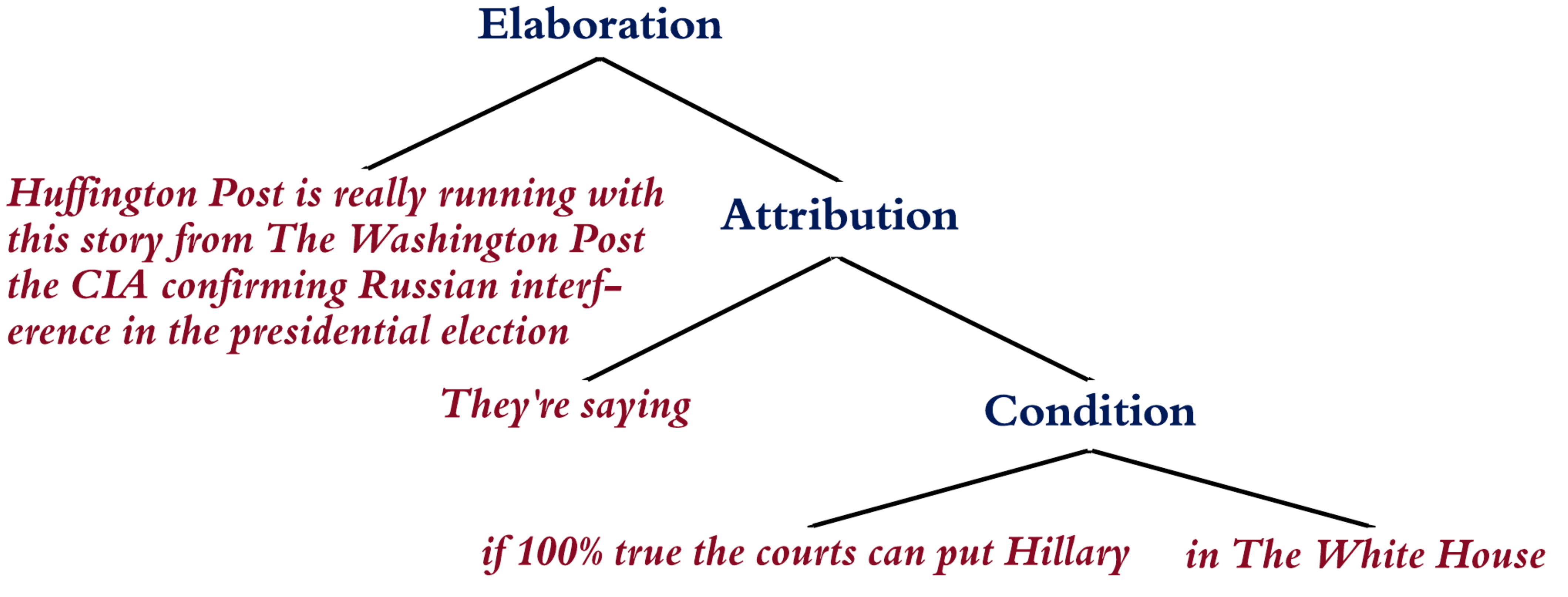}
    \caption{ \footnotesize{Rhetorical Structure
    for the partial content ``Huffington Post is really running with this story from The Washington Post about the CIA confirming Russian interference in the presidential election. They're saying if 100\% true, the courts can PUT HILLARY IN THE WHITE HOUSE!'' in a fake news article (directly from \cite{zhou2019content}). 
    Here, one elaboration, one attribution, and one condition rhetorical relationship exist. }
    }
    \label{fig:rst}
    \end{minipage}\vspace{-4mm}
\end{figure}

\paragraph{$\blacktriangleright$ General textual features} 
General textual features are often used to detect fake news within a traditional machine learning framework (detailed in Section \ref{subsubsec:mlModels}). These features describe content style from (at least) four language levels: (i) \textit{lexicon}, (ii) \textit{syntax}, (iii) \textit{discourse} and (iv) \textit{semantic}~\cite{conroy2015automatic}. 
The main task at the lexicon level is to assess the frequency statistics of lexicons, which can be basically conducted  using a Bag-Of-Word (BOW) model~\cite{zhou2019content}. At the syntax level, shallow syntax tasks are performed by Part-Of-Speech (POS)-taggers to assess POS (e.g., nouns and verbs) frequencies~\cite{zhou2019content,feng2012syntactic}. Deep syntax tasks are performed by Probabilistic Context-Free Grammar (PCFG) parse trees (see Fig. \ref{fig:cfg} for an example) that enable assessing the frequencies of rewrite rules (i.e., productions)~\cite{zhou2019content,feng2012syntactic,perez2017automatic}. { Four different encodings of rewrite rules can be considered for a PCFG parse tree~\cite{feng2012syntactic}:
\begin{itemize}
    \item $r$: unlexicalized rewrite rules, i.e., all rewrite rules except for those with leaf nodes such as IN$\rightarrow$``in'';
    \item $r*$: lexicalized rewrite rules, i.e., all rewrite rules;
    \item $\hat{r}$: unlexicalized rewrite rules with grandparent nodes, e.g., PP\^{}S$\rightarrow$IN NP; and
    \item $\hat{r}*$: lexicalized rewrite rules with grandparent nodes, e.g., IN\^{}PP$\rightarrow$``in''.
\end{itemize}}
At the discourse level, Rhetorical Structure Theory (RST) and rhetorical parsing tools can be used to capture the frequencies of rhetorical relations among sentences as features~\cite{zhou2019content,karimi2019learning} (see Fig. \ref{fig:rst} for an illustration).
Finally, at a semantic level, such frequencies can be assigned to lexicons or phrases that fall into each psycho-linguistic category (e.g., those defined in Linguistic Inquiry and Word Count (LIWC)~\cite{perez2017automatic}), or that fall into each self-defined psycho-linguistic attribute. {These attributes can be derived 
from experience, or be inspired by related deception theories (see news-related theories in Table \ref{tab::theories} or \cite{zhou2019content} as a typical interdisciplinary fake news study). 
Based on our investigation, such attributes and their corresponding computational features can be grouped along ten dimensions: \textit{quantity}~\cite{mccornack2014information}, \textit{complexity}, \textit{uncertainty}, \textit{subjectivity}~\cite{undeutsch1967beurteilung}, \textit{non-immediacy}, \textit{sentiment}~\cite{zuckerman1981verbal}, \textit{diversity}~\cite{undeutsch1967beurteilung}, \textit{informality}~\cite{undeutsch1967beurteilung}, \textit{specificity}~\cite{johnson1981reality}, and \textit{readability} (see Table \ref{tab::linguisticCues}), which are initially developed for identifying deception in computer-mediated communications~\cite{zhou2004automating,fuller2009decision} and testimonies~\cite{afroz2012detecting}, and recently used in fake news detection~\cite{bond2017lyin,potthast2017stylometric,perez2017automatic,zhou2019content}.}

\begin{table}[htbp]
\centering
\caption{Semantic-level Features in News Content}
\label{tab::linguisticCues}
\begin{adjustbox}{max width=\textwidth}
\small
\begin{threeparttable}
\begin{tabular}{|l|l|l|l|l|l|l|g|g|g|g|}
\multicolumn{1}{l}{\textbf{Attribute Type}}                     & \multicolumn{1}{l}{\textbf{Feature}}    
& \rot{\cite{zhou2004automating}}  
& \rot{\cite{fuller2009decision}}
& \rot{\cite{afroz2012detecting}}
& \rot{\cite{siering2016detecting}}
& \rot{\cite{zhang2016online}}
& \rot{\cite{bond2017lyin}}
& \rot{\cite{potthast2017stylometric}}
& \rot{\cite{perez2017automatic}}
& \rot{\cite{zhou2019content}}
\\ \hline \hline
\multirow{5}{*}{\textbf{Quantity}}     
    & \# Characters                                            
    &              &              & $\checkmark$ &              &              &            &              
    &              & $\checkmark$  \\ \cline{2-10} 
    & \# Words                                                 
    & $\checkmark$ & $\checkmark$ & $\checkmark$ & $\checkmark$ & $\checkmark$ &            &              
    &              & $\checkmark$  \\ \cline{2-10} 
    & \# Noun phrases                                     
    & $\checkmark$ &              &              &              &              &            &              
    &              &               \\ \cline{2-10} 
    & \# Sentences                                             
    & $\checkmark$ & $\checkmark$ & $\checkmark$ & $\checkmark$ &              &            &              
    &              & $\checkmark$  \\ \cline{2-10} 
    & \# Paragraphs                                            
    &              &              &              &              &              &            & $\checkmark$ 
    &              & $\checkmark$  \\ \hlineB{2.5} 
\multirow{4}{*}{\textbf{Complexity}}   
    & Average \# characters per word                      
    & $\checkmark$ & $\checkmark$ & $\checkmark$ & $\checkmark$ &              &            &              
    &              & $\checkmark$  \\ \cline{2-10}
    & Average \# words per sentence                       
    & $\checkmark$ & $\checkmark$ & $\checkmark$ & $\checkmark$ & $\checkmark$ &            &  
    &              & $\checkmark$  \\ \cline{2-10} 
    & Average \# clauses per sentence                     
    & $\checkmark$ &              &              & $\checkmark$ &              &            &              
    &              &               \\ \cline{2-10}
    & Average \# punctuations per sentence                
    & $\checkmark$ & $\checkmark$ & $\checkmark$ & $\checkmark$ &              &            &              
    &              &               \\ \hlineB{2.5}
\multirow{6}{*}{\textbf{Uncertainty}}  
    & \#/\% Modal verbs (e.g., ``shall'')                                 
    & $\checkmark$ & $\checkmark$ & $\checkmark$ & $\checkmark$ &              &              &             
    &              &               \\ \cline{2-10} 
    & \#/\% Certainty terms (e.g., ``never'' and ``always'')                             
    & $\checkmark$ & $\checkmark$ & $\checkmark$ & $\checkmark$ &              & $\checkmark$ &             
    &              & $\checkmark$  \\ \cline{2-10} 
    & \#/\% Generalizing terms (e.g., ``generally'' and ``all'')                          
    &              & $\checkmark$ &              &              &              &              &             
    &              &               \\ \cline{2-10} 
    & \#/\% Tentative terms (e.g., ``probably'')                             
    &              & $\checkmark$ & $\checkmark$ &              &              & $\checkmark$ &              
    &              & $\checkmark$  \\ \cline{2-10}
    & \#/\% Numbers and quantifiers                      
    &              &              & $\checkmark$ &              &              &              &              
    &              &               \\ \cline{2-10}
    & \#/\% Question marks                                   
    &              &              & $\checkmark$ &              &              &              &              
    &              & $\checkmark$  \\ \hlineB{2.5}
\multirow{4}{*}{\textbf{Subjectivity}} 
    & \#/\% Biased lexicons (e.g., ``attack'')                                
    &              &              &              &              &              &            &              
    &              & $\checkmark$  \\ \cline{2-10}
    & \#/\% Subjective verbs (e.g., ``feel'' and ``believe'')                            
    & $\checkmark$ &              &              &              & $\checkmark$ &            &              
    &              &               \\ \cline{2-10} 
    & \#/\% Report verbs (e.g., ``announce'')                                
    &              &              &              &              &              &            &              
    &              & $\checkmark$  \\ \cline{2-10} 
    & \#/\% Factive verbs (e.g., ``observe'')                               
    &              &              &              &              &              &            &              
    &              & $\checkmark$  \\ \hlineB{2.5}
\multirow{5}{*}{\textbf{Non-immediacy}} 
    & \#/\% Passive voice                                
    & $\checkmark$ & $\checkmark$ &              &              &              &            &              
    &              &               \\ \cline{2-10} 
    & \#/\% Self reference: \nth{1} person singular pronouns          
    & $\checkmark$ & $\checkmark$ & $\checkmark$  & $\checkmark$ & $\checkmark$ &           &              
    &              &               \\ \cline{2-10} 
    & \#/\% Group reference: \nth{1} person plural pronouns           
    & $\checkmark$ & $\checkmark$ & $\checkmark$  & $\checkmark$ & $\checkmark$ &           &              
    &              &               \\ \cline{2-10} 
    & \#/\% Other reference: \nth{2} and \nth{3} person pronouns       
    & $\checkmark$ & $\checkmark$ & $\checkmark$ & $\checkmark$ & $\checkmark$ &            &              
    &              &               \\ \cline{2-10} 
    & \#/\% Quotations                                       
    &              &              & $\checkmark$ &              &              &            & $\checkmark$ 
    &              &               \\\hlineB{2.5}
\multirow{6}{*}{\textbf{Sentiment}}    
    & \#/\% Positive words                               
    & $\checkmark$ & $\checkmark$ & $\checkmark$ & $\checkmark$ & $\checkmark$ & $\checkmark$ & 
    &              & $\checkmark$  \\ \cline{2-10} 
    & \#/\% Negative words                               
    & $\checkmark$ & $\checkmark$ & $\checkmark$ & $\checkmark$ & $\checkmark$ & $\checkmark$ & 
    &              & $\checkmark$  \\ \cline{2-10} 
    & \#/\% Anxiety/angry/sadness words                               
    &              &              &              &              &              & $\checkmark$ & 
    &              & $\checkmark$  \\ \cline{2-10}
    & \#/\% Exclamation marks                                
    &              &              & $\checkmark$ &              &              &              &              
    &              & $\checkmark$  \\ \cline{2-10} 
    & Content sentiment polarity                               
    &              &              &              &              &              &              & 
    &              & $\checkmark$  \\ \hlineB{2.5}    
\multirow{3}{*}{\textbf{Diversity}}    
    & Lexical diversity: \#/\% unique words or terms              
    & $\checkmark$ & $\checkmark$ & $\checkmark$ & $\checkmark$ & $\checkmark$ &              &               
    &              & $\checkmark$   \\ \cline{2-10} 
    & Content word diversity: \#/\% unique content words          
    & $\checkmark$ & $\checkmark$ &              &              & $\checkmark$ &              &              
    &              & $\checkmark$   \\ \cline{2-10} 
    & Redundancy: \#/\% unique function words                    
    & $\checkmark$ & $\checkmark$ & $\checkmark$ &              & $\checkmark$ &              &              
    &              & $\checkmark$   \\ \cline{2-10}
    & \#/\% Unique nouns/verbs/adjectives/adverbs                                
    &              &              &              &              &              &              & 
    &              & $\checkmark$   \\ \hlineB{2.5}   
\multirow{2}{*}{\textbf{Informality}}                   
    & \#/\% Typos (misspelled words)          
    & $\checkmark$ &              &              & $\checkmark$ & $\checkmark$ &              &                
    &              &               \\ \cline{2-10} 
    & \#/\% Swear words/netspeak/assent/nonfluencies/fillers                               
    &              &              &              &              &              &              & 
    &              & $\checkmark$  \\ \hlineB{2.5}   
\multirow{4}{*}{\textbf{Specificity}}  
    & Temporal/spatial ratio                                             
    & $\checkmark$ & $\checkmark$ &              &              &              & $\checkmark$ &               
    &              &               \\ \cline{2-10} 
    & Sensory ratio                                              
    & $\checkmark$ & $\checkmark$ &              & $\checkmark$ &              & $\checkmark$ &                
    &              & $\checkmark$  \\ \cline{2-10} 
    & Causation terms                                            
    &              & $\checkmark$ &              &              &              & $\checkmark$ &               
    &              & $\checkmark$  \\ \cline{2-10} 
    & Exclusive terms                                            
    &              & $\checkmark$ &              &              &              &            &               
    &              &               \\ \hlineB{2.5}
\multicolumn{2}{|l|}{\textbf{Readablity} (e.g., Flesch-Kincaid and Gunning-Fog index)}     
    &              &              & $\checkmark$ &              &              &            & $\checkmark$  
    & $\checkmark$ & $\checkmark$  \\ \hline
\end{tabular}
\begin{tablenotes}
\item The studies labeled with gray background color investigate news articles.
\end{tablenotes}
\end{threeparttable}
\end{adjustbox}
\end{table}

It should be noted that ``frequency'' can be defined and computed in three ways. Assume a corpus $\mathcal{C}$ contains $p$ news articles $\mathcal{C}=\{\mathcal{A}_1,\mathcal{A}_2,\cdots,\mathcal{A}_p\}$ and a total of $q$ words $W=\{w_1,w_2,\cdots,w_q\}$ (POS tags, rewrite rules, rhetorical relationships, etc.). $x^i_j$ denotes the number of $w_j$ appearing in $\mathcal{A}_i$. Then the ``frequency'' of $w_j$ for news $\mathcal{A}_i$ can be:
\begin{itemize}
    \item \textit{Absolute frequency} $f_a$, i.e., $f_a=x^i_j$;
    \item \textit{Standardized frequency} $f_s$, which removes the impact of content length, i.e., $f_s=\frac{x^i_j}{\sum_{j}x^i_j}$~\cite{zhou2019content}; or
    \item \textit{Relative frequency} $f_r$ by using Term Frequency-Inverse Document Frequency (TF-IDF), which further compares such frequency with that in other news articles in the corpus, i.e., $f_r=\frac{x^i_j}{\sum_{j}x^i_j} \ln p \frac{1}{\sum_{i}x^i_j}$~\cite{perez2017automatic}.
\end{itemize} 
In general, TF-IDF can be applied at various language levels, so can $n$-gram models which enable capturing the sequence of words (POS tags, rewrite rules, etc.)~\cite{perez2017automatic,feng2012syntactic}.

\paragraph{$\blacktriangleright$ Latent textual features} 
Latent textual features are often used for news text embedding. Such an embedding can be conducted at the word-level~\cite{mikolov2013efficient,pennington2014glove}, sentence-level~\cite{arora2016simple,le2014distributed}, or document-level~\cite{le2014distributed}; results are vectors representing a news article and can be directly used as the input to classifiers (e.g., SVMs) when predicting fake news within a traditional machine learning framework~\cite{zhou2019content} (detailed in Section \ref{subsubsec:mlModels}). Such embeddings (often at the word-level) can be further incorporated into neural network architectures (e.g., Convolution Neural Networks,  CNNs~\cite{lecun1989backpropagation,krizhevsky2012imagenet,simonyan2014very,szegedy2015going,he2016deep,huang2017densely}, Recurrent Neural Networks, RNNs~\cite{hochreiter1997long,cho2014learning,schuster1997bidirectional}, and the 
Transformer~\cite{vaswani2017attention,devlin2018bert}) to predict fake news within a deep learning framework (see details in Section \ref{subsubsec:dlModels}), where CNNs represent news text from a local to global view, and RNNs and the Transformer capture the sequences with news text. Theoretically, such latent representation can also be obtained by matrix or tensor factorization; current style-based fake news detection studies have rarely considered them, while a few studies focusing on the propagation aspects of fake news have considered these methods, which we will review later in Section \ref{sec::propagation}.

\subsubsection{News Images}
\label{subsubsec:image_representation}

Currently, not many studies exist on detecting fake news by exploring news images~\cite{shu2017fake}. News images can be represented by hand-crafted (i.e., non-latent) features, e.g., the visual features defined in \cite{jin2017novel} (see Table \ref{tab:image_features}). On the other hand, to be further processed by neural networks such as VGG-16/19~\cite{simonyan2014very} to obtain a latent representation, each image is often embedded as a pixel matrix or tensor with size $width\times height\times \#channel(s)$, where channels can be the gray value (\#channels=1) or RGB data (\#channels=3).

\begin{table}[t]
\centering
\caption{Visual Features in News Content (defined in \cite{jin2017novel})}\vspace{-3mm}
\label{tab:image_features}
\begin{adjustbox}{max width=\textwidth}
\begin{tabular}{|l|l|}
\hline
\rowcolor{gray!20}\textbf{Feature} & \textbf{Description} \\ \hline \hline
\textbf{Visual Clarity Score} & Distribution difference between the image set of a news article and that in the corpus \\ \hline
\textbf{Visual Coherence Score} & Average similarities between pairs of images in a news article \\ \hline
\textbf{Visual Similarity Distribution Histogram} & Distribution histogram of the similarity matrix of images in a news article \\ \hline
\textbf{Visual Diversity Score} & Weighted average of dissimilarities between image pairs in a news article \\ \hline
\textbf{Visual Clustering Score} & The number of image clusters in a news article \\ \hline 
\end{tabular}
\end{adjustbox}
\end{table}\vspace{-3mm}

\subsection{Style Classification}
\label{subsec:style_classification}

We detail style-based fake news detection models that rely on traditional Machine Learning (ML) (in Section \ref{subsubsec:mlModels}) and on Deep Learning (DL)~\cite{lecun2015deep} (in Section \ref{subsubsec:dlModels}), along with their performance results.

\subsubsection{Traditional Machine Learning-based Models}
\label{subsubsec:mlModels}

Within a traditional ML framework, news content is represented by using a set of manually selected [latent and non-latent] features, which can be extracted from news images, or text at various language levels (lexicon-, syntax-, semantic-, and discourse-level)~\cite{zhou2019content,perez2017automatic,feng2012syntactic}. Machine learning models that can detect news type (e.g., true or fake) based on this representation can be supervised, semi-supervised, or unsupervised, where supervised methods (classifiers) have been mainly used for style-based fake news detection, e.g., style-based methods have relied on SVMs~\cite{perez2017automatic,feng2012syntactic}, Random Forests (RF)~\cite{zhou2019content}, and XGBoost~\cite{chen2016xgboost,zhou2019content}.

Within the same experimental setup as that in \cite{zhou2019content}, the performance of [latent and non-latent] features at various levels is presented and compared in Table \ref{tab:mlModel}. Results indicate that when predicting fake news using a traditional ML framework, 
(1)~non-latent features often outperform latent ones; 
(2)~combining features across levels can outperform using single-level features; and 
(3)~[standardized] frequencies of lexicons and rewrite rules better represent fake news content style and perform better (while more time-consuming to compute) than other feature groups.

It should be noted that as classifiers perform best for machine learning settings they were initially designed for, it is unjustified to determine algorithms that perform best for fake news detection (related discussions can be found in \cite{kotsiantis2007supervised,fernandez2014we}). 

\begin{table}[t]
\centering
\small
\caption{Feature Performance (accuracy \textbf{Acc.} and $\mathbf{F_1}$-score) in Fake News Detection using Traditional Machine Learning (\textbf{R}andom \textbf{F}orests and \textbf{XGBoost} classifiers)~\cite{zhou2019content}.
Results show that 
(1)~non-latent features can outperform latent ones; 
(2)~combining features across levels can outperform using single-level features; and 
(3)~the [standardized] frequencies of lexicons and rewrite rules better represent fake news content style and perform better (while more time-consuming to compute) than other feature groups.}\vspace{-3mm}
\label{tab:mlModel}
\begin{adjustbox}{max width=\textwidth}
\begin{tabular}{|c|c|l|l|l|l|l|l|l|l|l|}
\hline
\multicolumn{3}{|c|}{\multirow{3}{*}{\textbf{Feature Group}}} & \multicolumn{4}{c|}{\textbf{PolitiFact data}~\cite{shu2018fakenewsnet}} & \multicolumn{4}{c|}{\textbf{BuzzFeed data}~\cite{shu2018fakenewsnet}} \\ \cline{4-11} 
    \multicolumn{3}{|c|}{} & \multicolumn{2}{c|}{\textbf{XGBoost}} & \multicolumn{2}{c|}{\textbf{RF}} & \multicolumn{2}{c|}{\textbf{XGBoost}} & \multicolumn{2}{c|}{\textbf{RF}} \\ \cline{4-11} 
    \multicolumn{3}{|c|}{} & \multicolumn{1}{c|}{\textbf{Acc.}} & \multicolumn{1}{c|}{$\mathbf{F_1}$} & \multicolumn{1}{c|}{\textbf{Acc.}} & \multicolumn{1}{c|}{$\mathbf{F_1}$} & \multicolumn{1}{c|}{\textbf{Acc.}} & \multicolumn{1}{c|}{$\mathbf{F_1}$} & \multicolumn{1}{c|}{\textbf{Acc.}} & \multicolumn{1}{c|}{$\mathbf{F_1}$} \\ \hline \hline
\multirow{9}{*}{\rotatebox{90}{\textbf{Non-latent Features}}} &   \multirow{2}{*}{\textbf{Lexicon}} & BOWs ($f_s$) & 0.856 & 0.858 & 0.837 & 0.836 & 0.823 & 0.823 & 0.815 & 0.815 \\ \cline{3-11} 
    &   & Unigram+bigram ($f_r$) & 0.755 & 0.756 & 0.754 & 0.755 & 0.721 & 0.711 & 0.735 & 0.723 \\ \cline{2-11}
    &   \multirow{3}{*}{\textbf{Syntax}} & POS tags ($f_s$) & 0.755 & 0.755 & 0.776 & 0.776 & 0.745 & 0.745 & 0.732 & 0.732 \\ \cline{3-11} 
    &   & Rewrite rules ($r*, f_s$) & \textbf{0.877} & \textbf{0.877} & 0.836 & 0.836 & 0.778 & 0.778 & 0.845 & 0.845 \\ \cline{3-11} 
    &   & Rewrite rules ($r*, f_r$) & 0.749 & 0.753 & 0.743 & 0.748 & 0.735 & 0.738 & 0.732 & 0.735 \\ \cline{2-11} 
    &   \multirow{2}{*}{\textbf{Semantic}} & LIWC & 0.645 & 0.649 & 0.645 & 0.647 & 0.655 & 0.655 & 0.663 & 0.659 \\ \cline{3-11} 
    &   & Theory-driven~\cite{zhou2019content} & 0.745 & 0.748 & 0.737 & 0.737 & 0.722 & 0.750 & 0.789 & 0.789 \\ \cline{2-11} 
    &   \textbf{Discourse} & Rhetorical relationships & 0.621 & 0.621 & 0.633 & 0.633 & 0.658 & 0.658 & 0.665 & 0.665 \\ \clineB{2-11}{2}
    &   \textbf{Combination} & \cite{zhou2019content} & 0.865 & 0.865 & \textbf{0.845} & \textbf{0.845} & \textbf{0.855} & \textbf{0.856} & \textbf{0.854} & \textbf{0.854} \\ \hlineB{2}
\multicolumn{2}{|c|}{\multirow{2}{*}{\textbf{Latent Features}}} 
    & \textsc{Word2Vec}~\cite{mikolov2013efficient} & 0.688 & 0.671 & 0.663 & 0.667 & 0.703 & 0.714 & 0.722 & 0.718 \\ \cline{3-11}
    \multicolumn{2}{|c|}{} & \textsc{Doc2Vec}~\cite{le2014distributed} & 0.698 & 0.684 & 0.712 & 0.698 & 0.615 & 0.610 & 0.620 & 0.615 \\ \hline
\end{tabular}
\end{adjustbox}
\end{table}\vspace{-2mm}

\subsubsection{Deep Learning-based Models}
\label{subsubsec:dlModels}

\begin{figure}[t]
    \subfigure[\textsc{EANN} (directly from \cite{wang2018eann})]{\label{subfig:eann}
    \includegraphics[width=0.38\textwidth]{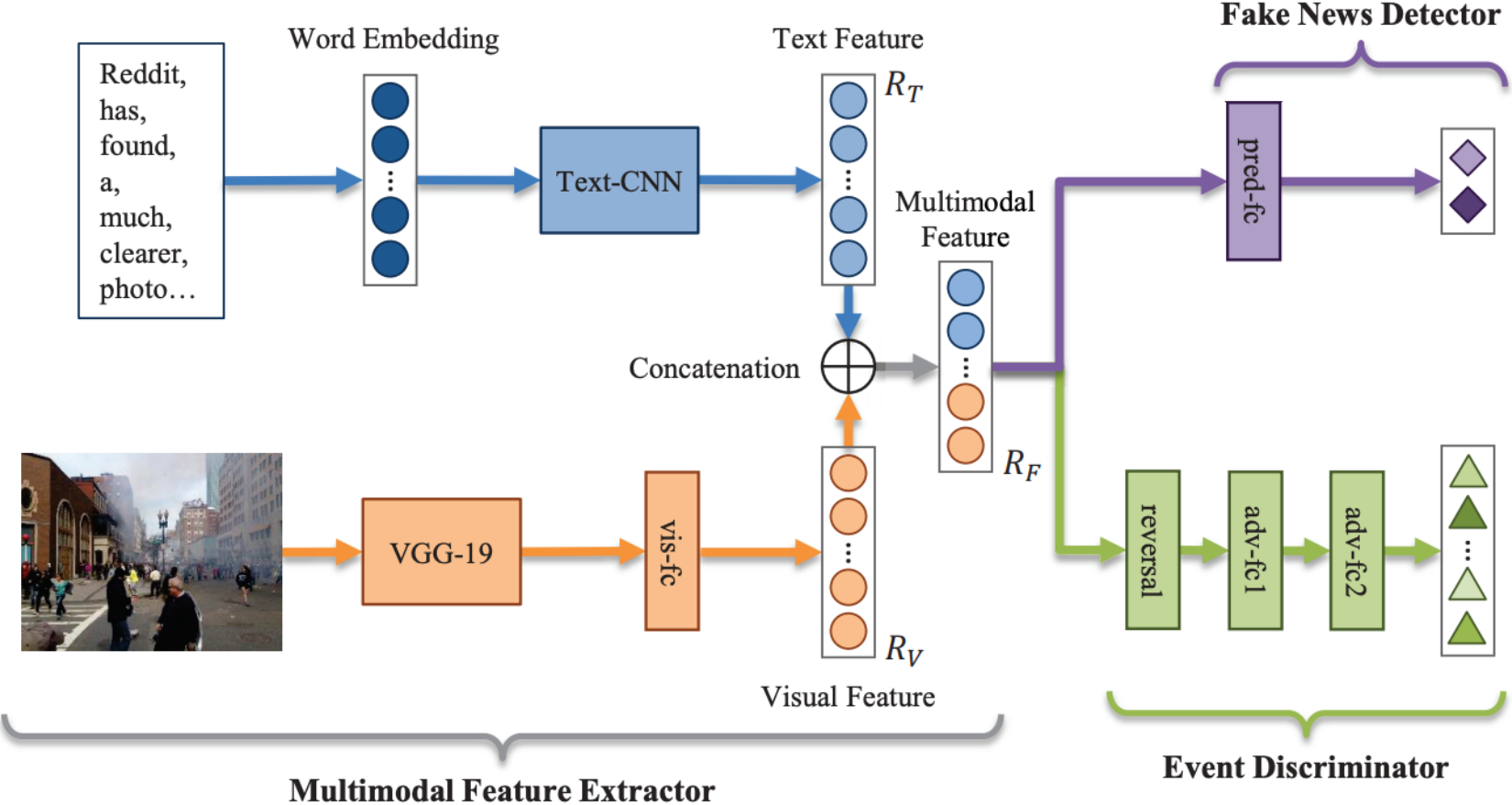}}
    \subfigure[\textsc{SAFE} (directly from \cite{zhou2020multimodal})]{\label{subfig:safe}
    \includegraphics[width=0.6\textwidth]{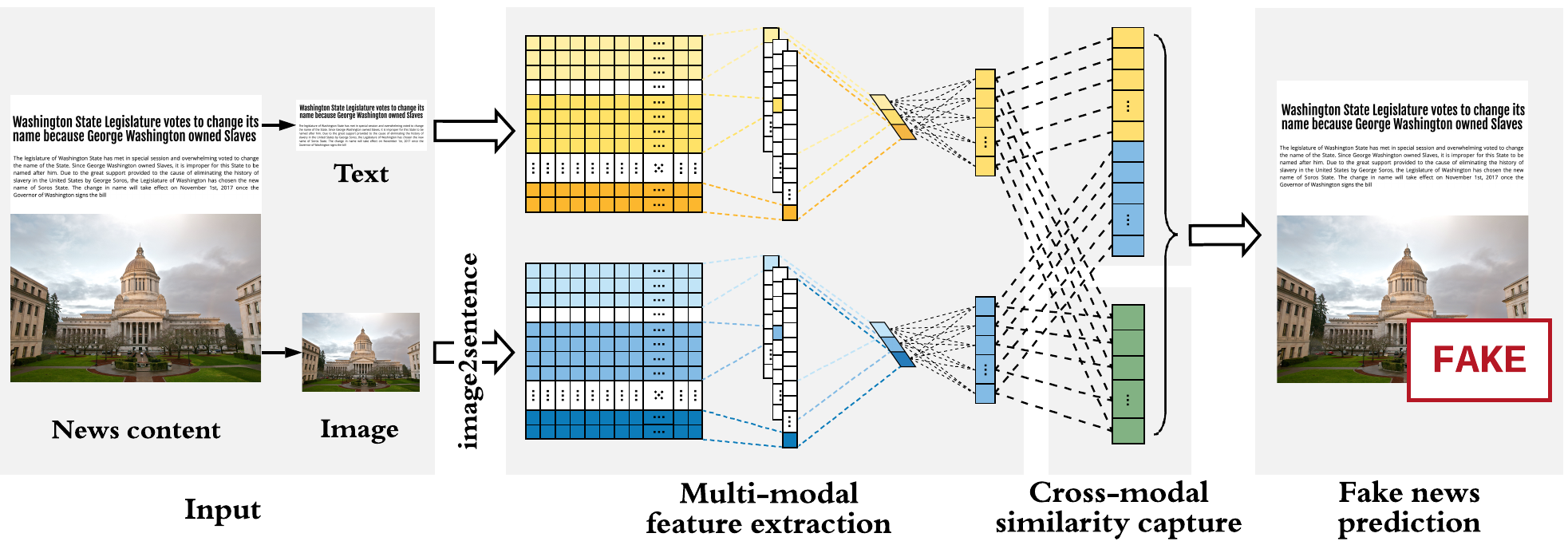}}
    \vspace{-3mm}
    \caption{Multimodal Fake News Detection Models}\vspace{-4mm}
    \label{fig:multimodal}
\end{figure}

Within a deep learning framework, news content (text and/or images) is often first embedded at the word-level~\cite{mikolov2013efficient} (for text), or as a pixel matrix or tensor (for images). Then, such an embedding is  processed by a well-trained neural network (e.g., 
CNNs~\cite{lecun1989backpropagation,krizhevsky2012imagenet,szegedy2015going,he2016deep,huang2017densely} such as VGG-16/19~\cite{simonyan2014very} and Text-CNN~\cite{kim2014convolutional};  
RNNs such as LSTMs~\cite{hochreiter1997long}, GRUs~\cite{cho2014learning}, and BRNNs~\cite{schuster1997bidirectional}; and the 
Transformer~\cite{vaswani2017attention,devlin2018bert}) to extract latent textual and/or visual features of news content. Ultimately, the given news content is  classified as true news or fake news often by concatenating and feeding all these features to a well-trained classifier such as a softmax.

This general procedure can be improved to, e.g., facilitate explainable fake news detection and enhance feature representativeness -- more discussions on \textit{explainable fake news detection} are provided in Section \ref{sec::discussion}. 
An example is the Event Adversarial Neural Network
(\textsc{EANN})~\cite{wang2018eann}, which can enhance feature representativeness by extracting features that are invariant under different world events to represent news content (text and images). Fig. \ref{subfig:eann} presents the architecture of the \textsc{EANN} model, which has three components: 
(1) multi-modal feature extractor, which extracts both textual and visual features from a given news article using neural networks; 
(2) event discriminator, which further captures event-invariant features of the given news by playing a \textit{min-max game}; and 
(3) fake news detector for news classification (true or fake). In \textsc{EANN}, the event discriminator ensures that the extracted features are representative. Another example is \textsc{SAFE}, a multimodal fake news detection method (see its architecture in Fig.~\ref{subfig:safe}). \textsc{SAFE} explores the relationship between the textual and visual features in a news article to detect fake news based on the falsity that can exist in the news multimodal and/or relational information~\cite{zhou2020multimodal}. Specifically, \textsc{SAFE} assumes that a ``gap'' often exists between the textual and visual information of fake news by observing that 
(i)~to attract public attention, some fake news writers prefer to use attractive while irrelevant images, and (ii)~when a fake news article tells a fake story, it is difficult to find both pertinent and nonmanipulated images to match such fake contents.

\begin{figure}[t]
    \centering
    \includegraphics[width=.245\textwidth]{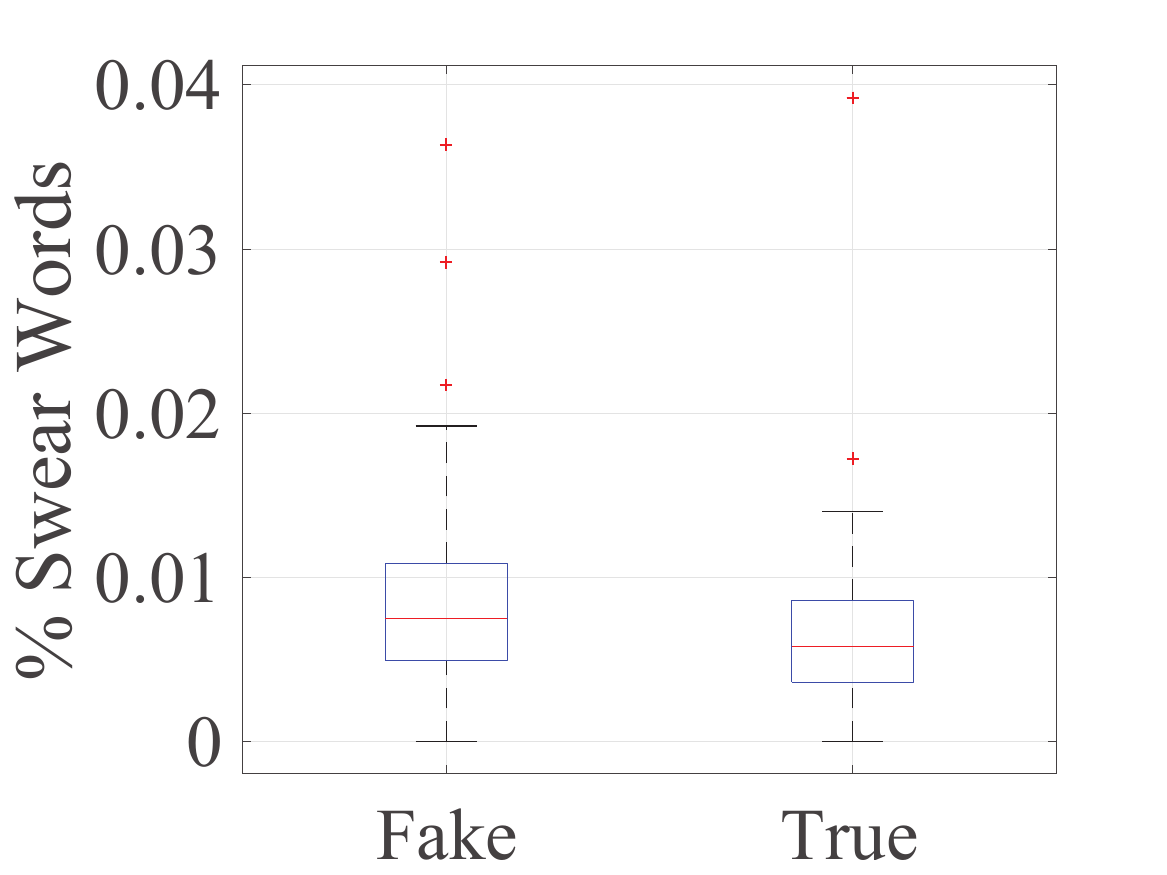}
    \includegraphics[width=.245\textwidth]{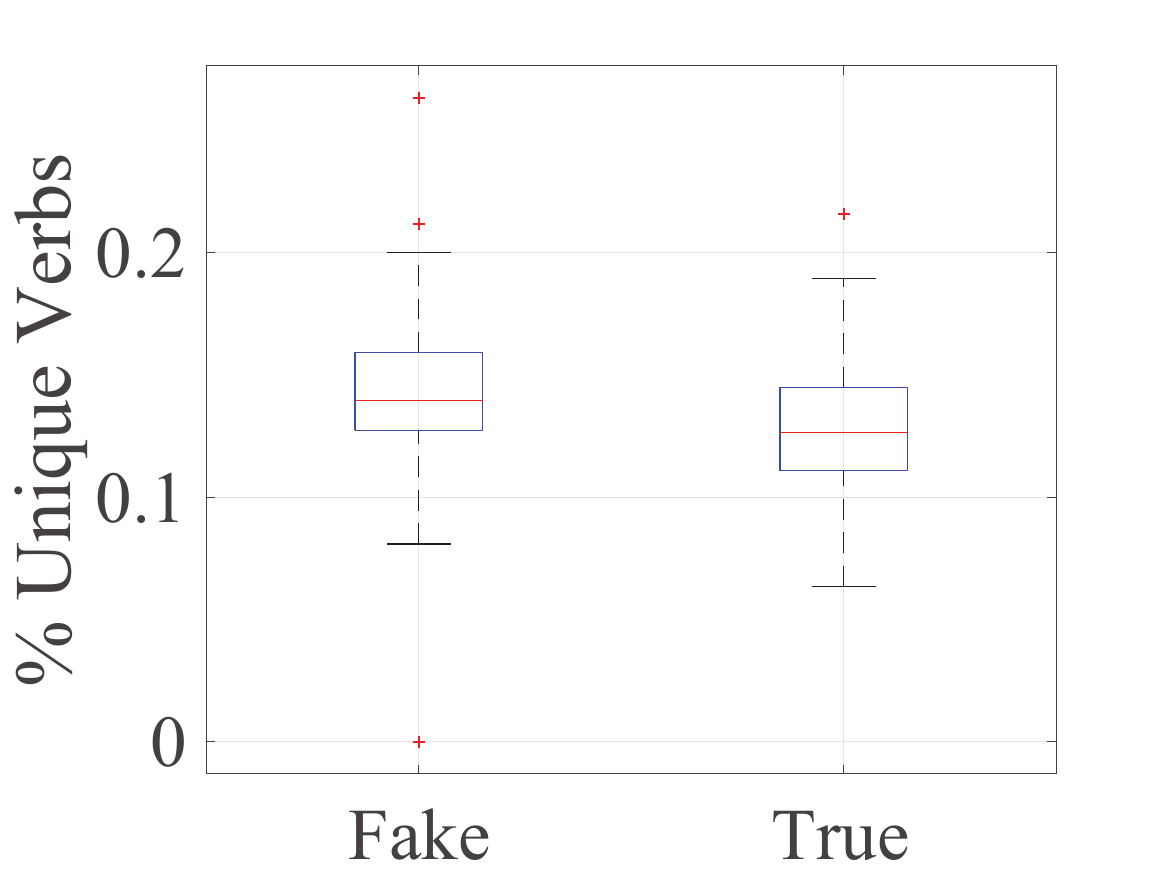} 
    \includegraphics[width=.245\textwidth]{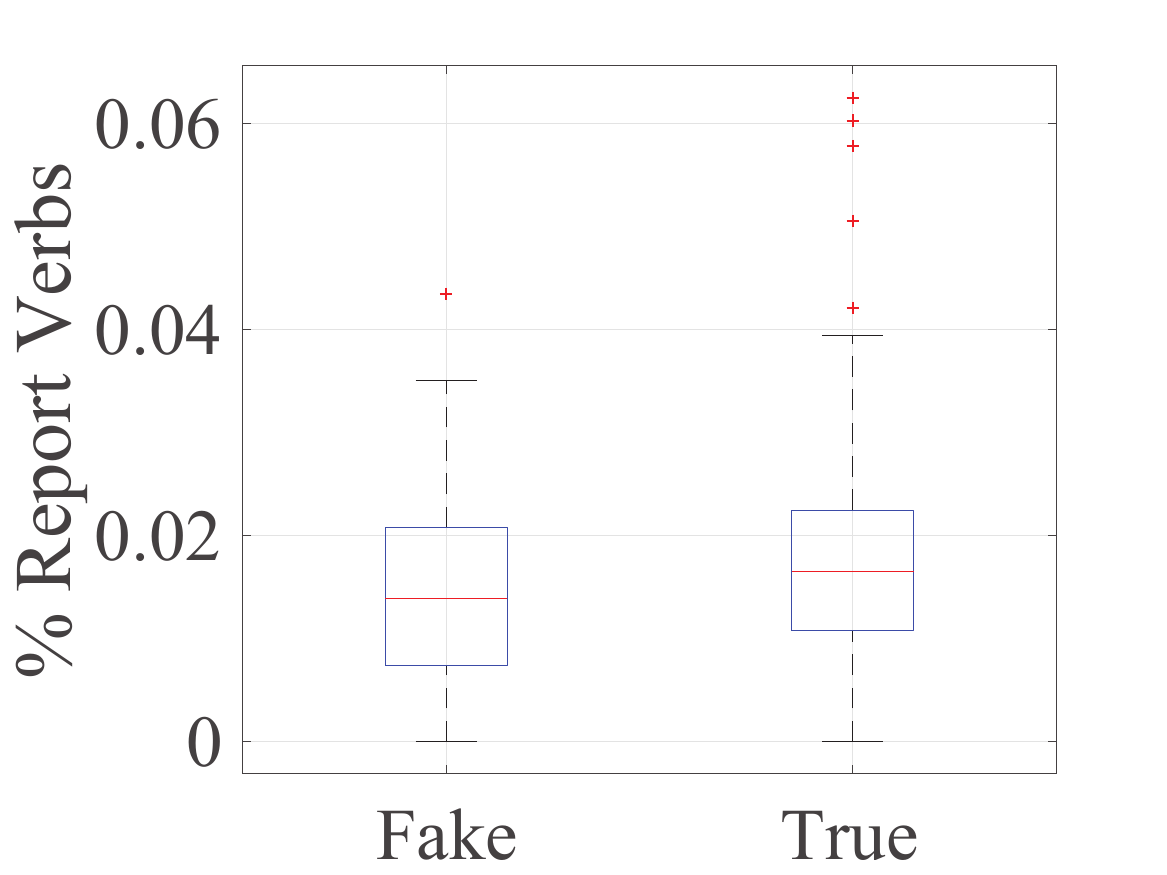}
    \includegraphics[width=.245\textwidth]{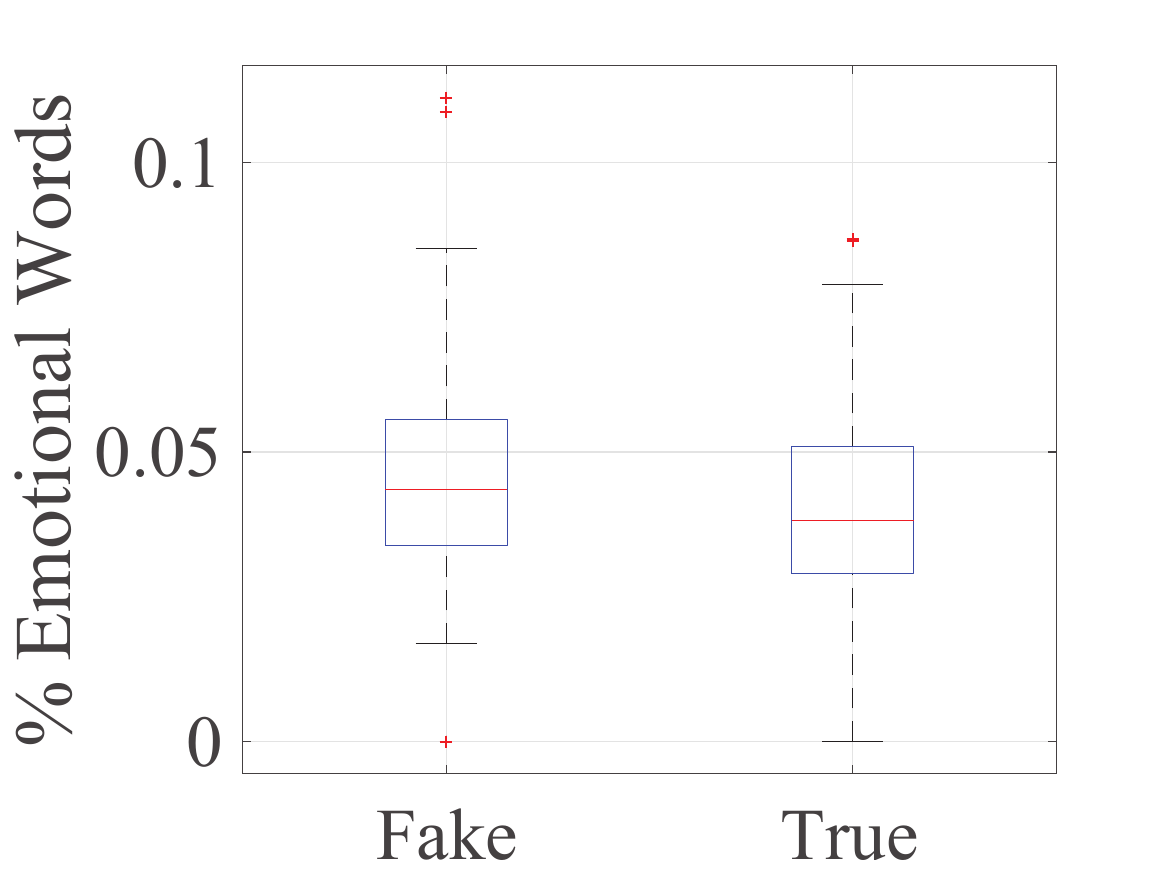} 
    \vspace{-1.5mm}\caption{Fake News Textual Patterns~\cite{zhou2019content} (PolitiFact, data is from FakeNewsNet~\cite{shu2018fakenewsnet}): Compared to true news text, fake news text has (i) higher informality (\% swear words), (ii) diversity (\% unique verbs), (iii) subjectivity (\% report verbs), and is (iv) more emotional (\% emotional words).
    }\vspace{-2.5mm}
    \label{fig:pattern_text}
\end{figure}
\begin{figure}
    \centering
    \includegraphics[width=.24\textwidth]{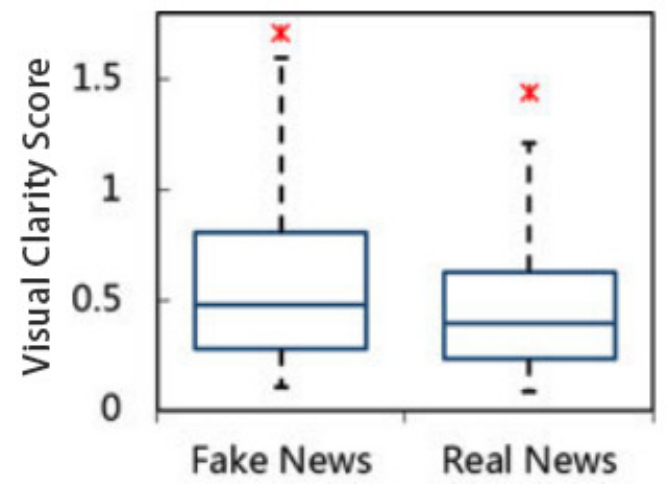}
    \includegraphics[width=.24\textwidth]{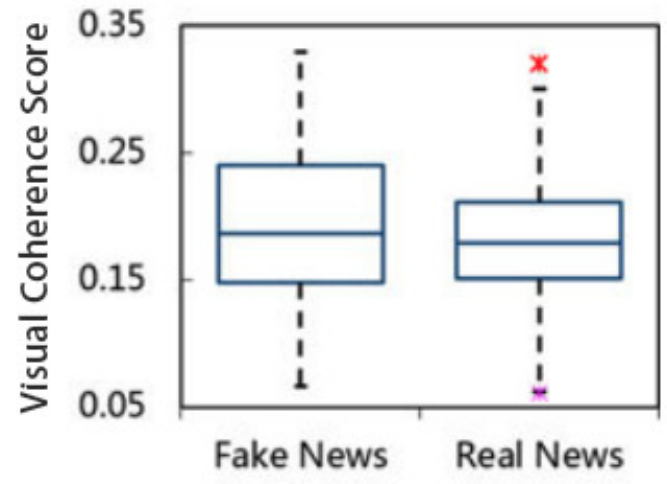}
    \includegraphics[width=.24\textwidth]{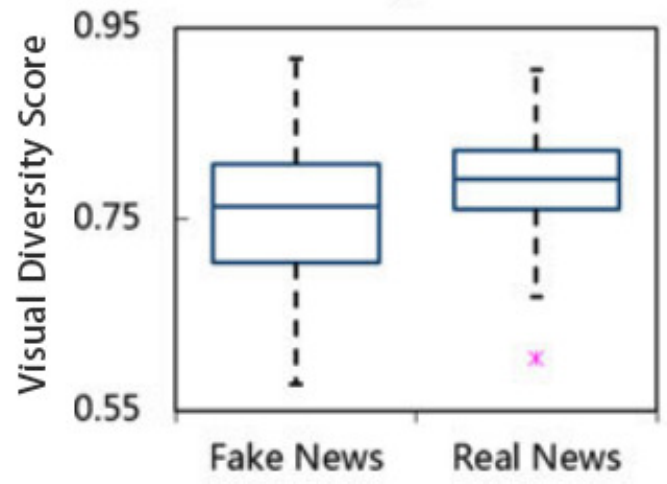}
    \includegraphics[width=.24\textwidth]{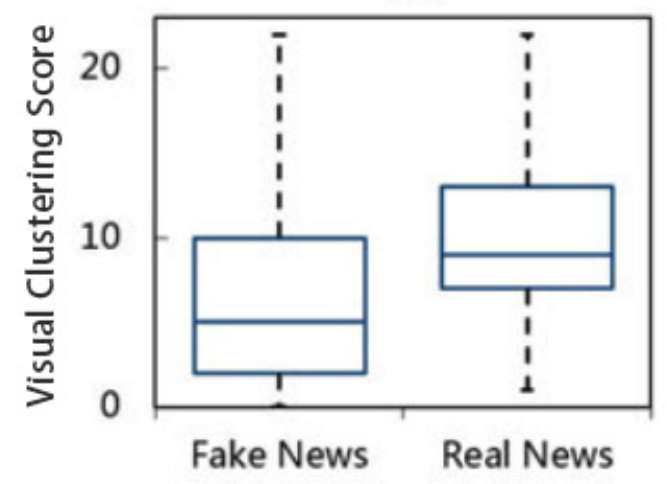}
    \vspace{-1.5mm}\caption{Fake News Visual Patterns (Twitter+Weibo, directly from \cite{jin2017novel}): Compared to true news images, fake news images often have higher clarity and coherence, while lower diversity and clustering score (see Table~\ref{tab:image_features} for a description of these features).
    }\vspace{-3mm}
    \label{fig:pattern_image}
\end{figure}

\subsection{Patterns of Fake News Content Style}
\label{subsec:style_pattern}

Some patterns of fake news content (text and images) style are distinguishable from those of the true news. Recent studies have revealed such patterns~\cite{zhou2019content,jin2017novel}. Particularly,
\begin{itemize}[leftmargin=*]
    \item \textit{Fake news text}, compared to true news text as shown in Fig. \ref{fig:pattern_text}, has higher (i) informality (\% swear words), (ii) diversity (\% unique verbs), (iii) subjectivity (\% report verbs), and is (iv) more emotional (\% emotional words); and
    \item \textit{Fake news images}, compared to true news images as illustrated in Fig. \ref{fig:pattern_image}, often have higher clarity and coherence, while lower diversity and clustering scores (see Table~\ref{tab:image_features} for a description of these features).
\end{itemize}

\subsection{Discussion}
\label{subsec:style_discussion}

We have detailed how to (1) represent and (2) classify news content style, the two main components of style-based fake news detection, along with some [textual and visual] patterns within fake news content that can help distinguish it from true news content. It should be pointed out that patterns in Fig. \ref{fig:pattern_text} are limited to political news articles and those in Fig. \ref{fig:pattern_image} target a mix of English (Twitter) and Chinese (Weibo) news articles from vague domains. Hence, a more comprehensive analysis of fake news content style across different domains, languages, time periods, etc. is highly encouraged as fake news content style may vary across domains and languages, evolve over time, and ultimately impact prediction performance~\cite{perez2017automatic} (more discussions on this topic are presented in Section \ref{sec::discussion}).
Furthermore, the writing style can be manipulated. Style-based (or knowledge-based) fake news detection relies heavily on news content, which enables the models to identify fake news before it has been propagated on social media (i.e., to achieve \textit{fake news early detection}, which we discuss more in Section \ref{sec::discussion}). However, such heavy dependence ``helps'' malicious entities bypass style-based models by changing their writing style. In other words, style-based fake news detection sometimes can be a cat-and-mouse game; any success at detection, in turn, will inspire future countermeasures by fake news writers. To resist such an attack, one can further involve social context information into news analysis to enhance model robustness, which we will discuss next in Sections \ref{sec::propagation} and \ref{sec::credibility}.

\section{Propagation-based Fake News Detection} 
\label{sec::propagation}

When detecting fake news from a propagation-based perspective, one can investigate and utilize the information related to the dissemination of fake news, e.g., how users spread it. Similar to style-based fake news detection, propagation-based fake news detection is often formulated as a binary (or multi-label) classification problem as well, however, with a different input. Broadly speaking, the input to a propagation-based method can be either a (I) \textit{news cascade}, a direct representation of news propagation, or a (II) self-defined graph, an indirect representation capturing additional information on news propagation. Hence, propagation-based fake news detection boils down to classifying (I) news cascades or (II) self-defined graphs. We review fake news detection using news cascades in Section \ref{subsec:propagation_cascades} and fake news detection using self-defined propagation graphs in Section \ref{subsec:propagation_graphs}, and provide our discussions in Section \ref{subsec:propagation_discussion}.

\subsection{Fake News Detection using News Cascades}
\label{subsec:propagation_cascades}

We first define a \textit{news cascade} in Definition \ref{def::fnCascade}, a formal representation of news dissemination that has been adopted by many studies (e.g., \cite{vosoughi2018spread,ma2018rumor,wu2015false,castillo2011information}).

\begin{myDef}[News Cascade] 
\label{def::fnCascade}
A news cascade is a tree or tree-like structure that directly captures the propagation of a certain news article on a social network (Fig. \ref{fig::fnCascade} provides examples). The root node of a news cascade represents the user who first shared the news article (i.e., initiator); Other nodes in the cascade represent users that have subsequently spread the article by forwarding it after it was posted by their parent nodes, which they are connected to via edges. A  news cascade can be represented in terms of the number of steps (i.e., hops) that the news has traveled (i.e., hop-based news cascade) or the times that it was posted (i.e., time-based news cascade).\vspace{1mm}

\noindent
\parbox[t]{2.7in}{\raggedright%
\textbf{Hop-based news cascade}, often a standard tree, allowing natural measures such as
\begin{compactenum}
\item[-] Depth: the maximum number of steps (hops) that the news has traveled within a cascade;
\item[-] Breadth (at hop $k$): the number of users that have spread the news $k$ steps (hops) after it was initially posted within a cascade; and
\item[-] Size: the total number of users in a cascade.
\end{compactenum}
}\hfill
\parbox[t]{2.9in}{\raggedright%
\textbf{Time-based news cascade}, often a tree-like structure, allowing natural measures such as
\begin{compactenum}
\item[-] Lifetime: the longest interval during which the news has been propagated;
\item[-] Real-time heat (at time $t$): the number of users posting/forwarding the news at time $t$; and
\item[-] Overall heat: the total number of users that have forwarded/posted the news.
\end{compactenum}
}
\begin{figure}[htbp]
\begin{minipage}{0.45\textwidth}
  \centering
  \includegraphics[scale=0.7]{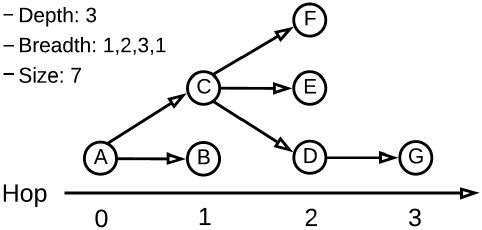}
\end{minipage}
\begin{minipage}{0.45\textwidth}
  \centering
  \includegraphics[scale=0.7]{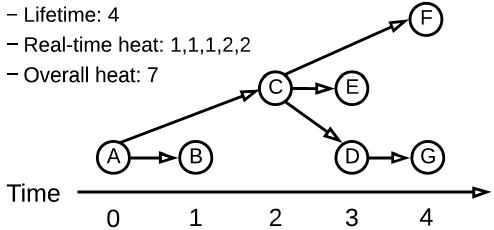}  
\end{minipage}
\vspace{-3mm}
\caption{Illustrations of News Cascades}
\label{fig::fnCascade}
\vspace{-3mm}
\end{figure}
\end{myDef}

Note that a specific news article can lead to multiple simultaneous cascades due to multiple initiating users. Furthermore, often within a news cascade, nodes (users) are represented with a series of attributes and additional information, e.g., whether they (support or oppose) the fake news, their profile information, previous posts, and their comments. 

Based on Definition \ref{def::fnCascade}, classifying a news article (true or fake) using its cascade(s) boils down to classifying its cascade(s) as true or fake. To perform this classification, some proposed methods rely on (I)  traditional machine learning, while others utilize (II) [deep] neural networks.

\begin{table}[t]
\centering
\caption{News Cascade Features}
\label{tab:cascade_features}
\small
\begin{adjustbox}{width = \textwidth}
\begin{threeparttable}
\begin{tabular}{|l|l|c|c||c|}
\hline
\rowcolor{gray!20}\textbf{Feature Type} &\textbf{Feature Description} & \textbf{H} & \textbf{T} & \textbf{P} \\ \hline \hline
\textbf{Cascade Size} 
    & Overall number of nodes in a cascade~\cite{castillo2011information,vosoughi2018spread} 
        & \checkmark    & \checkmark               & \checkmark \\ \hline
\textbf{Cascade Breath}
    & Maximum (or average) breadth of a news cascade~\cite{vosoughi2018spread} 
        & \checkmark    &                          & \checkmark \\ \hline
\textbf{Cascade Depth}
    & Depth of a news cascade~\cite{castillo2011information,vosoughi2018spread} 
        & \checkmark    &                           & \checkmark \\ \hline
\textbf{Structural Virality} 
    & Average distance among all pairs of nodes in a cascade~\cite{vosoughi2018spread}
        & \checkmark    &                           & \checkmark \\ \hline
\multirow{2}{*}{\textbf{Node Degree}} 
    & Degree of the root node of a news cascade~\cite{castillo2011information} 
        & \checkmark    &            &       \\ \cline{2-5}
    & Maximum (or average) degree of non-root nodes in a news cascade~\cite{castillo2011information} 
        & \checkmark    &            &  \\ \hline
\multirow{2}{*}{\textbf{Spread Speed}} 
    & Time taken for a cascade to reach a certain depth (or size)~\cite{vosoughi2018spread} 
        & \checkmark    &                & \checkmark \\ \cline{2-5} 
    & Time interval between the root node and its child nodes~\cite{wu2015false} 
        &               & \checkmark &             \\ \hline
\textbf{Cascade Similarity} & Similarity scores between a cascade and other cascades in the corpus~\cite{wu2015false} 
        & \checkmark \* &            &            \\ \hline
\end{tabular}
\begin{tablenotes}
\item \textbf{H}: Hop-based news cascades; \textbf{T}: Time-based news cascades; \textbf{P}: Pattern-driven features.
\end{tablenotes}
\end{threeparttable}
\end{adjustbox}
\end{table}

\paragraph{$\blacktriangleright$ Traditional Machine Learning (ML) Models} Within a traditional ML framework, to classify a news cascade that has been represented as a set of features, one often relies on supervised learning methods such as SVMs~\cite{castillo2011information,kwon2013prominent,wu2015false}, decision trees~\cite{castillo2011information,kwon2013prominent}, decision rules~\cite{castillo2011information}, na\"ive Bayes~\cite{castillo2011information}, and random forests~\cite{kwon2013prominent}. Table \ref{tab:cascade_features} summarizes the news cascade features used in current research. 

\begin{figure}[t]
  \begin{center}
  \begin{minipage}{\textwidth}
  \centering
  \includegraphics[width=0.245\textwidth]{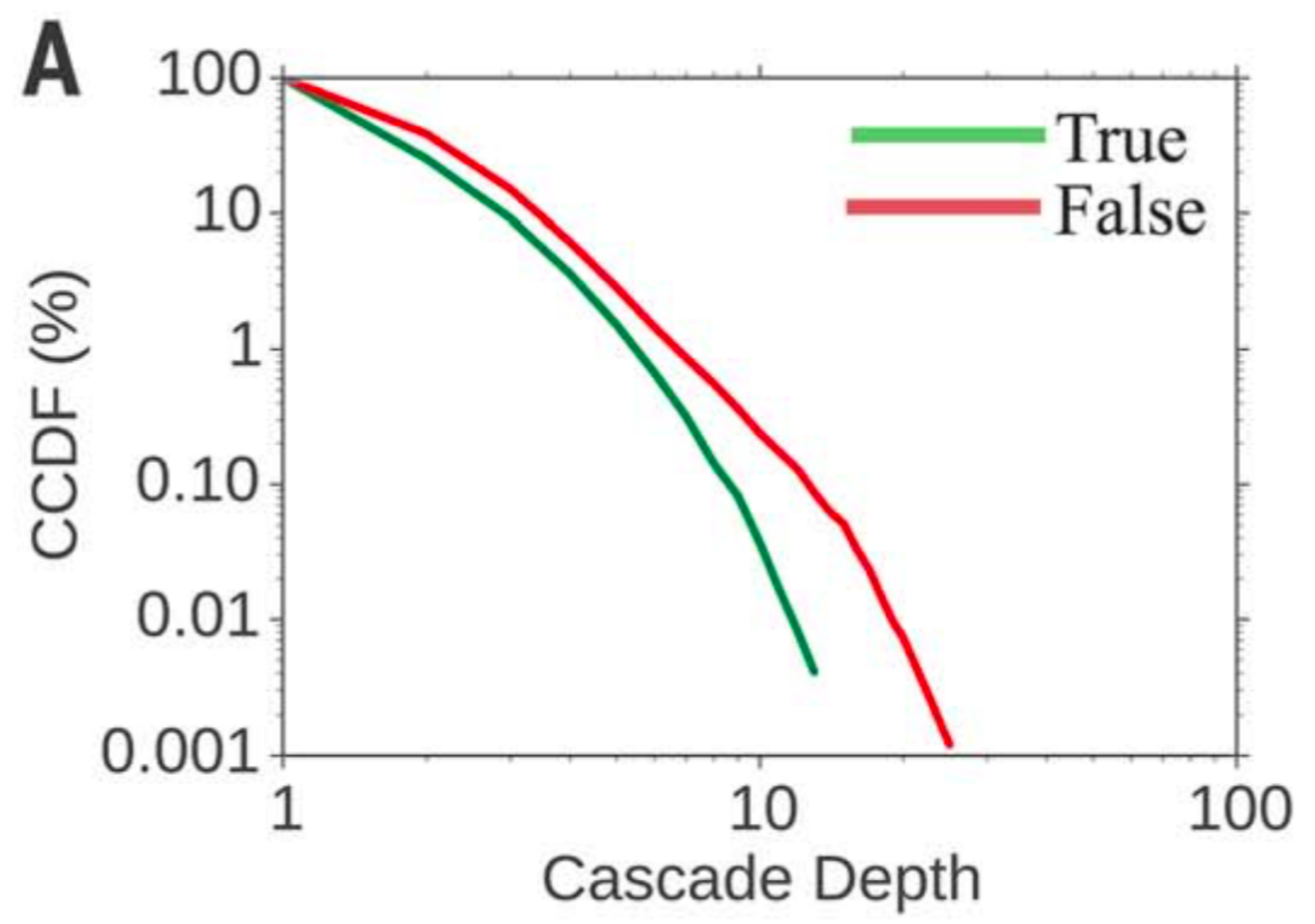}
  \includegraphics[width=0.245\textwidth]{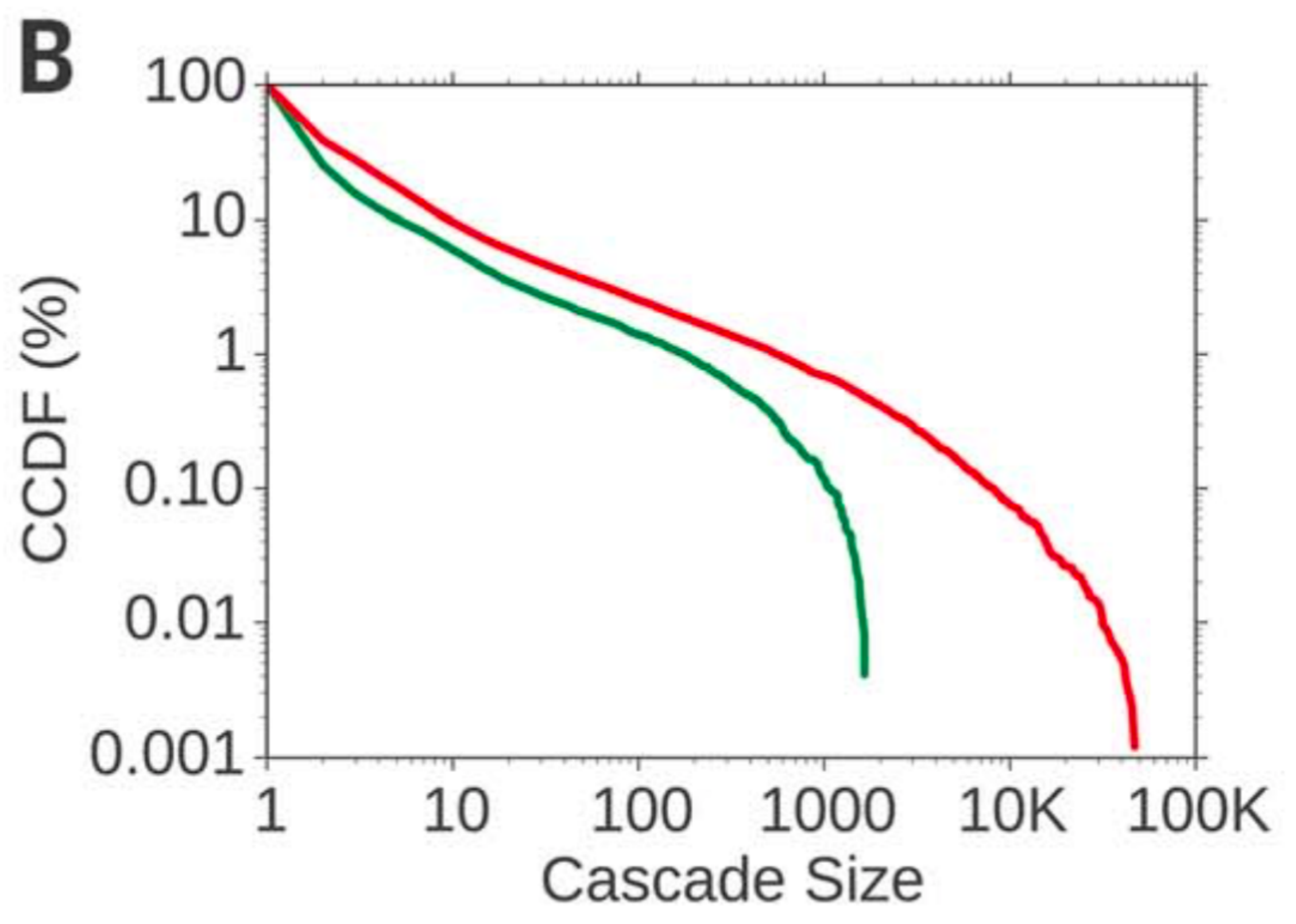}
  \includegraphics[width=0.245\textwidth]{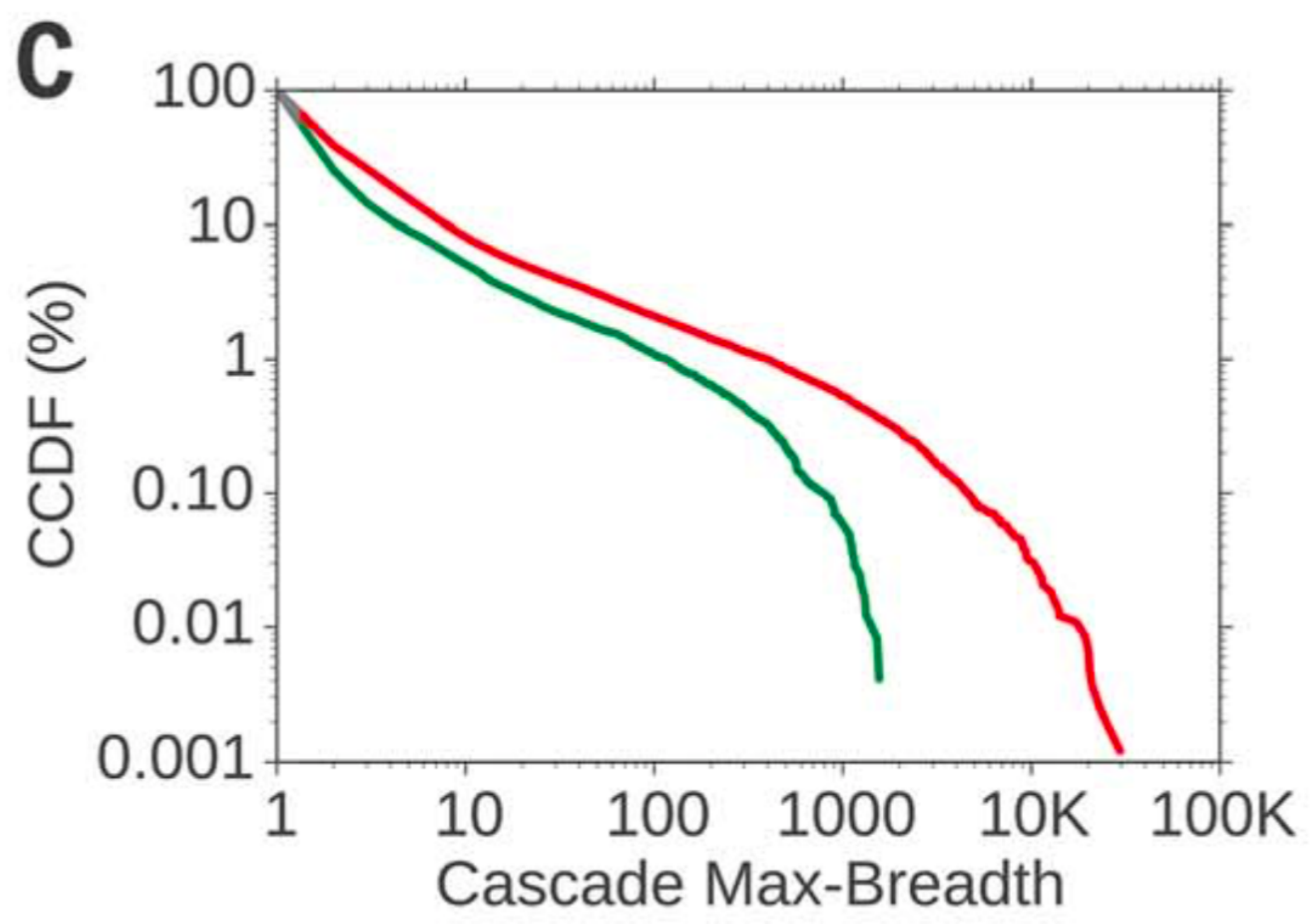}
  \includegraphics[width=0.245\textwidth]{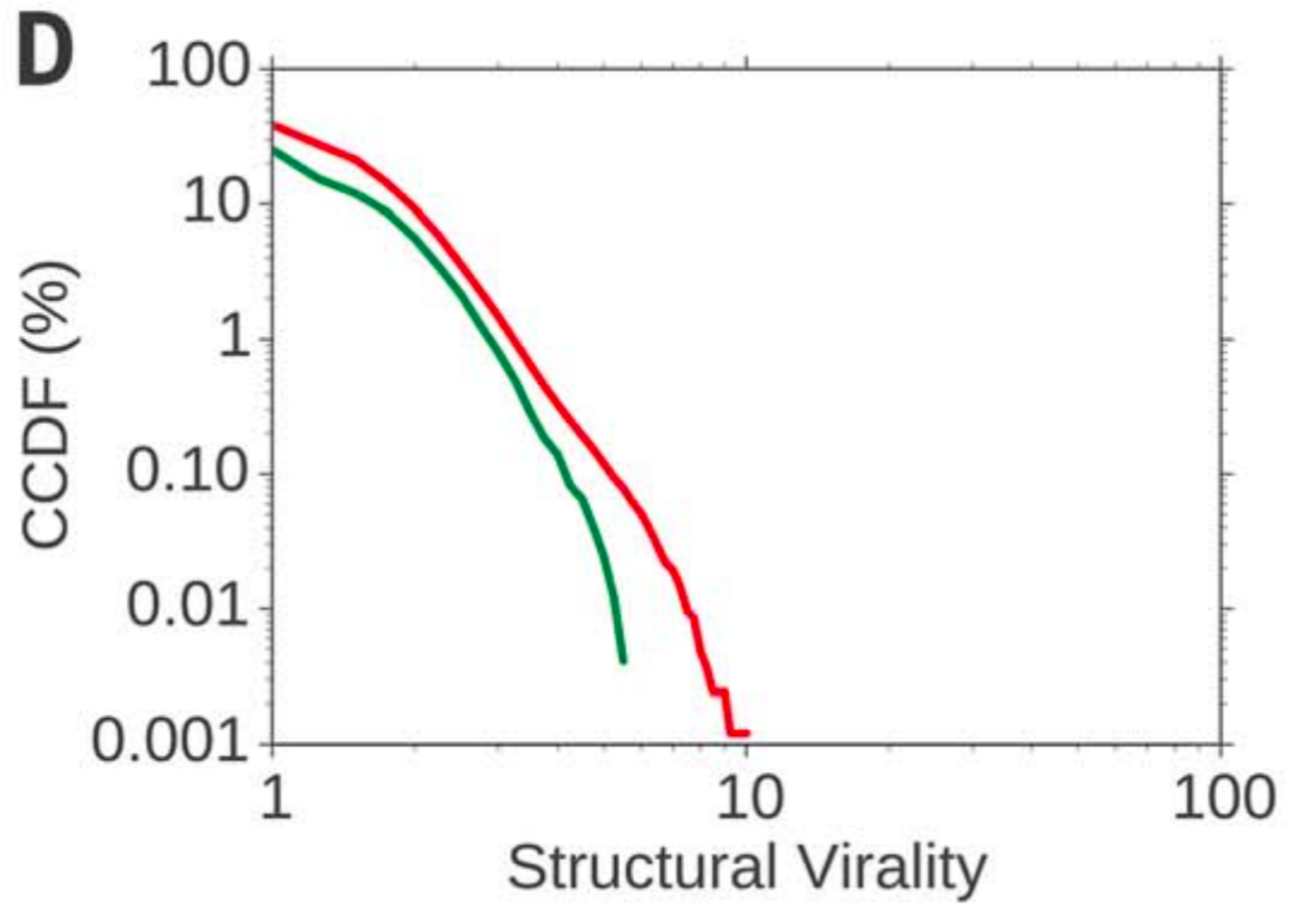}
  \includegraphics[width=0.245\textwidth]{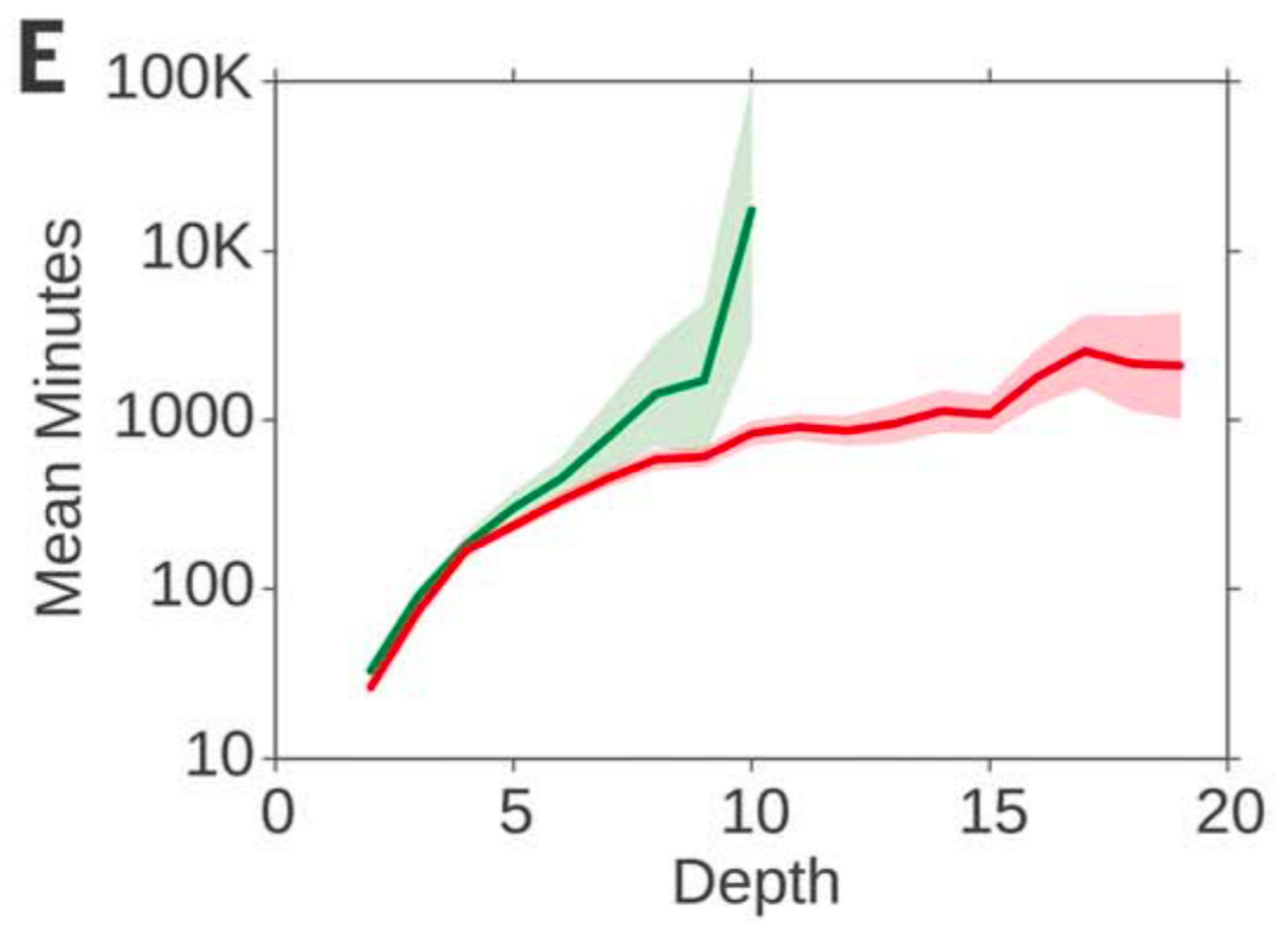}
  \includegraphics[width=0.245\textwidth]{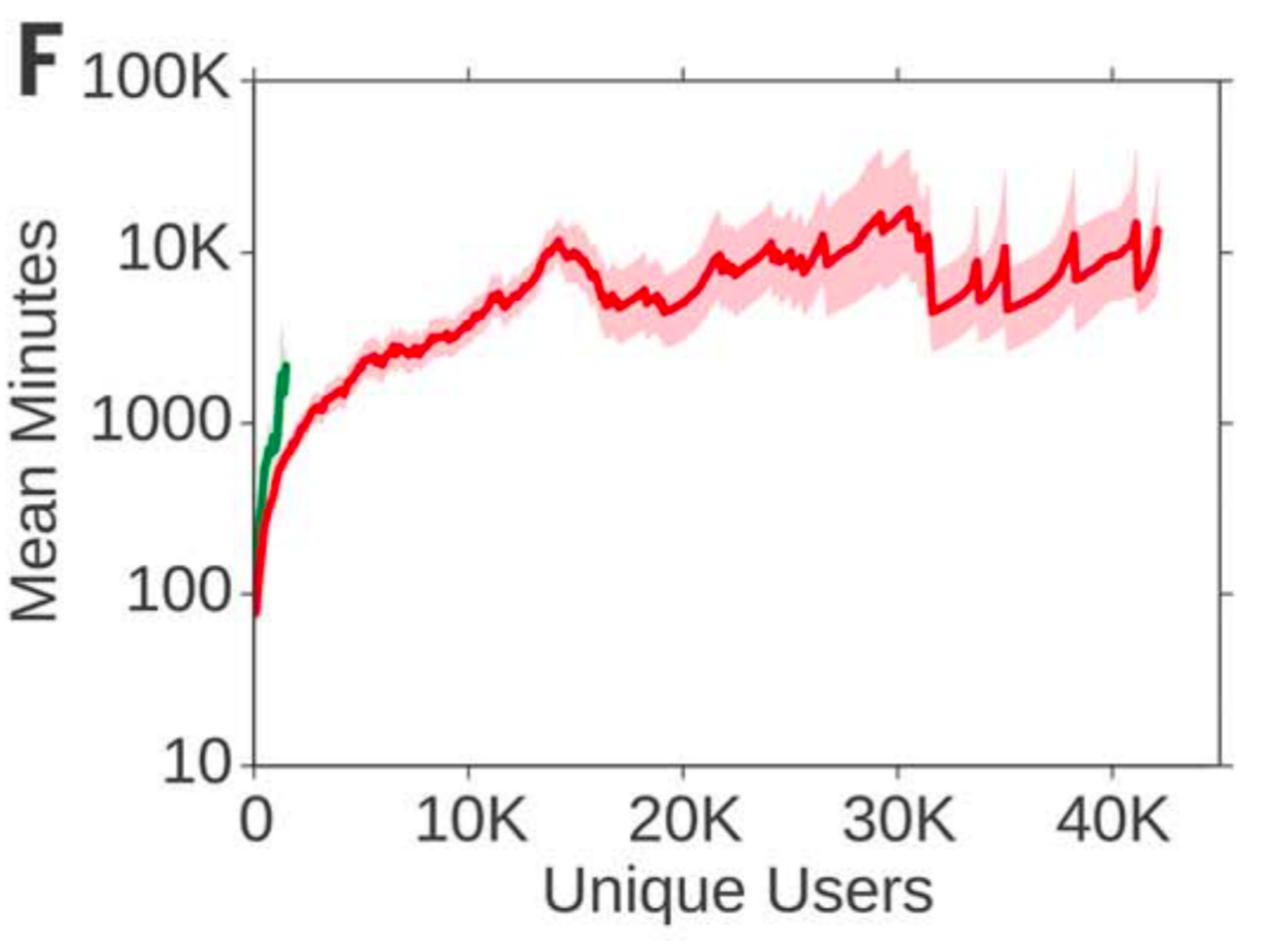}
  \includegraphics[width=0.245\textwidth]{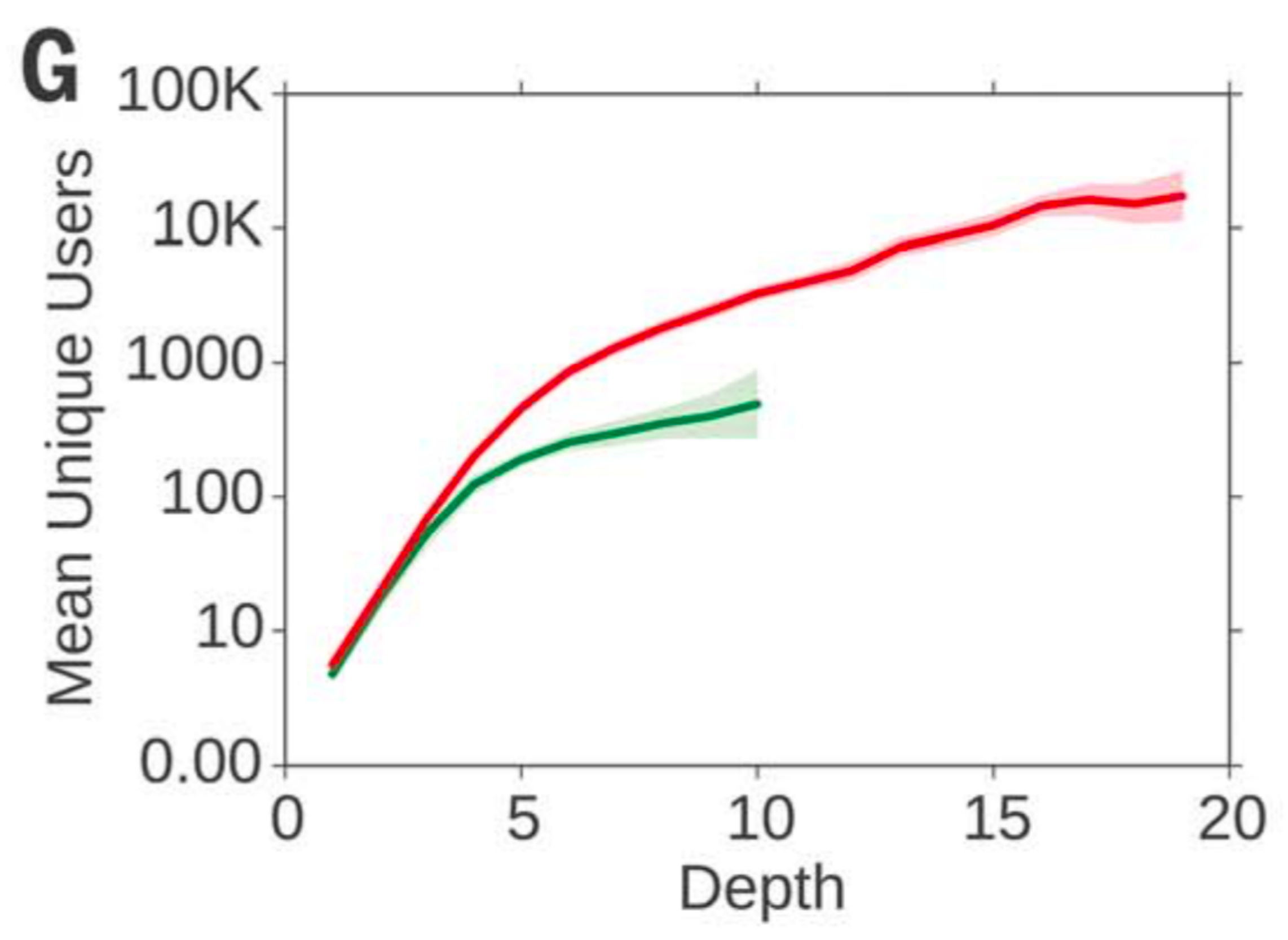}
  \includegraphics[width=0.245\textwidth]{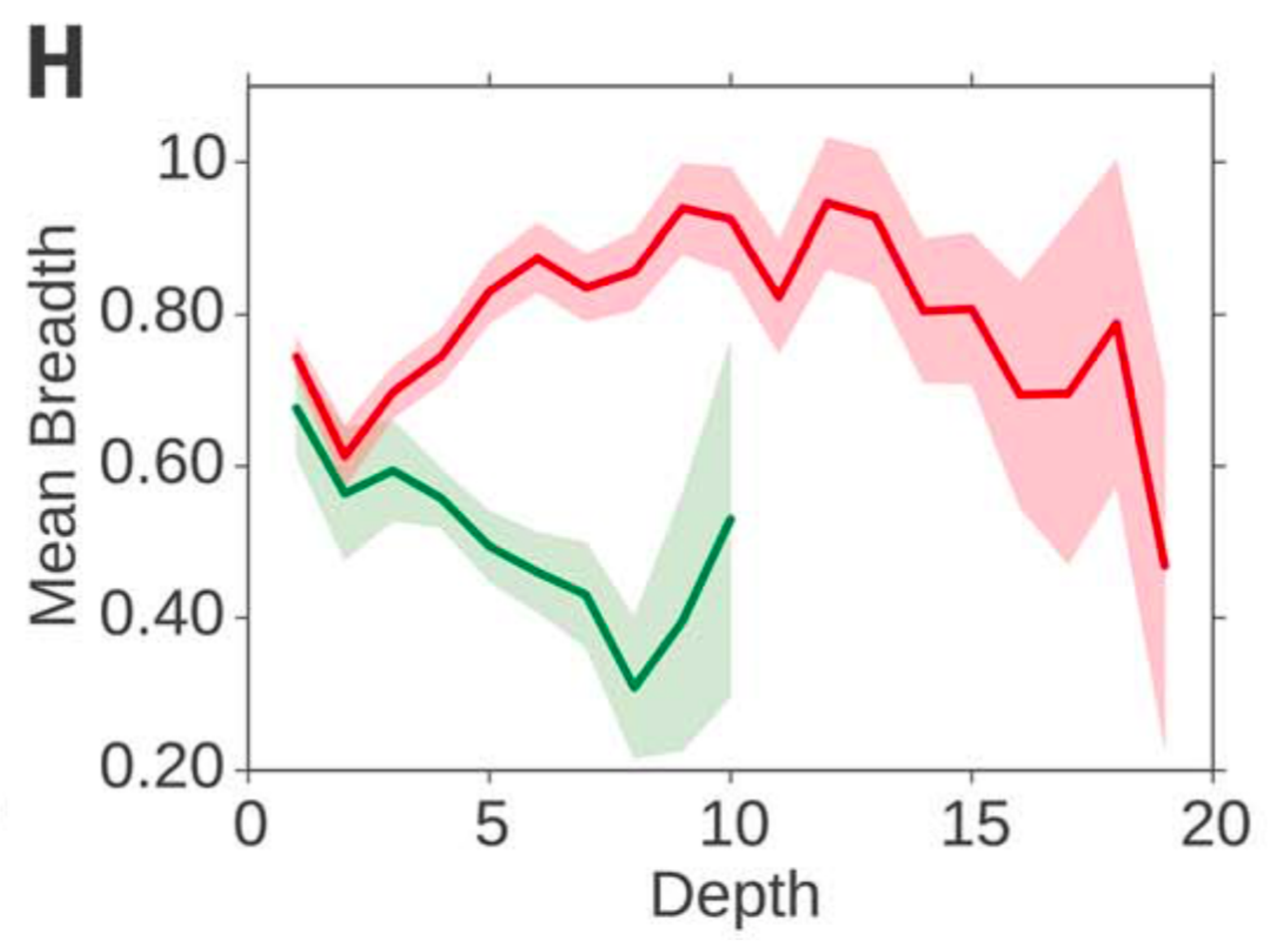}
  \end{minipage}
  \vspace{-3mm}
  \caption{Fake News Cascade-based Propagation Patterns (Twitter data, directly from \cite{vosoughi2018spread}). 
  (\textbf{A}-\textbf{D}): CCDF (Complementary Cumulative Distribution Function) distributions of cascade depth, size, max-breadth and structural virality of fake news are always above that of true news; 
  (\textbf{E}-\textbf{F}): The average time taken for fake news cascades to reach a certain depth and a certain number of unique users are both less than that for true news cascades; and (\textbf{G}-\textbf{H}): For fake news cascades, their average number of unique users and breadth at a certain depth are always greater than that of true news cascades.}
\label{fig::pattern_propagation}
\vspace{-3mm}
\end{center}
\end{figure}

As shown in Table \ref{tab:cascade_features}, cascade features can be inspired by  \textit{fake news propagation patterns} observed in empirical studies. For example, a recent study has investigated the differences in diffusion patterns of all verified true and fake news stories on Twitter from 2006 to 2017~\cite{vosoughi2018spread}. The study reveals that \textit{fake news spreads faster, farther, more widely, and is more popular with a higher structure virality score compared to true news}. Specifically, the cascade depth, maximum breadth (and mean breadth at every depth), size, and structural virality~\cite{goel2015structural} (i.e., the average distance among all pairs of nodes in a cascade) for fake news are generally greater than that of true news. Also, the time it takes for fake news cascades to reach any depth (and size) is less than that for true news cascades (see Fig. \ref{fig::pattern_propagation}). 

\begin{figure}[t]
  \begin{center}
\subfigure[A Traditional Machine Learning Model~\cite{wu2015false}]{  
\label{subfig::cascade1}
   \begin{minipage}{0.4\textwidth}
   \centering
     \includegraphics[width=0.85\textwidth]{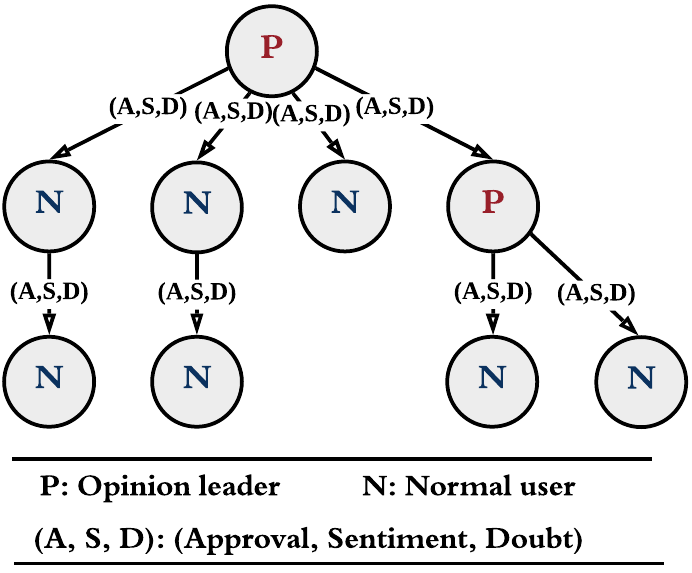}
     \end{minipage}} \qquad 
\subfigure[A Deep Learning Model~\cite{ma2018rumor}]{ \label{subfig::cascade2}
   \begin{minipage}{0.4\textwidth}
   \centering
     \includegraphics[width=0.90\textwidth]{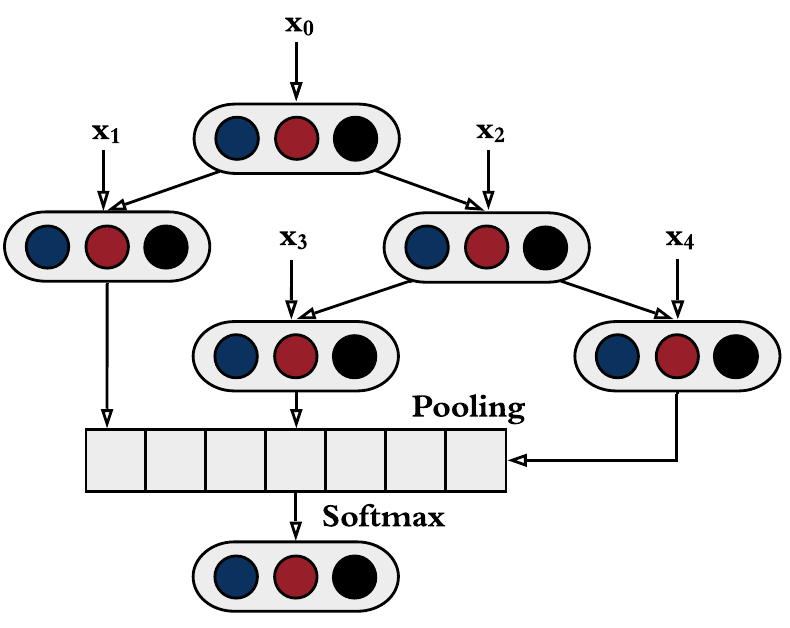}
     \end{minipage}} 
  \vspace{-3mm}
  \caption{Examples of Cascade-based Fake News Detection Models}
  \vspace{-3mm}
\label{fig::cascadeDetection}
\end{center}
\end{figure}

The cascade features in Table \ref{tab:cascade_features} can be viewed from another perspective, where they can either capture the local structure of a cascade, e.g., its depth and width~\cite{castillo2011information,vosoughi2018spread}, or allow comparing the overall structure of a cascade with that of other cascades by computing similarities~\cite{wu2015false}.
A common strategy to compute such graph similarities is to use \textit{graph kernels}~\cite{vishwanathan2010graph}.
For example, Wu et al. develop a \textit{random walk graph kernel} to compute the similarity between two hop-based news cascades that contain additional information, e.g., the user roles (\textit{opinion leader} or \textit{normal user}) as well as approval, sentiment, and doubt scores for user posts (see the structure of such a cascade in Fig.~\ref{subfig::cascade1}).  


\vspace{-2mm}
\paragraph{$\blacktriangleright$ Deep Learning (DL) Models}
Within a DL framework, learning the representation of news cascades often relies on neural networks, where a softmax function often acts as a classifier. For example, Ma et al. develop Recursive Neural Networks (RvNNs), a tree-structured neural network, based on news cascades~\cite{ma2018rumor,bian2020rumor}. A top-down RvNN model with Gated Recurrent Units (GRUs) is shown in Fig.~\ref{subfig::cascade2}. Specifically, for each node $j$ with a post on a certain news report represented as a TF-IDF vector $\mathbf{x}_j$, its hidden state $\mathbf{h}_j$ is recursively determined by $\mathbf{x}_j$ and the hidden state of its parent node $\mathcal{P}(j)$, denoted as $\mathbf{h}_{\mathcal{P}(j)}$. Formally, $\mathbf{h}_j$ is derived using a standard GRU formulation:
\begin{equation}
\begin{array}{rcl}
\mathbf{r}_j & = & \sigma (\bm{W}_r \mathbf{x}_j \mathbf{V}+\mathbf{U}_r \mathbf{h}_{\mathcal{P}(j)})~,\\
\mathbf{z}_j & = &\sigma (\bm{W}_z \mathbf{x}_j \mathbf{V}+\mathbf{U}_z \mathbf{h}_{\mathcal{P}(j)})~,\\
\mathbf{h}_j & = &\mathbf{z}_j \odot \tanh (\mathbf{W}_h \mathbf{x}_j \mathbf{V} + \mathbf{U}_h (\mathbf{h}_{\mathcal{P}(j)}\odot \mathbf{r}_j)) +(1-\mathbf{z}_j) \odot \mathbf{h}_{\mathcal{P}(j)}~,
\end{array}
\end{equation}
where $\mathbf{r}_j$ is a reset gate vector, $\mathbf{z}_j$ is an update gate vector, $\mathbf{W}_*$, $\mathbf{U}_*$, and $\mathbf{V}$ denote parameter matrices, $\sigma(\cdot)$ is the sigmoid function, $\tanh(\cdot)$ is the hyperbolic tangent, and $\odot$ denotes entry-wise product. In this way, a representation (hidden state) is learned for each leaf node in the cascade. These representations are used as inputs to a \textit{max-pooling} layer, which computes the final representation $\mathbf{h}$ for the news cascade. Max-pooling layer takes the maximum value of each dimension of the hidden state vectors over all the leaf nodes. Finally, the label of news cascade is predicted as 
\begin{equation}
\tilde{y} =\mathrm{Softmax}(\mathbf{Q}\mathbf{h}+\mathbf{b}),
\end{equation} 
where $\mathbf{Q}$ and $\mathbf{b}$ are parameters. The model (parameters) can be trained (estimated) by minimizing some cost function, e.g., squared error~\cite{ma2018rumor} and cross-entropy~\cite{zhang2018fake}, using some optimization algorithm, e.g., Stochastic Gradient Descent (SGD)~\cite{wang2018eann}, Adam~\cite{kingma2014adam}, and Alternating Direction Method of Multipliers (ADMM)~\cite{boyd2011distributed,shu2019beyond}.

\subsection{Fake News Detection using Self-defined Propagation Graphs}
\label{subsec:propagation_graphs}

When detecting fake news using self-defined propagation graphs (networks), one constructs flexible networks to indirectly capture fake news propagation. These networks can be (1) homogeneous, (2) heterogeneous, or (3) hierarchical.

\begin{figure}[t]
\begin{minipage}{\textwidth}
\begin{minipage}{0.29\textwidth}
\centering
    \subfigure[Spreader Net.~\cite{zhou2019network}]{
    \label{subfig:spreaderNet}
      \includegraphics[width=1.02\textwidth]{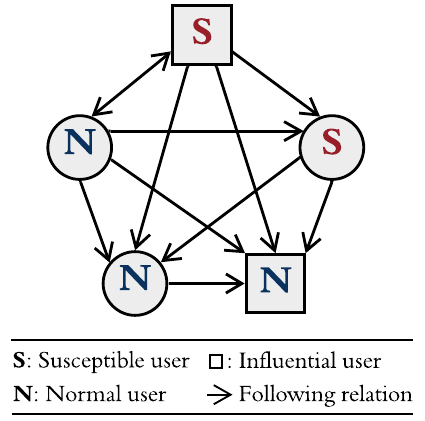}} 
    \subfigure[Stance Net.~\cite{jin2016news}]{
    \label{subfig:stanceNet}
      \includegraphics[width=0.85\textwidth]{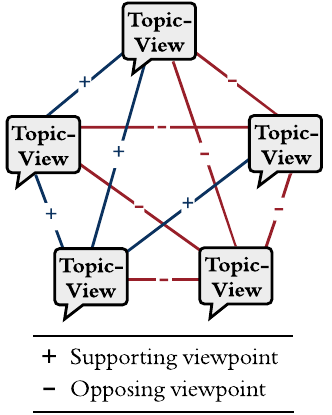}}
    \vspace{-2mm}
    \caption{Homogeneous Networks}
    \label{fig:homoNet}
    \vspace{-2mm}
\end{minipage}\quad
\begin{minipage}{0.33\textwidth}
\centering
    \subfigure[\cite{shu2019beyond}]{
    \label{subfig:heteroNet1}
      \includegraphics[width=0.95\textwidth]{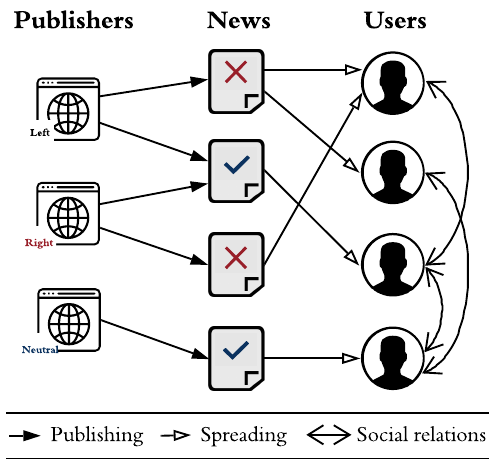}}
    \subfigure[\cite{zhang2018fake}]{
    \label{subfig:heteroNet2}
      \includegraphics[width=0.9\textwidth]{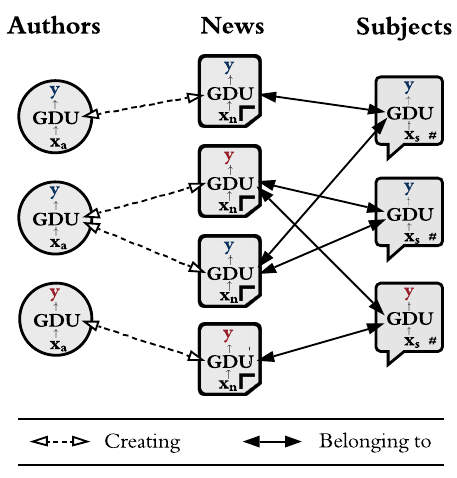}}
    \vspace{-2mm}
    \caption{Heterogeneous Networks}
    \label{fig:heteroNet}
    \vspace{-2mm}
\end{minipage}\quad 
\begin{minipage}{0.33\textwidth}
\centering
    \subfigure[\cite{shu2019hierarchical}]{
    \label{subfig:hieraNet1}
      \includegraphics[width=0.95\textwidth]{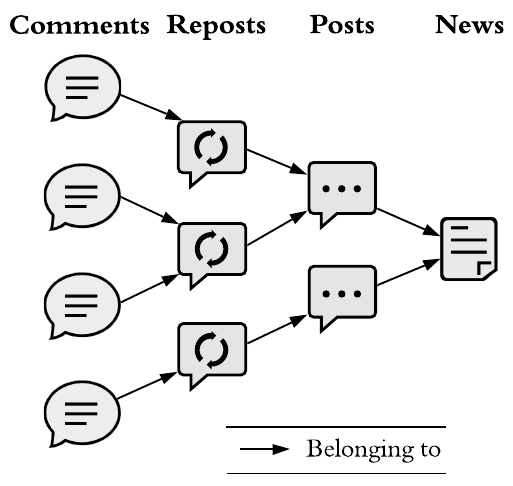}}         
    \subfigure[\cite{jin2014news}]{
    \label{subfig:hieraNet2}
      \includegraphics[width=.91\textwidth]{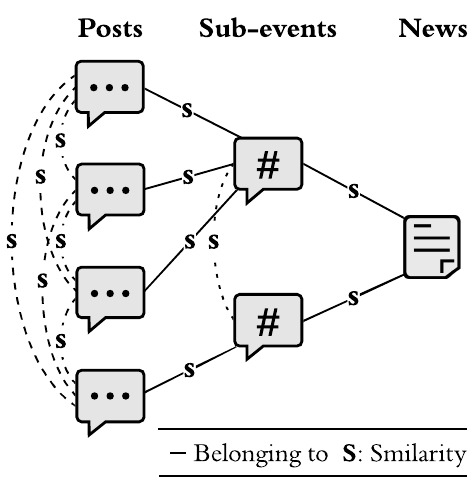}}
    \vspace{-2mm}
    \caption{Hierarchical Networks}
    \label{fig:hieraNet}
    \vspace{-2mm}
\end{minipage}
\end{minipage}
\end{figure}

\begin{figure}[t]
    \centering
    \subfigure[\textsc{Farther-Distance}]{
    \includegraphics[width=.235\textwidth]{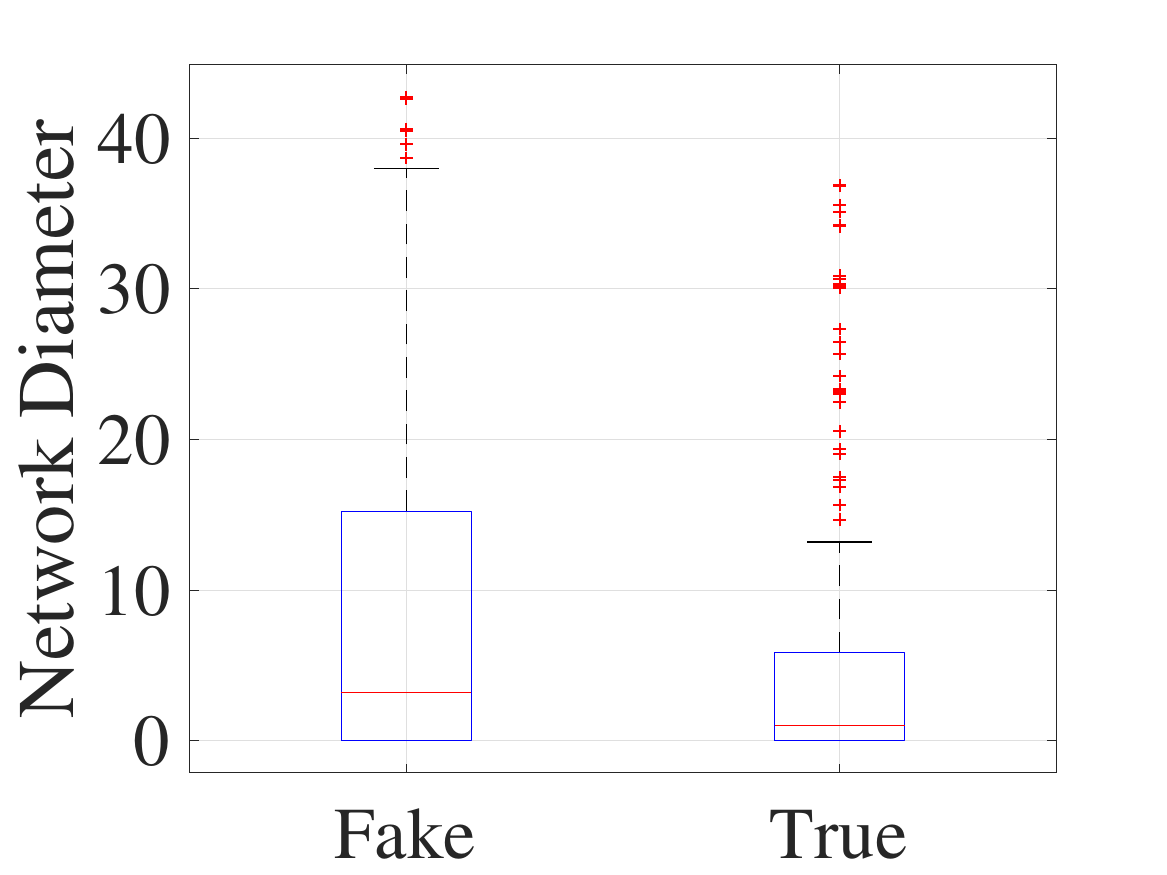}}
    \subfigure[\textsc{More-Spreaders}]{
    \includegraphics[width=.235\textwidth]{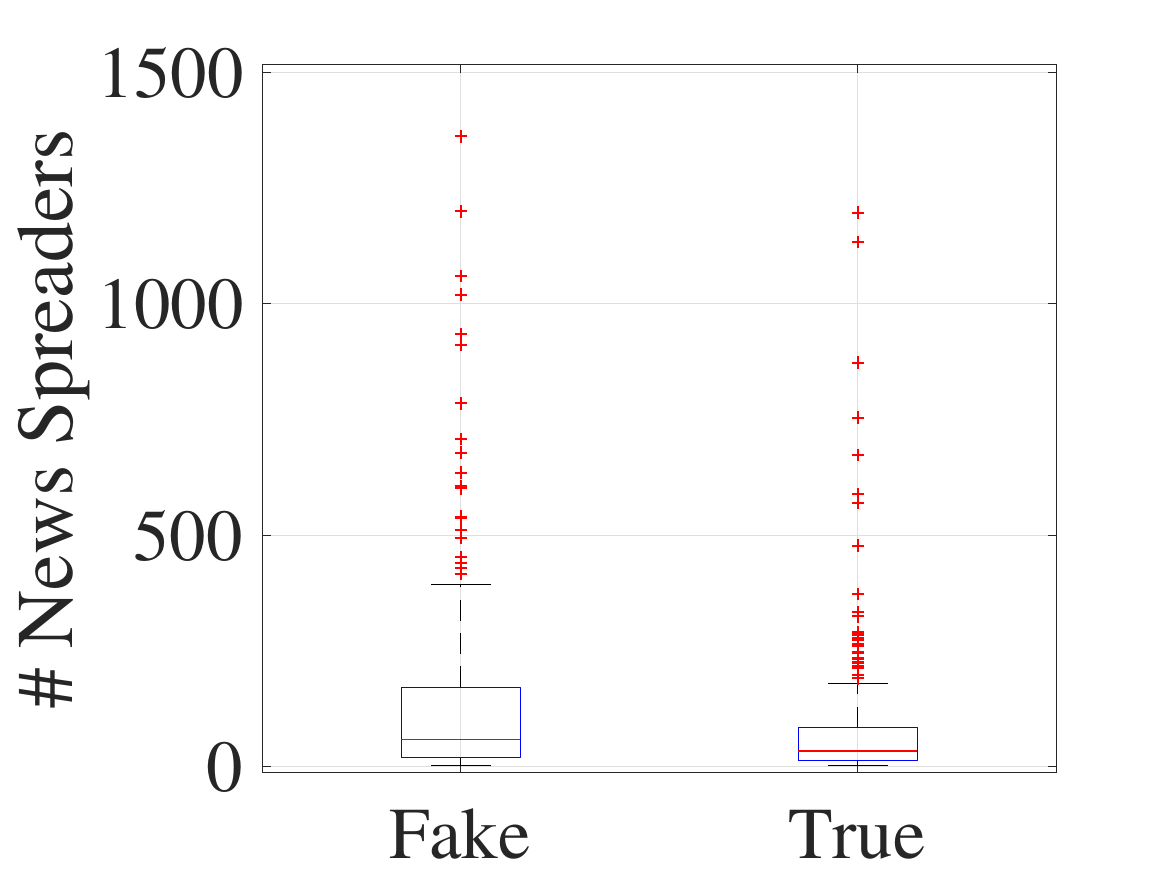}}
    \subfigure[Stronger-Engagements]{
    \includegraphics[width=.235\textwidth]{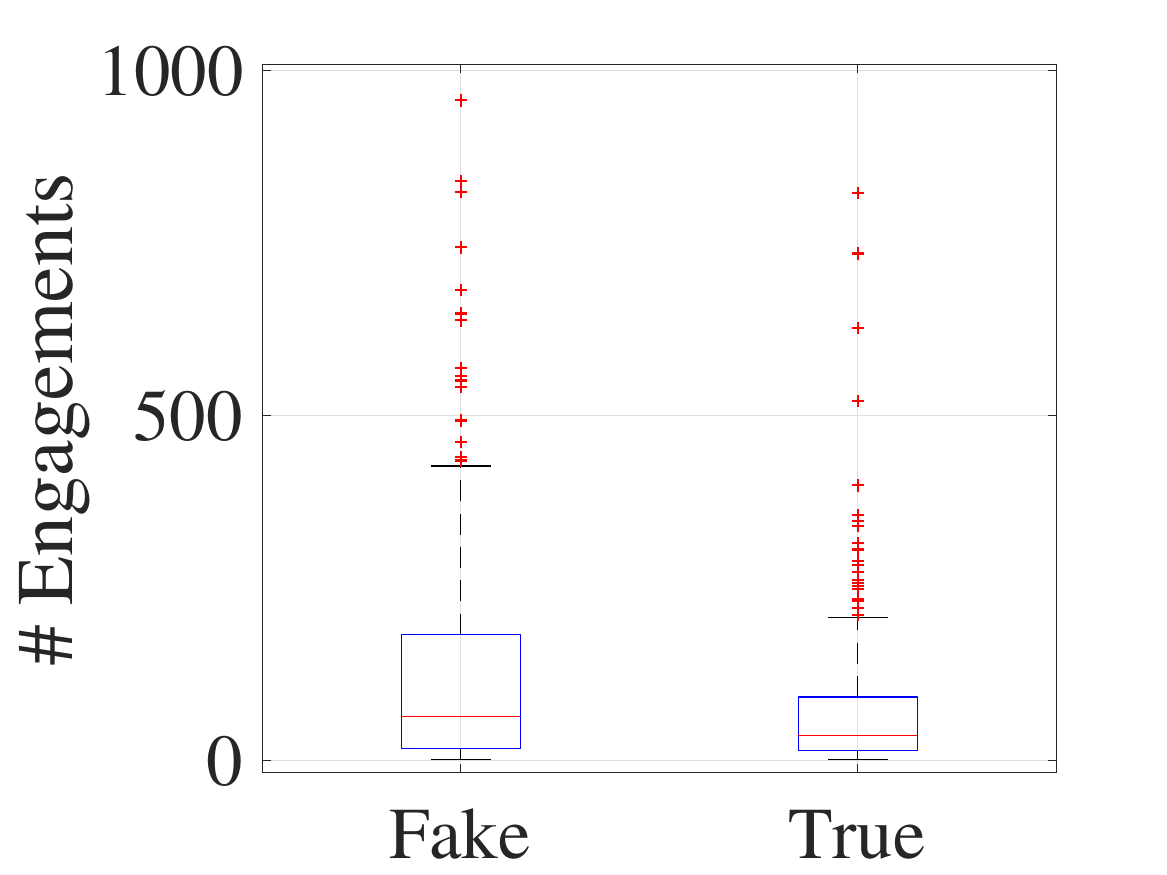}}
    \subfigure[Denser-Networks]{
    \includegraphics[width=.235\textwidth]{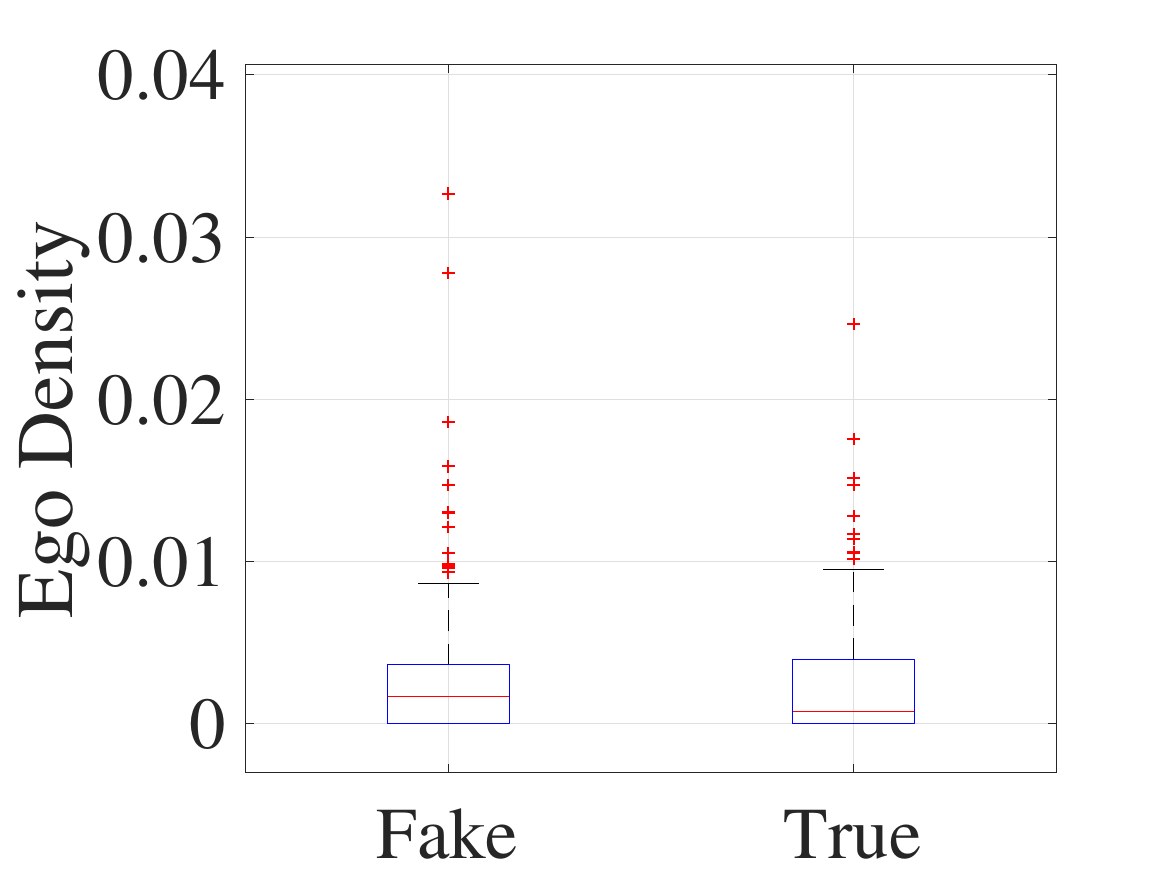}}
    \vspace{-1mm}
    \caption{Fake News Network-based Propagation Patterns~\cite{zhou2019network} (PolitiFact+BuzzFeed, data are from FakeNewsNet~\cite{shu2018fakenewsnet}). Compared to the truth, fake news can (a) spread farther and (b) attract more spreaders, where these spreaders are often (c) more strongly engaged with fake news and (d) more densely connected within fake news spreader networks.}\vspace{-3mm}
    \label{fig:pattern_network}
\end{figure}

\paragraph{$\blacktriangleright$ Homogeneous Network} Homogeneous networks are networks containing a single type of node and a single type of edge. One example is the \textit{news spreader network} (see Fig. \ref{subfig:spreaderNet}), a subgraph of social network of users, where each network is for a news article; each node in the network is a user spreading the news; and an edge between two nodes indicates the following relationship of two news spreaders~\cite{zhou2019network}. Classifying a news article (fake or true) using its spreader network is equivalent to classifying the network. Recently, Zhou et al. analyze such networks at the level of \textit{node}, \textit{ego}, \textit{triad}, \textit{community}, and the overall \textit{network}~\cite{zhou2019network}, respectively, and reveal four patterns within fake news spreader networks: 
(1)  \textsc{More-Spreader Pattern}, i.e., more users spread fake news than true news; 
(2) \textsc{Farther-Distance Pattern}, i.e., fake news spreads farther than true news;
(3) \textsc{Stronger-Engagement Pattern}, i.e., spreaders engage more strongly with fake news than with true news; and 
(4) \textsc{Denser-Network Pattern}, i.e., fake news spreaders form denser networks compared to true news spreaders (see Fig. \ref{fig:pattern_network}).
Another example of a homogeneous network is a \textit{stance network}~\cite{jin2016news}, where nodes are news-related posts by users and edges represent supporting (+) or opposing (-) relations among each pair of posts, e.g., the similarity between each pair of posts that can be calculated using a distance measure such as Jensen-Shannon divergence~\cite{jin2016news} or Jaccard distance~\cite{jin2014news}. The stance network is shown in Fig. \ref{subfig:stanceNet}. 
Fake news detection using a stance network boils down to evaluating the credibility of news-related posts (i.e., lower credibility~=~fake news), which can be further cast as a graph optimization problem. Specifically,
let $\mathbf{A} \in \mathbb{R}^{n \times n}$ denote the adjacency matrix of the aforementioned stance network with $n$ posts and $\mathbf{c}=  (c_1, \dots, c_n)$ denote the vector of post credibility scores. Assuming that supporting posts have similar credibility scores, the cost function in \cite{zhou2004learning} can be used and the problem can be defined as
\begin{equation} \label{eq::opt1}
\arg \underset{\mathbf{c}}{\min}~\mu ||\mathbf{c}-\mathbf{c}_0||^2 + (1-\mu) \sum_{i,j=1}^n \mathbf{A}_{ij} (\frac{\mathbf{c}_i}{\sqrt{\mathbf{D}_{ii}}}- \frac{\mathbf{c}_j}{\sqrt{\mathbf{D}_{jj}}})^2,
\end{equation}
where $\mathbf{c}_0$ refers to true credibility scores of training posts, $\mathbf{D}_{ij}=\sum_k\mathbf{A}_{ik}$, and $\mu \in [0,1]$ is a weight parameter.

\paragraph{$\blacktriangleright$ Heterogeneous Network} Heterogeneous networks have multiple types of nodes or edges. An early instance is a (users)-(tweets)-(news events) network using which Gupta et al.~\cite{gupta2012evaluating} evaluate news credibility by designing an algorithm similar to PageRank~\cite{page1999pagerank}. Their algorithm determines credibility relationships among these three entities with the assumption that users with low credibility tend to post tweets that are often related to fake news. Another example is the network capturing relationships among news publishers, news articles, and news spreaders (users) shown in Fig. \ref{subfig:heteroNet1}. Using this network, Shu et al.~\cite{shu2019beyond} classify news articles (as fake news or true news) using a linear classifier, where
\begin{itemize}
\item[(\textbf{I})] News article are represented using latent features, derived using Nonnegative Matrix Factorization (NMF):
\begin{eqnarray}
\min \left \| \mathbf{X}-\mathbf{DV}^{\top} \right \|_{F}^{2} 
& \text{s.t.}~\mathbf{D,V}\geq 0,
\end{eqnarray}
where $\mathbf{X} \in \mathbb{R}_+^{m \times t}$ is the given article-word matrix for $m$ news articles. $\mathbf{X}$ will be factorized as $\mathbf{D} \in \mathbb{R}_+^{m \times d}$ (i.e., article-latent feature matrix) and some weight matrix $\mathbf{V} \in \mathbb{R}_+^{t \times d}$;
\item[(\textbf{II})] Assuming that the political bias for each publisher is known: left ($-1$), least-biased ($0$), or right ($+1$), let $\mathbf{b} \in \{-1,0,1\}^{l}$ denote the political biases for $l$ publishers. Assuming that the political biases of publishers can be represented by the latent features of their published articles, the publisher-article relationship is derived by
\begin{equation} \label{eq:news_publisher}
\min\left \| \bar{\mathbf{P}}\mathbf{Dq} -\mathbf{b}\right \|_{2}^{2}, 
\end{equation}
where $\bar{\mathbf{P}} \in \mathbb{R}^{l \times m}$ is the normalized publisher-article relation (adjacency) matrix, and $\mathbf{q} \in \mathbb{R}^{d}$ is the weights; and
\item[(\textbf{III})] Assuming that non-credible (credible) users spread fake (true) news, a similar optimization formulation can be utilized to derive the relationship between spreaders (users) and new articles.

\end{itemize}

{Finally, Zhang et al. develop a framework for a heterogeneous network of users (news authors), news articles, and news subjects to detect fake news (see Fig.~\ref{subfig:heteroNet2}). They introduce \textit{Gated Diffusive Units} (GDUs), a neural network component that can jointly learn the representations of users, news articles, and news
subjects~\cite{zhang2018fake}.}

\vspace{-2mm}
\paragraph{$\blacktriangleright$ Hierarchical Network} In hierarchical networks, various types of nodes and edges form set-subset relationships (i.e., a \textit{hierarchy}). {One example is the news-tweet-retweet-reply network (see Fig.~\ref{subfig:hieraNet1}), an extension of the news cascade defined in Definition \ref{def::fnCascade}~\cite{shu2019hierarchical}. Hence, the same features listed for news cascades in Table \ref{tab:cascade_features} (e.g., cascade depth and size) can be used to predict fake news using this hierarchical network within a traditional ML framework.}
Another example of a hierarchical network is shown in Fig. \ref{subfig:hieraNet2}, which includes relationships across (i.e., \textit{hierarchical relationships}) and within (i.e., \textit{homogeneous relationships}) news events, sub-events, and posts. In such networks, news verification can be transformed into a graph optimization problem~\cite{jin2014news}, extending the optimization in Eq. (\ref{eq::opt1}).

\subsection{Discussion}
\label{subsec:propagation_discussion}

We have discussed the current solutions for detecting fake news by investigating how news spreads online. By involving dissemination information (i.e., social context) in fake news detection, propagation-based methods are more robust against writing style manipulation by malicious entities. However, propagation-based fake news detection is inefficient for \textit{fake news early detection} (see Section \ref{sec::discussion} for more details) as it is difficult for propagation-based models to detect fake news before it has been disseminated, or to perform well when limited news dissemination information is available. 

Furthermore, mining news propagation and news writing style allow one to assess news intention. As discussed, the intuition is that (1) news created with a malicious intent, that is, to mislead and deceive the public, aims to be ``more persuasive'' compared to those not having such aims, and (2) malicious users often play a part in the propagation of fake news to enhance its social influence~\cite{leibenstein1950bandwagon}. However, to evaluate if news intentions are properly assessed one relies on the ground truth (news labels) in training datasets often annotated by domain experts. This ground truth dependency particularly exists when predicting fake news by (semi-) supervised learning within graph optimization~\cite{shu2019beyond}, traditional statistical learning~\cite{zhou2019network}, or deep neural networks~\cite{zhang2018fake}. Most current fake news datasets have not provided a clear-cut declaration on whether the annotations within the datasets consider news intention, or how the annotators have manually evaluated it, which motivates the construction of fake news datasets or repositories that provide information on news intention.
    
Finally, research has shown that \textit{political fake news spreads faster, farther, more widely, and is more popular with a higher structure virality score than fake news in other domains such as business, terrorism, science, and entertainment}~\cite{vosoughi2018spread}. Discovering more dissemination patterns of fake news are hence highly  encouraged by comparing it with true news, or fake news from different domains and languages. Such patterns can deepen the public understanding of fake news and enhance explainability of fake news detection (we discuss \textit{explainable fake news detection} in Section \ref{sec::discussion}).

\section{Source-based Fake News Detection}
\label{sec::credibility}

{One can detect fake news by assessing the \textit{credibility} of its source, where credibility is often defined in the sense of quality and believability -- ``offering reasonable grounds for being believed''~\cite{castillo2011information,viviani2017credibility}. 
As presented in Fig.~\ref{fig::framework}, there are three stages within a [fake] news life cycle: being created, published online, and propagating on social media. This section will present how credibility of news stories can be assessed based on that of their sources at each stage. In other words, we deem ``sources'' as a general concept that includes (I) the sources creating the news stories, i.e., the news writers (Section \ref{subsec:publisherCred}), (II) the sources that publish the news stories, i.e., the news publishers (Section \ref{subsec:publisherCred}), as well as (III) the sources that spread the news stories on social media, i.e., the social media accounts (users, Section \ref{subsec:userCred}). We combine (I) and (II) into one subsection for two reasons: (i) not many studies have investigated news authors and (ii) news authors and publishers often form an employee-employer relationship; hence, their credibilities intuitively should have some correlations.
Note that though the role of news authors, publishers, and spreading users in detecting fake news has been illustrated in Sections \ref{sec::knowledge} to \ref{sec::propagation} (see related work such as \cite{zhou2019network,shu2019beyond,zhang2018fake}), the focus has been on each news story. In contrast, in this section, the focus is on each author, publisher, and user who might have written, published, or spread the news stories. That is why we, to some extent, regard assessing the credibility of news sources as an indirect way to detect fake news, i.e., one can consider news articles from unreliable news sources as fake news though it is not unlikely for these sources to post true news. Such an approach to detecting fake news might seem arbitrary but is efficient~\cite{norregaard2019nela}, as evidence has revealed that many fake news stories come from either fake news websites that only publish hoaxes, or from hyper-partisan websites that present themselves as publishing real news~\cite{silverman2016analysis}.}\vspace{-2mm}

\begin{figure}[t]
    \begin{minipage}{0.37\textwidth}
    \centering
    \includegraphics[width=\textwidth]{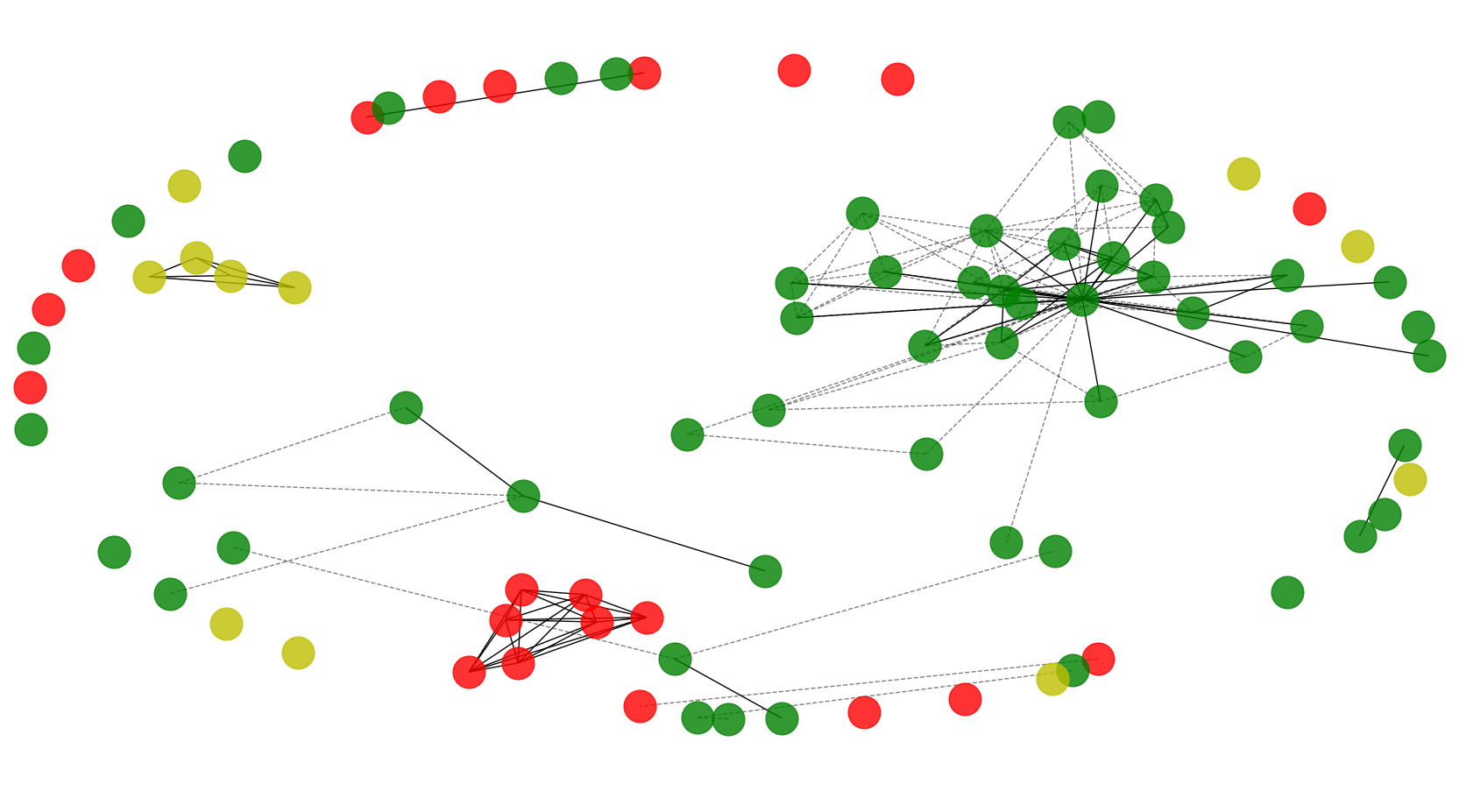}
    \caption{Coauthorship Network (directly from \cite{sitaula2020credibility}, data is from FakeNewsNet~\cite{shu2018fakenewsnet}): Red nodes represent authors only associated with [two or more] fake news articles; Green nodes represent authors only associated with [two or more] true news articles; yellow nodes represent authors writing both fake and true news; dashed lines indicate that two authors collaborate only once; and solid lines indicate that two authors collaborate at least twice.}
    \label{fig:authorCollaborationNet}
    \end{minipage}\qquad
    \begin{minipage}{0.55\textwidth}
    \centering
    \includegraphics[width=0.78\textwidth]{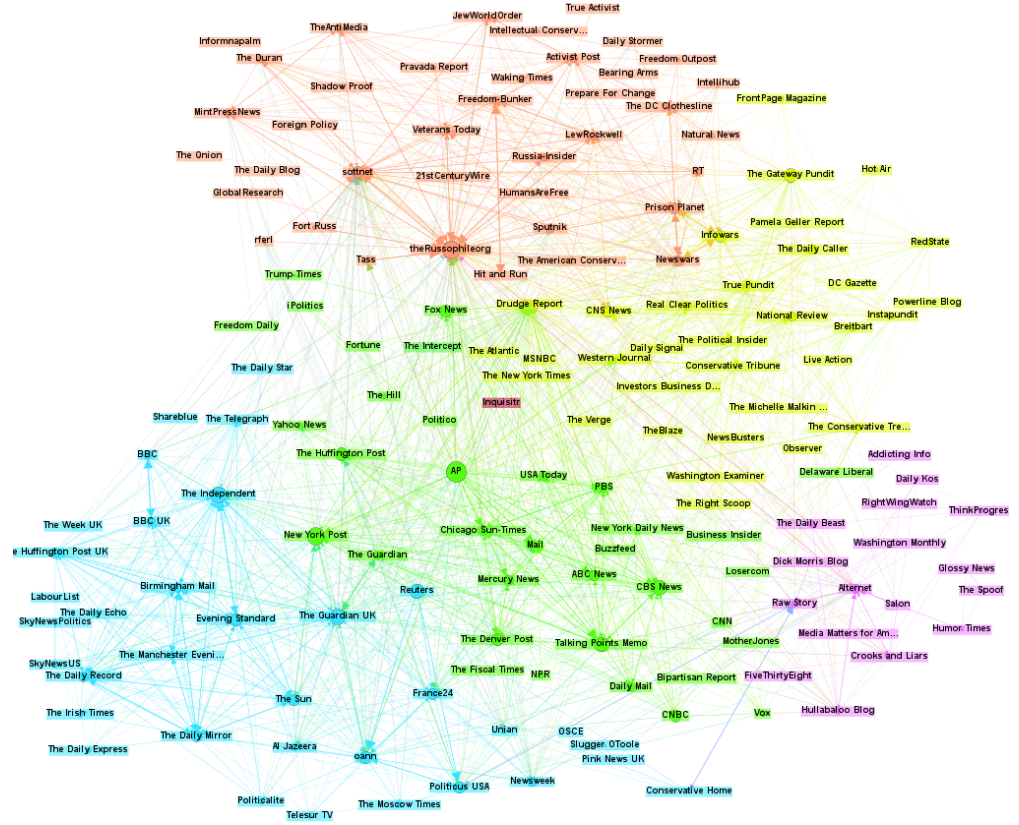}
    \vspace{-3mm}
    \caption{Content Sharing Network (directly from \cite{horne2019different}): Nodes are news publishers and edges are the flows of news articles among publishers. Orange: Russian/conspiracy community; Yellow: Right-wing/conspiracy community; Green: U.S. mainstream community; Magenta: Left-wing blog community; and Cyan: U.K. mainstream community.}
    \label{fig:contentSharingNet}
    \end{minipage}
\end{figure}

{\subsection{Assessing Source Credibility based on News Authors and Publishers}
\label{subsec:publisherCred}

Based on existing related studies, we organize this subsection as follows. We first present (I)~some patterns shared among (un)reliable authors or publishers. These patterns can help assess the credibility of unknown authors or publishers by exploring their relationships with other authors or publishers. Next, we review (II)~Web spam detection as a representation of the way to detect unreliable websites. Such methods can be used to identify spam publishers. Finally, we introduce (III)~resources that can help obtain the credibility (or political bias) of news publishers (ground truth).\vspace{-2mm}

\paragraph{I. Patterns of (Un)reliable Authors or Publishers.}
Research has shown that news authors and publishers exhibit \textit{homogeneity} in the networks that they form. Specifically, Sitaula et al.~\cite{sitaula2020credibility} construct the collaboration network of news authors (i.e., \textit{coauthorship network}) based on FakeNewsNet dataset~\cite{shu2018fakenewsnet}. The network is presented in Fig.~\ref{fig:authorCollaborationNet}. In this network, each node represents a news author, and the edge between two authors indicates that they collaborate in writing one (the dashed line) or more (the solid line) news articles. All nodes (authors) are grouped as (i) \textit{true-news authors} (green nodes) who are associated with two or more true news stories; (ii) \textit{fake-news authors} (red nodes), who are associated with two or more fake news stories; and (iii) the authors who have published both fake and true news stories (yellow nodes). As we can observe from Fig.~\ref{fig:authorCollaborationNet}, the coauthorship network exhibits homogeneity: the authors within the same group are more densely connected compared to the authors from different groups. Similarly, such homogeneity has been observed in the network formed by news publishers (i.e., \textit{content sharing network}). See Fig.~\ref{fig:contentSharingNet} for an illustration~\cite{horne2019different}. In this directed network, each node represents a news publisher, and the edge between two publishers indicates the flow of news articles. Nodes (publishers) can be clearly grouped into five communities:
(1)~Russian/conspiracy community (orange nodes); 
(2)~right-wing/conspiracy community (yellow nodes); 
(3)~U.S. mainstream community (green nodes);
(4)~left-wing blog community (magenta nodes); and 
(5)~U.K. mainstream community (cyan nodes), representing various nations and importantly, various political biases. 
}

\paragraph{II. Web (Publisher) Spam Detection.}
News publishers often publish their news articles on their own websites. Detecting unreliable publishers thus can be reduced to detecting low-credible websites. To assess website credibility, many practical techniques have been developed such as Web ranking algorithms. Traditional Web ranking algorithms such as PageRank~\cite{page1998pagerank} and HITS~\cite{kleinberg1999authoritative} assess website credibility to improve search engine responses to user search queries. However, the weaknesses of these Web ranking algorithms provide opportunities for \textit{Web spam}, to try to unjustifiably improve website rankings; thus, motivating the development of \textit{Web spam detection}. A comprehensive survey is available in \cite{spirin2012survey}. Web spam can be categorized as 
(i)~\textit{content spam}, which leads to a spam webpage appearing among normal search results primarily due to fake word frequencies (e.g., TF-IDF scores). Content spam includes spamming of title, body, meta-tags, anchors, and URLs; 
(ii)~[outgoing and incoming] \textit{link spam}, where the former mostly targets algorithms similar to HITS to achieve high hub scores, and the latter enhances website authority scores by attacking algorithms similar to PageRank; and 
(iii)~other types of spam such as \textit{cloaking}, \textit{redirection}, and \textit{click spams}. Algorithms to identify Web spam thus can be classified into 
(i)~content-based algorithms, which analyze Web content features, such as word counts and content duplication~\cite{fetterly2005detecting,ntoulas2006detecting,baly2018predicting}; 
(ii)~link-based algorithms, which detect Web spam by utilizing graph information~\cite{zhou2009osd,horne2019different}, learning statistical anomalies~\cite{dong2015knowledge}, and performing techniques such as (dis)trust propagation~\cite{gyongyi2004combating}, link pruning~\cite{bharat1998improved} and graph regularization~\cite{abernethy2010graph}; and 
(iii)~other algorithms that are often based on click streams~\cite{dou2008click} or user behavior~\cite{liu2015towards}.

\paragraph{III. Resources for Understanding News Publishers}
We introduce several resources that can help obtain the ground truth on the credibility (or political bias) of news publishers.
One resource is the \textit{Media Bias/Fact Check} website,\footnote{\url{https://mediabiasfactcheck.com/}} which provides a list of media along with their political slant: left, left-center, least biased, right-center, and right.
Another source is NewsGuard,\textsuperscript{\ref{note:newsguard}} which relies on expert-based evaluations and rates the reliability of a news source in terms of nine criteria: if it repeatedly publishes false content, responsibly presents information, regularly corrects or clarifies errors, responsibly handles the difference between news and opinion, avoids deceptive headlines, discloses ownership and financing, clearly labels advertising, reveals who's in charge that includes any possible conflicts of interest, and provides information about content creators. Finally, a system named MediaRank ranks news sources associated with their peer reputation, reporting bias, bottomline financial pressure, and popularity~\cite{ye2019mediarank}.

{\subsection{Assessing Source Credibility based on Social Media Users}
\label{subsec:userCred}

Social media users can be the initiating source for a news story spreading on social media. Intuitively, users with low credibility are more likely to become the spreading source of a fake news story than reliable users. Here we define and group such low-credible users as (I) malicious users and (II) normal users that are vulnerable to spreading fake news (or any disinformation).

\paragraph{I. Identifying Malicious Users}
Identifying malicious users can often be reduced to detecting \textit{social bots}, a software application that runs automated tasks (scripts) over the Internet. Note that social bots can be benign and designed to provide useful services, e.g., automatically aggregating content from different sources, like simple news feeds. 


Meanwhile, some bots are created to harm by manipulating social media discourse and deceiving social media users, for example, by spreading fake news~\cite{ferrara2016rise}.
One recent study -- that classifies accounts based on observable features such as sharing behavior, number of social ties, and linguistic features -- estimates that between 9 to 15 percent of active Twitter accounts are bots~\cite{varol2017online}.
It has been suggested that millions of social bots have participated in online discussions around the 2016 U.S. presidential election,\footnote{\url{http://comprop.oii.ox.ac.uk/research/public-scholarship/resource-for-understanding-political-bots/}} and some of the same bots were later used to attempt to influence the 2017 French election~\cite{ferrara2017disinformation}. Research has significantly contributed to differentiating bots from humans on social media. For example, a feature-based system, Botometer (formally BotOrNot),\footnote{\url{https://botometer.iuni.iu.edu/}} extracts six main groups of features (\textit{network}, \textit{user}, \textit{friend}, \textit{temporal}, \textit{content}, \textit{sentiment}) of a Twitter account and uses random forests to predict whether it is a bot or not with a 0.95 AUC (Area Under the ROC Curve) performance. Similarly, Cai et al. propose a bot detection model that uses deep learning to learn features from both user posts and behavior~\cite{cai2017detecting}. Morstatter et al. work on striking the balance between precision and recall value in predicting social bots~\cite{morstatter2016new}.
From 14 million messages spreading 400 thousand articles on Twitter during a ten-month period in 2016 and 2017, Shao et al. find evidence that bots spread unreliable news at an earlier time compared to humans to increase the chances that an article goes ``viral.'' (see Fig.~\ref{fig:bot1})~\cite{shao2018spread}. They also discover that humans do most of the retweeting, and they retweet articles posted by [likely] bots almost as much as those by other humans (see Fig.~\ref{fig:bot2}). Such discoveries demonstrate the difficulty in recognizing social bots and the vulnerability of humans (or normal users) to online information with low credibility. Next, we will discuss the possible ways to identify this group of vulnerable normal users, which, thus far, has been rarely investigated.

\begin{figure}[t]
    \begin{minipage}{0.41\textwidth}
    \vspace{17mm}
    \includegraphics[width=\textwidth]{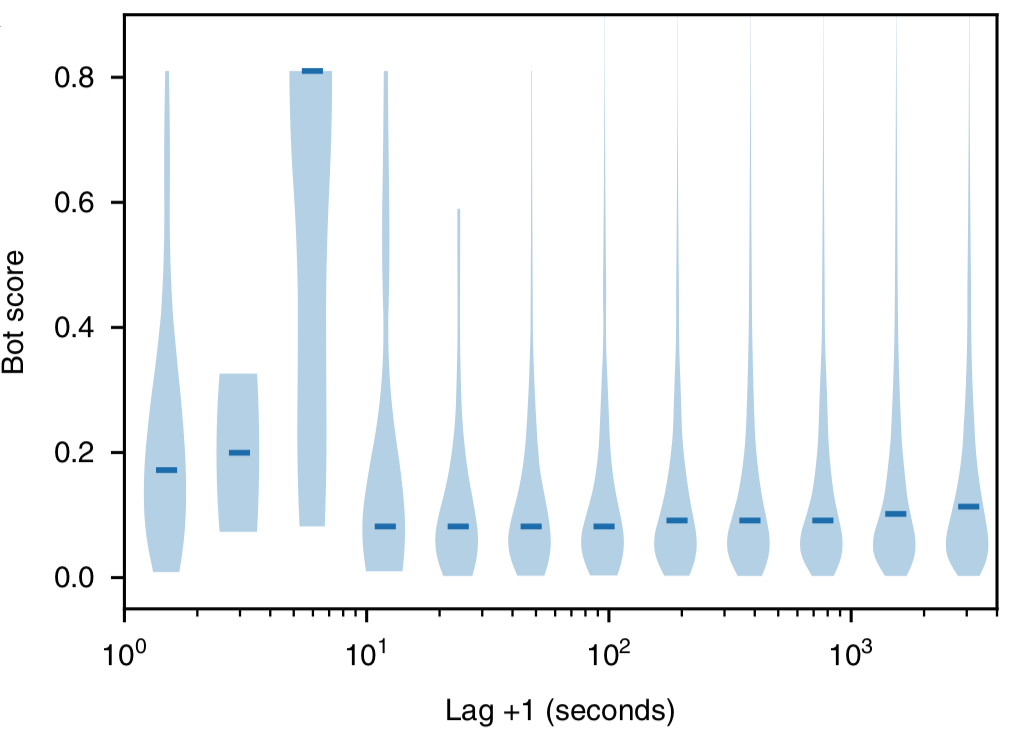}
    \caption{Temporal Engagements of Social Bots (directly from \cite{shao2018spread}), which indicates that bots spread unreliable news at an earlier time compared to humans.}
    \label{fig:bot1}
    \vspace{-3mm}
    \end{minipage}\quad
    \begin{minipage}{0.55\textwidth}
    \includegraphics[width=\textwidth]{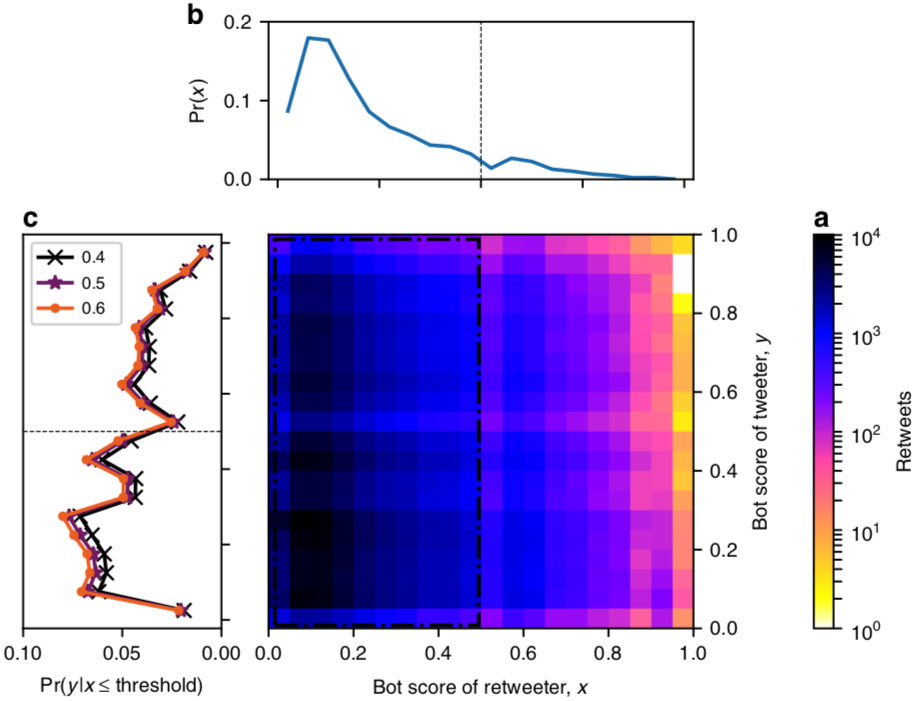}
    \caption{Impacts of Bots on Humans (directly from \cite{shao2018spread}). (\textbf{a}): Joint distribution of bot scores for tweeters and retweeters of unreliable news; (\textbf{b}): Distribution of bot scores for retweeters, who are mostly humans; and (\textbf{c}): Distribution of bot scores for tweeters, where bots and humans both share a significant proportion.}
    \label{fig:bot2}
    \vspace{-3mm}
    \end{minipage}
\end{figure}

\paragraph{II. Identifying Vulnerable Normal Users}
Fake news, unlike information such as fake reviews~\cite{jindal2008opinion}, can ``attract'' both malicious and normal users, where each group is attracted for different reasons (i.e., different intentions). Malicious users such as some social bots~\cite{ferrara2016rise}) often spread fake news intentionally. In contrast, some normal users can frequently and unintentionally do so without recognizing the falsehood. Compared to normal users that are ``immune" to fake news, these normal users are ``susceptible" and vulnerable to fake news, i.e., their credibility is relatively lower compared to other normal users. Theories in Table \ref{tab::theories} imply that a user engages in spreading fake news less intentionally when spreading bears a greater (i)~\textit{social influence}, e.g., more users spreading the news ~\cite{leibenstein1950bandwagon,deutsch1955study,ashforth1989social,kuran1999availability,boehm1994validity,macleod1986attentional} and (ii)~\textit{self-influence}, e.g., high similarity between the knowledge within the fake news and the preexisting knowledge of the user~\cite{nickerson1998confirmation,freedman1965selective,fisher1993social}. Preexisting knowledge can be captured by user posts. Nevertheless, few studies have quantified the impact of social and self-influence on the vulnerability of users to fake news, or have considered the effect of intention in assessing user credibility and further in fake news intervention (see Section \ref{sec::discussion} for more discussions on \textit{fake news intervention}).}

\section{Discussion and Future Work}
\label{sec::discussion}

While we have detailed four fake news detection strategies (\textit{knowledge}-based, \textit{style}-based, \textit{propagation}-based and \textit{source}-based) separately in Sections \ref{sec::knowledge}-\ref{sec::credibility}, they are not independent. Predicting fake news jointly from multiple perspectives is encouraged, where one can combine their strengths. Furthermore, there is a motivation to detail open issues shared among different strategies. Based on fake news characteristics and current state of fake news research, we highlight the following potential research tasks that can facilitate a deeper understanding of fake news, and improve the interpretability and performance of current fake news detection studies.

\paragraph{I. Detection of Non-traditional Fake News} Based on how fake news was defined, fake news can also be news articles with (1) outdated knowledge, e.g., ``Britain has control over fifty-six colonial countries'' or (2) false claims in \underline{some} parts of the news content (text and images), i.e., news content is partially (in)correct. These non-traditional forms of fake news emphasize different aspects of detection and motivate one to develop more thorough and comprehensive detection strategies. For example, to detect fake news with outdated knowledge, one is encouraged to construct a dynamic knowledge graph; To detect fake news that is only partially correct, extending fake news detection to a multi-label classification or regression problem may be more appropriate than defining it as a binary classification problem. 

\paragraph{II. Fake News Early Detection}
Fake news early detection aims to detect fake news at an early stage before it becomes wide-spread, so that one can take early actions for fake news mitigation and intervention. Early detection is especially crucial for fake news as the more fake news spreads, the more likely for people to trust it (i.e., \textit{validity effect}~\cite{boehm1994validity}). 
To detect fake news at an early stage, one has to primarily and efficiently rely on news content and limited social context information, which leads to facing multiple challenges. First, newly emerged events often generate new and unexpected knowledge that has not been stored in existing knowledge graphs, or is difficult to be inferred.  Second, features that have well-represented the style of fake news in one domain or in the past, may not be as useful in the other domains or in the future, especially due to the different patterns of fake news in different domains, and the constant evolution of deceptive writing style~\cite{castelo2019topic}. Finally, limited information may adversely impact the performance of machine learning methods. To address these challenges and detect fake news early, one can focus on\vspace{-1mm}
\begin{enumerate}[leftmargin=*]
    \item \textit{timeliness of ground truth}, for example, technologies related to dynamic (real-time) Knowledge Graph (KG) construction should be developed to facilitate timely updates of the ground truth;
    \item \textit{feature compatibility}~\cite{wang2018eann}, specifically, features that can capture the commonalities of deceptive writing style across topics, domains, languages, and the like, as well as the evolution of deceptive writing style; and
    \item \textit{verification efficiency}~\cite{liu2018early,zhou2019content}, for example, by improving the efficiency of fake news detection by effectively using a small amount of information in a news article (e.g., headlines), or by identifying \textit{check-worthy content} and topics, which we will discuss next.
\end{enumerate}

\paragraph{III. Identifying Check-worthy Content} 
With new information created and circulated online at an unprecedented rate, identifying check-worthy content or topics can improve the efficiency of fake news detection and intervention by prioritizing content or topics that are check-worthy~\cite{hassan2017toward}. Whether a given news content or topic is check-worthy can be measured by, e.g., (i)~its \textit{newsworthiness} or potential to influence the society, or (ii)~its historical likelihood of being fake news. Thus a content or topic that is newsworthy and generally favored by fake news creators is more check-worthy. Newsworthiness can be assessed in many ways, e.g., by verifying if the title of the news article is a clickbait~\cite{zhou2019content}, if its content will trigger widespread discussions on social media~\cite{vosoughi2018spread}, or if its topic relates to national affairs and matches with public concerns~\cite{hassan2017toward}. As shown in Fig. \ref{fig::politiFact}, the historical likelihood of a topic being fake news is available in some online resources, e.g., ``the PolitiFact scorecard''\textsuperscript{\ref{url:politifact}} gives statistics on the authenticity distribution of all the statements on a specific topic~\cite{zhang2018fake}.

{Furthermore, identifying check-worthy portions within a news content (e.g., sentences) is a pathway to \textit{explainable fake news detection}, which we will also discuss as a potential research task in this section. A news portion that contributes more to detecting fake news is more check-worthy; thus, an \textit{attention mechanism} becomes a natural choice for identifying (i.e., weighting) such portions in a \textit{Recurrent Neural Network}-based fake news detection model~\cite{shu2019defend}. Expert-based analyses and justifications provided for each verified news article on fact-checking websites (detailed in Section \ref{subsec::manual}) can be potentially combined with such identified portions to provide explanations; however, to date, such research directions have not been well explored.}

\paragraph{IV. Cross-domain (-topic, -website, -language) Fake News Analysis} We highlight this potential research task for two reasons. First, current fake news studies emphasize on distinguishing fake news from truth with experimental settings that are generally limited to a particular social network and a specific language. Second, analyzing fake news across domains, topics, websites, and languages allows one to gain a deeper understanding of fake news and identify its unique non-varying characteristics, which can further assist in fake news early detection and in the identification of check-worthy content,  both of which we have discussed in this section.

\paragraph{V. Explainable Fake News Detection} Facilitating interpretability has been of great interest in artificial intelligence~\cite{vilone2020explainable} and machine learning~\cite{du2018techniques} research. For fake news detection models, interpretability can be provided by mining social feedback, e.g., the stance taken within (re-)posts and comments~\cite{shu2019defend}, and mining expert analyses available on fact-checking websites, while both have been rarely utilized. Model interpretability can also be provided and enhanced by conducting interdisciplinary research, e.g., by relying on fundamental theories identified in Table \ref{tab::theories} (see Section \ref{sec::theories}). While there have been studies that have conducted theory- or pattern-driven feature engineering to detect fake news within a traditional machine learning framework~\cite{zhou2019content,zhou2019network}, there has been limited research on using related theories or domain knowledge to guide the learning process in machine learning, e.g., in [deep] neural networks.

\paragraph{VI. Fake News Intervention} Fake news studies have emphasized the importance of new business models adopted by social media sites to address fake news intervention, which suggests shifting the emphasis from maximizing user engagement to that on increasing information quality, e.g., by using self- or government regulations~\cite{lazer2018science}. In addition to introducing new policies and regulations, efficiently blocking and mitigating the spread of fake news also demands technical innovations and developments. Technically, a fake news intervention strategy can be based on network structure, or based on users as we have discussed in Section \ref{sec::theories}. When intervening based on network structure, one aims to stop fake news from spreading by blocking its propagation paths, relying on analyzing the network structure of its propagation and predicting how fake news is going to further spread. From a user perspective, fake news intervention relies on specific roles users play in fake news dissemination. One such role is being an (i) \textit{influential user} (i.e., \textit{opinion leader}). When blocking certain fake news in a social network, targeting these influential spreaders leads to a more efficient intervention compared to those with a negligible social influence on others. Another beneficial role is being a (ii) \textit{corrector}, users on social networks who take an active role in mitigating the spread of fake news by attaching links to their posts or comments that debunk the fake news~\cite{vo2018rise}. Furthermore, the intervention strategy for (iii) malicious users and (iv) normal users should be different, while they can both spread fake news; malicious users should be penalized, while normal users should be assisted to improve their ability to distinguish fake news. For example, \textit{personal recommendation} of true news articles and/or articles with refuting evidence can be helpful to normal users. These recommendation should not only cater to the topics that the users want to read but also those topics that they are most gullible to due to their political biases or preexisting knowledge.

\section{Conclusion}
\label{sec::conclusion}

This survey extensively reviews and evaluates current fake news research by
(1)~defining fake news, differentiating it from deceptive news, false news, satire news, misinformation, disinformation, clickbaits, cheery-picking, and rumors based on three characteristics: authenticity, intention, and being news; (II)~detailing interdisciplinary fake news research by firstly and comprehensively identifying related fundamental theories in, e.g., social sciences; 
(III)~reviewing the methods that detect fake news from four perspectives: the false  \underline{knowledge} fake news communicates, its writing \underline{style}, its \underline{propagation} patterns, and the credibility of its \underline{source}; and
(IV)~highlighting challenges in current research and some research opportunities that go with these challenges. 
As fake news research is evolving, we accompany this survey with an online repository \url{http://fake-news.site}, which will provide summaries and timely updates on the research developments on fake news, including tutorials, recent publications and methods, datasets, and other related resources.

\bibliographystyle{ACM-Reference-Format}
\bibliography{reference}

\end{document}